\RequirePackage{silence}
\documentclass{article}

\usepackage[numbers]{natbib}

\usepackage[final,eandd]{neurips_2026}

\usepackage[utf8]{inputenc} 
\usepackage[T1]{fontenc}    
\usepackage{hyperref}       
\hypersetup{
    colorlinks=true,
	linkcolor=blue,
	filecolor=magenta,      
	urlcolor=blue,
	citecolor=blue,
	pdfinfo={
        Title={ExpertBench},
        Subject={machine learning, large language models, datasets and benchmarks},
        Keywords={machine learning, large language models, datasets and benchmarks},
    }
}

\usepackage{url}            
\usepackage{booktabs}       
\usepackage{amsfonts}       
\usepackage{nicefrac}       
\usepackage{microtype}      
\usepackage{graphicx}
\usepackage{wrapfig}
\usepackage{caption}
\usepackage{subcaption}
\usepackage{amsmath}
\usepackage{amssymb}
\usepackage{mathtools}
\usepackage{amsthm}
\usepackage{enumitem}
\usepackage[capitalize,noabbrev]{cleveref}
\usepackage{listings}
\usepackage{epigraph} 

\usepackage{multirow}
\usepackage{verbatimbox}
\usepackage{tcolorbox} 
\tcbuselibrary{breakable,skins}
\usepackage{fancyvrb} 
\usepackage[dvipsnames,svgnames,table]{xcolor}
\usepackage[table]{xcolor}
\usepackage{makecell}
\usepackage{titletoc}
\usepackage{tabularx}
\usepackage{siunitx}
\usepackage{pgf}
\usepackage{longtable}

\newtcolorbox{CaseBox}{
  enhanced,
  breakable,
  colback=gray!10,
  colframe=black,
  width=\linewidth,
  arc=1mm,
  auto outer arc,
  boxrule=0.5pt,
  left=6pt,right=6pt,top=6pt,bottom=6pt,
  before skip=6pt,
  after skip=10pt
}

\newcommand{\BlackBar}[1]{%
  \par\smallskip\noindent
  \begingroup
  \setlength{\fboxsep}{4pt}%
  \colorbox{black}{%
    \parbox{\dimexpr\linewidth-2\fboxsep\relax}{\centering\color{white}\textbf{#1}}%
  }%
  \endgroup
  \par\smallskip\nopagebreak[2]
}

\usepackage{rotating}

\definecolor{pal-coral}{HTML}{F8807F}   
\definecolor{pal-apricot}{HTML}{FFBB74} 
\definecolor{pal-olivine}{HTML}{A5C476} 
\definecolor{pal-mint}{HTML}{B3EFCD}    
\definecolor{pal-aqua}{HTML}{81D7D4}    
\definecolor{Ai2Pink}{HTML}{F0529C}
\definecolor{DarkGray}{HTML}{b2b2b2}
\definecolor{LightGray}{HTML}{f2f2f2}
\colorlet{Ai2PinkLight}{Ai2Pink!70!white}

\hypersetup{
  colorlinks,
  linkcolor=black,
  urlcolor=blue
}

\theoremstyle{plain}

\theoremstyle{definition}

\theoremstyle{remark}

\usepackage{xspace}

\newcommand{\dataset}{\$OneMillion-Bench\xspace}
\newcommand{\benchmark}{\$OneMillion-Bench\xspace}

\newcommand\coauth{$^\star$}

\makeatletter
\DeclareRobustCommand\onedot{\futurelet\@let@token\@onedot}
\def\@onedot{\ifx\@let@token.\else.\null\fi\xspace}
\def\eg{\emph{e.g}\onedot} 

\def\ie{\emph{i.e}\onedot} 

\def\cf{\emph{cf.}\xspace}

\def\vs{\textit{vs}\onedot}

\makeatother

\usepackage{tikz}
\usetikzlibrary{calc}

\usepackage[table]{xcolor}
\usepackage{colortbl}
\usepackage{textcomp}
\usepackage{dsfont}

\usepackage{ifdraft}

\newcommand{\yang}[1]{\ifdraft{\textcolor{OliveGreen}{\textsf{\small[LY: #1]}}}{}}

\let\svthefootnote\thefootnote
\newcommand\freefootnote[1]{
  \begin{small}%
  \let\thefootnote\relax%
  \footnotetext{#1}%
  \let\thefootnote\svthefootnote%
  \end{small}%
}

\newcommand{\passofk}{pass\textasciicircum
k\ }
\newcommand{\passatk}{pass@k\ }

\newcommand{\WebLogo}{\raisebox{-0.15em}{\includegraphics[height=1.1em]{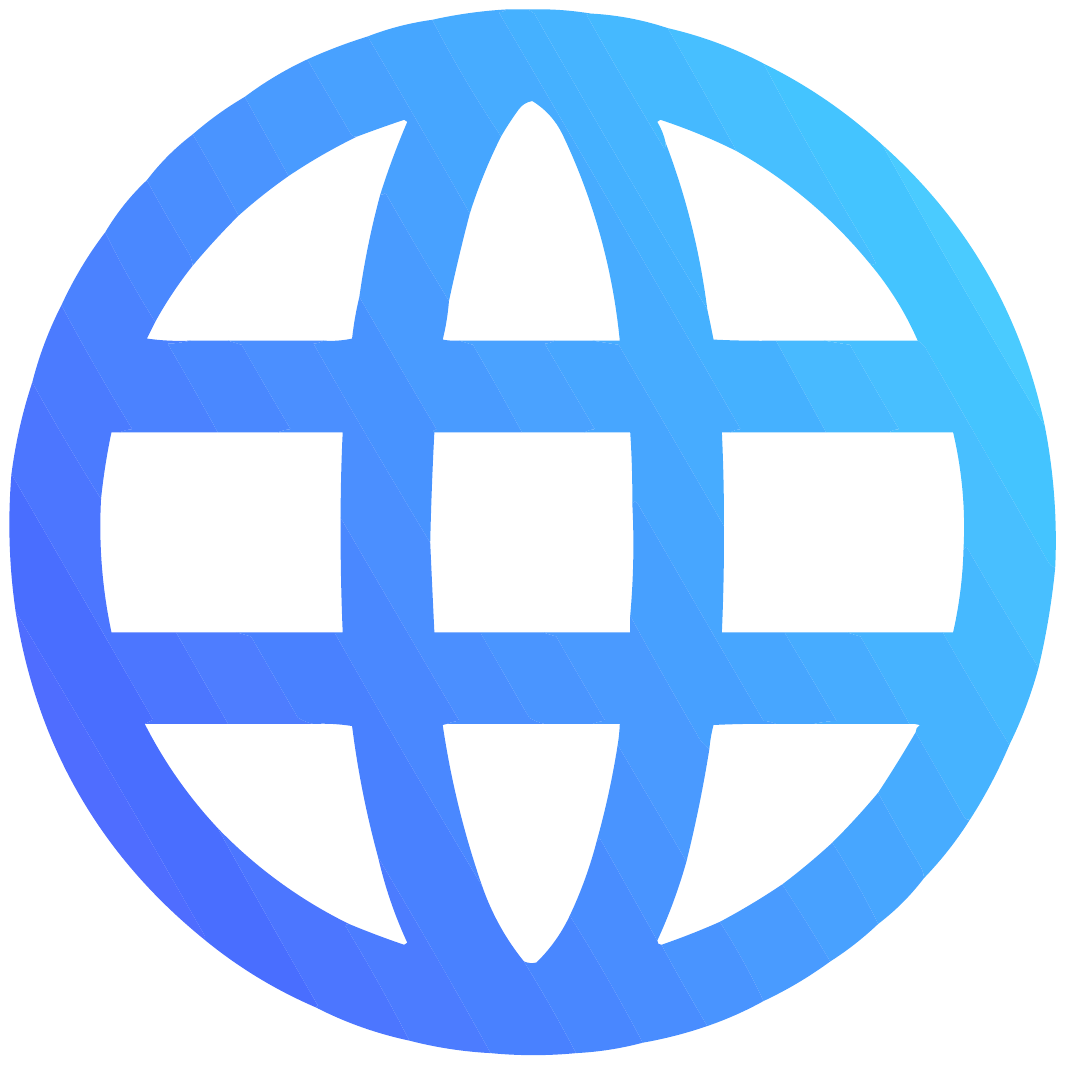}\xspace}}

\title{\$OneMillion-Bench:\\ How Far are Language Agents from Human Experts?}

\author{%
  Yang Liu$^2$\coauth{}\textnormal{,} Jiaqi Li$^2$\coauth{}\textnormal{,} Jun Bai$^2$\coauth{}\textnormal{,} Qianyu Yang$^1$\coauth{}\textnormal{,} \\
  \textbf{Xiaobo Hu}$^3$,
  \textbf{Tao Peng}$^1$,
  \textbf{Zaiyuan Wang}$^1$,
  \textbf{Ran Tian}$^1$,
  \textbf{Jiayun Dong}$^1$,
  \textbf{Chun Zhang}$^1$, \\
  \textbf{Zixia Jia}$^2$,
  \textbf{Kaiyuan Chen}$^3$, \textbf{Yixin Ren}$^3$, \textbf{Yang Liu}$^3$,  \textbf{Yanglihong Xiao}$^1$, \\
  \textbf{Lingyue Yin}$^1$, \textbf{Tiliang Duan}$^1$, \textbf{Ge Zhang}$^4$, \textbf{Gang Yao}$^2$, \textbf{Hao Chen}$^2$, \\
  \textbf{Yuan Gong}$^{3\dagger}$,
  \textbf{Jianpeng Jiao}$^{1\dagger}$, 
  \textbf{Zilong Zheng}$^{2\dagger}$
  \vspace{0.5em} \\
  $^1$\textbf{Humanlaya}, $^2$\textbf{BIGAI}, $^3$\textbf{xbench}, $^4$\textbf{M-A-P}
  \vspace{-1.5em}\\
}

\begin{document}
\maketitle
\freefootnote{\hspace{-1em}$^*$Co-first Authors. $\dagger$ Corresponding Authors. Complete list of dataset contributors is in Appendix \ref{app:authors}. Correspondence to \url{jasper@humanlaya.com} and \url{zlzheng@bigai.ai}.}

\begin{abstract}
As language models (LMs) evolve from chat assistants to long-horizon agents capable of multi-step reasoning and tool use, existing benchmarks remain largely confined to structured or exam-style tasks that fall short of real-world professional demands. To this end, we introduce \textbf{\$OneMillion-Bench} (\$OMB), a benchmark of 400 expert-curated tasks spanning Law, Finance, Industry, Healthcare, and Natural Science, built to evaluate agents across economically consequential scenarios. Unlike prior work, the benchmark requires retrieving authoritative sources, resolving conflicting evidence, applying domain-specific rules, and making constraint decisions, where correctness depends as much on the reasoning process as the final answer. We adopt a rubric-based evaluation protocol scoring factual accuracy, logical coherence, practical feasibility, and professional compliance, focusing on expert-level problems to ensure meaningful differentiation across agents. Together, \$OMB provides a unified testbed for assessing agentic reliability, professional depth, and an indicator of practical readiness in domain-intensive scenarios.
\end{abstract}

\begin{figure*}[hb!]
    \centering
    \vspace{-0.5em}
    \includegraphics[width=0.95\columnwidth]
    {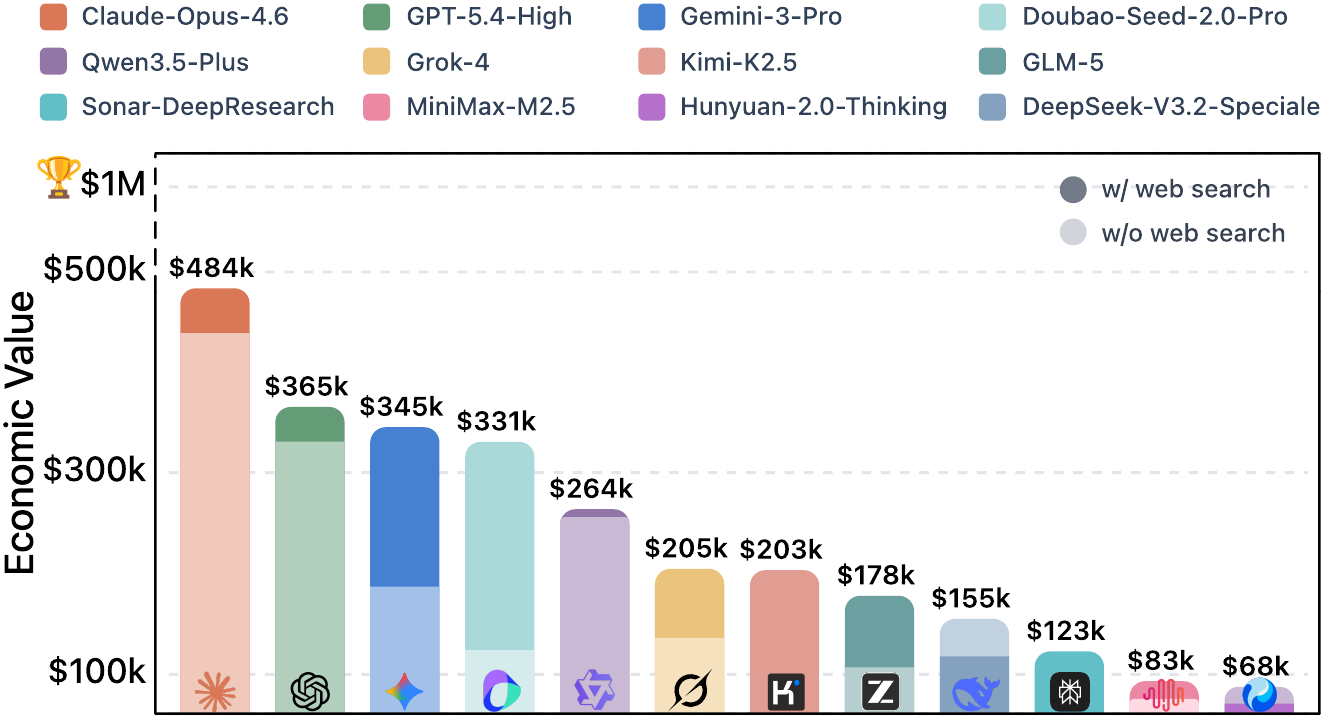} 
    \caption{Leaderboard performance on \dataset.}
    \label{fig:intro}
    \vspace{-3em}
\end{figure*}
\newpage


\section{Introduction}\label{sec:introduction}

Consider work performed daily in professional industries: a senior actuary auditing a life-insurance reserve valuation under IFRS 17; an M\&A lawyer verifying cross-border compliance clauses; or an investment analyst constructing a high-pressure valuation model. These workloads are not merely prompts to be answered, but context-heavy professional deliverables that require specialized knowledge, long-horizon execution, and strict adherence to domain constraints. While language agents are capable of supporting complex reasoning~\cite{jimenez2024swe,merrill2026terminal,phan2026benchmark,mialon2023gaia,chen2025xbench}, multi-turn interactions~\cite{yao2025tau,barres2025tau,xie2024travelplanner}, and realistic decision making~\cite{backlund2025vending,zeng2025futurex}, a critical gap remains between exam-style evaluations and the rigorous demands of professional labor. Existing benchmarks evaluate agents on solving isolated questions but lack clarity and provide limited insight into whether they can perform professional work and reliably create value in economically consequential professional scenarios~\cite{schwartz2025reality,jabbour2025evaluation,meimandi2025measurement}.

To effectively quantify the economic value that language agents can deliver in realistic tasks, we create \textbf{\dataset}, a comprehensive benchmark built around expert tasks designed to measure not only precision, but value. It spans high-impact professional domains, requires intensive retrieval, evidence-based reasoning, and tests reliability under explicit constraints. The benchmark comprises 400 challenging open‑ended tasks in five domains: Finance, Law, Healthcare, Natural Science, and Industry, curated through a collaborative effort spanning over 2,000 expert hours. Each task is assigned a realistic monetary value, calculated by the estimated time required for the task completion by a senior professional and its prevailing market hourly wage\footnote{\url{https://data.bls.gov/oes/\#/industry/000000}}\footnote{\url{https://rsj.gz.gov.cn/zwdt/tzgg/content/post_10619688.html}}. The estimated value of all tasks exceeds \$1 million, giving the benchmark its name. Performance on \dataset is thus expressed by translating agent capability directly into a value measure of professional utility.

\dataset offers several key features: (1) \textbf{Economic evaluation} that quantifies agentic capabilities through the lens of labour costs. It distinguishes the capabilities of language agents through how much they earn in million-value professional tasks. (2) \textbf{Comprehensive domain coverage} spans five professional areas, with balanced sub-domain partitioning. (3) \textbf{Skill-oriented taxonomy} including retrieval, reasoning, verbalization, and instruction following that enables fine-grained analysis, instead of overfitting evaluation to any domain idiosyncrasy.
(4) \textbf{Rubric-based evaluation} mechanism avoids evaluation metric gaming \cite{thomas2022reliance} and well aligns with domain-specific policies and expectations designed by domain experts to provide fine-grained automated scoring. 

Ultimately, \dataset is a step towards value quantification of language agents in real-world professional scenarios which turns capability into an interpretable quantity: how much reliable work an agent can actually deliver, and how much that work is worth, accelerating progress towards AI systems that are \emph{not only powerful, but trustworthy and economically meaningful}. \textbf{We release all the artifacts of \benchmark in \url{https://github.com/humanlaya/OneMillion-Bench}.}

\section{Constructing \dataset}

\subsection{Data Curation Pipeline}
\begin{figure*}[htbp]
    \centering
    \includegraphics[width=0.9\columnwidth]{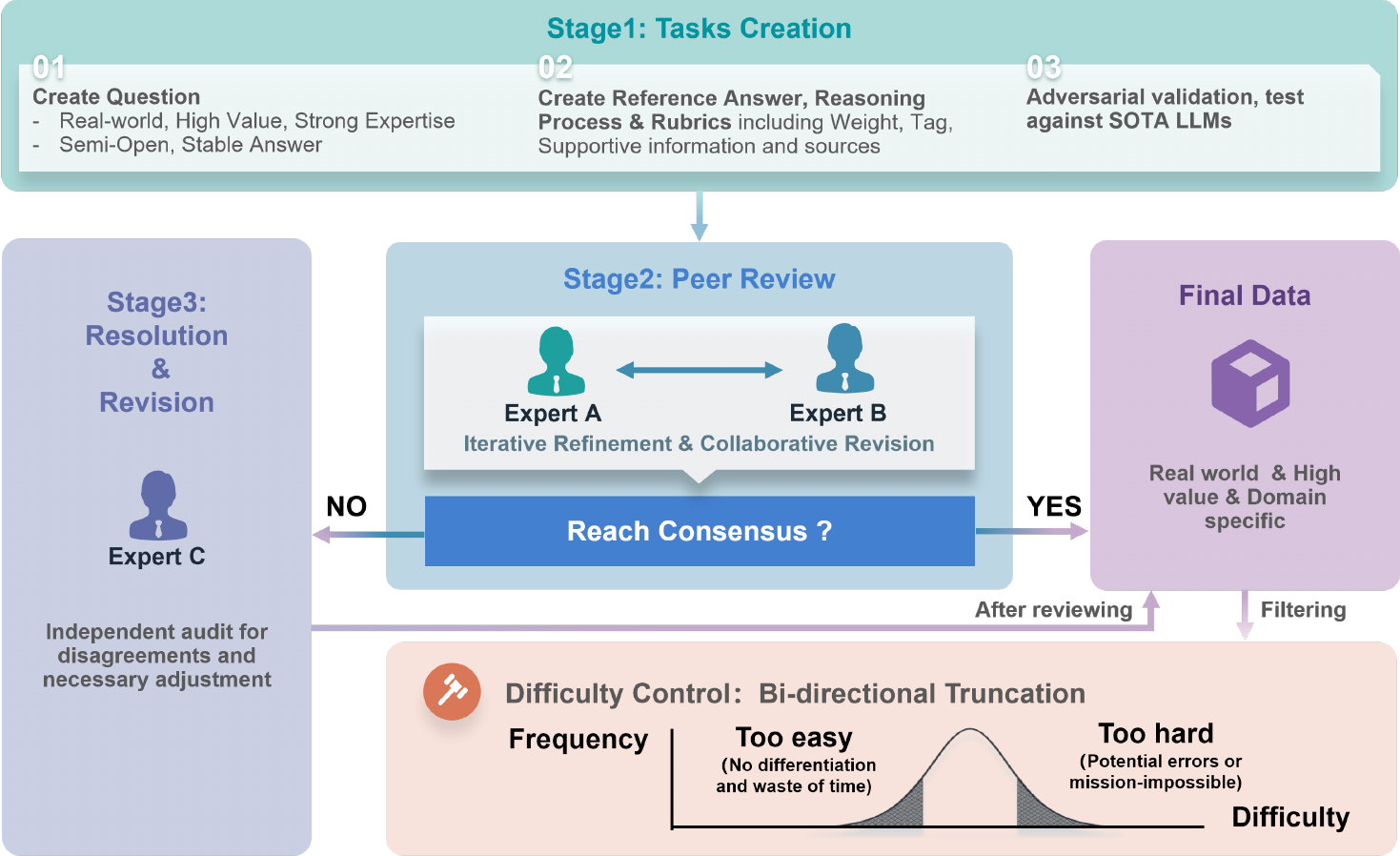} 
    \caption{Data curation pipeline of \dataset.}
    \label{fig:examples1}
    \vspace{-15pt}
\end{figure*}

Following the principle: (1) realistic workflow problems with high economic value; (2) tasks with objective and quantifiable rubrics; (3) diverse rubric structures without metric gaming, the data curation follows a rigorous three-stage expert annotation process to ensure objectivity and reliability in \cref{fig:examples1}. 

\paragraph{Stage 1: Task Creation.} 

An expert with professional domain knowledge begins by designing a specialized task, which includes: (1) formulating a specific and valuable task; (2) drafting a comprehensive reference answer that illustrates the underlying rationale; and (3) defining detailed scoring rubrics. These tasks are semi-open-ended, allowing different valid approaches, and are time-agnostic, yielding stable benchmarks over time. A critical subsequent step is concurrent adversarial validation. The task is tested against state-of-the-art agents from different model families to minimize bias, and retained only if several agents fail to meet a predefined passing threshold based on rubrics. 

\begin{figure*}[b]
    \centering
    \includegraphics[width=\columnwidth]{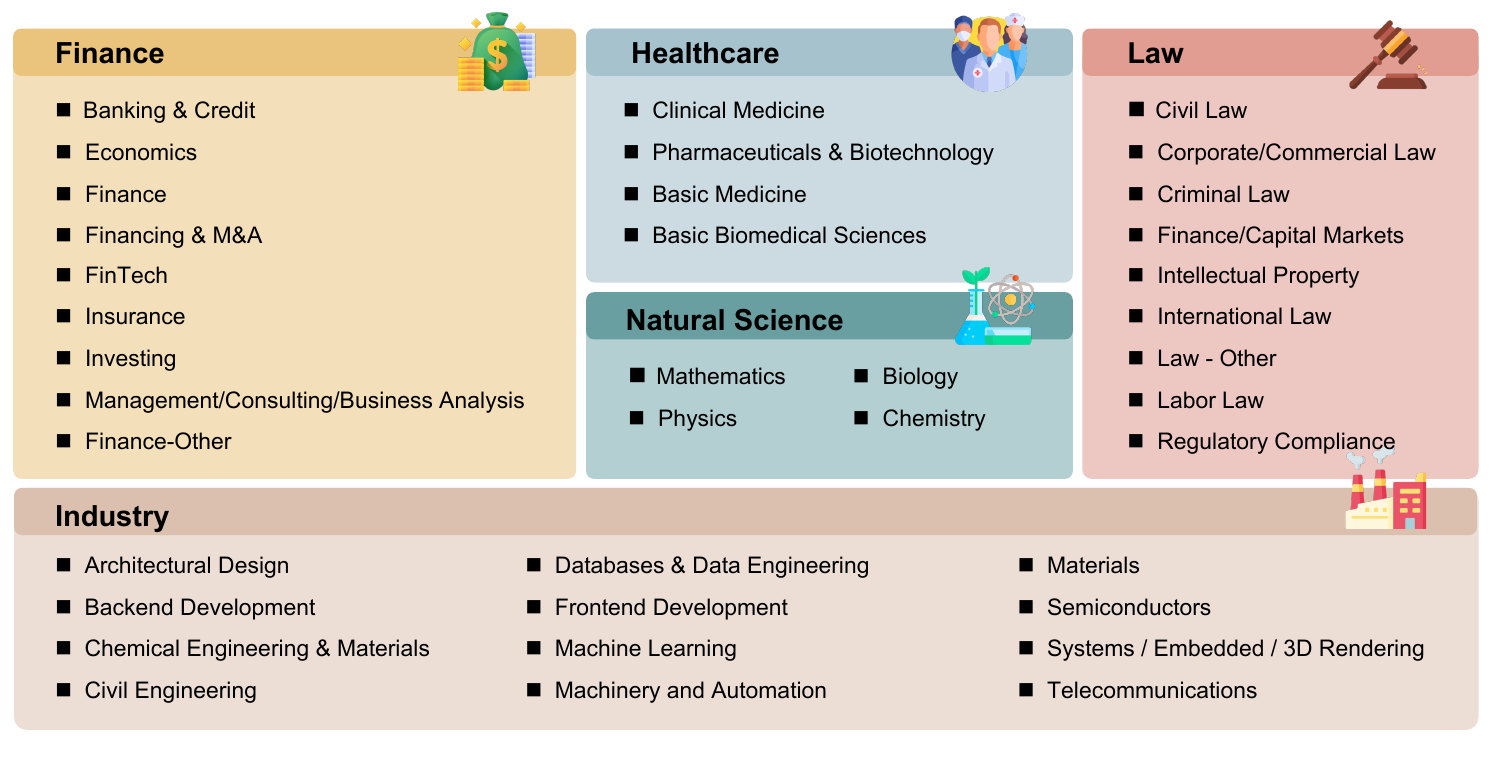} 
    \caption{\dataset consists of 5 macro domains, 37 sub-domains and 92 third-level categories covering a wide variety of real applications and professional scenarios  (\cf Appendix \ref{category}).}
    \label{fig:examples2}
\end{figure*}

\paragraph{Stage 2: Peer Review.}
After the initial draft is completed, the task undergoes an independent peer review by a second expert in the same sub-domain. The reviewer conducts a holistic evaluation of the task's clarity, domain specialization, adherence to design principles, and objectivity of the rubrics. Structured feedback leads to iterative discussions and collaborative revisions between the two experts, continuing until they reach a final agreement on the task, the reference response, and the evaluation criteria.

\paragraph{Stage 3: Resolution and Revision.} For cases exhibiting potential risk or instances where the first two experts can not achieve consensus, a third expert performs an independent audit. This final stage involves necessary adjustments to guarantee the high quality and reliability of the curated data.

To enhance the overall discriminability, we further filtered the candidate dataset with bidirectional truncation by removing or revising tasks at both extremes:
\begin{itemize}[leftmargin=*, topsep=0pt]
    \item \textbf{Lower Bound Elimination:} Tasks with excessively low difficulty are removed. They are solved by all the agents tested that are deemed a waste of evaluation resources without differentiation.
     \item \textbf{Upper Bound Review:} Tasks with extreme difficulty, where agents consistently yield universally low scores, undergo a secondary review to avoid unanswerable or infeasible tasks.
\end{itemize}


\begin{table*}[htbp]
\small
\centering
\caption{Data Statistics of \dataset \yang{More stats needed.}}
\label{tab:omb_stats}
\setlength{\tabcolsep}{8pt}
\renewcommand{\arraystretch}{1.05}
\begin{tabular*}{\textwidth}{c l c c c c}
\toprule
\textbf{Subset} & \textbf{Domain} & \textbf{\# Questions} & \textbf{Question Len.} & \textbf{\# Rubrics} & \textbf{\# Negative Rubrics} \\
\midrule
\multirow{5}{*}{\textbf{CN}}
& Finance         & 40 &  554.8 & 18.8 & 4.2 \\
& Law                & 40 &  709.1 & 16.3 & 3.9 \\
& Healthcare & 40 &  584.2 & 16.9 & 4.3 \\
& Natural Science   & 40 &  555.8 & 16.8 & 4.3 \\
& Industry           & 40 &  965.2 & 15.9 & 4.0 \\
\midrule
\multirow{5}{*}{\textbf{Global}}
& Finance        & 40 & 1810.9 & 18.5 & 4.1 \\
& Law                & 40 & 3277.8 & 16.0 & 4.2 \\
& Healthcare & 40 & 2029.9 & 16.9 & 4.2 \\
& Natural Science   & 40 & 1634.6 & 16.8 & 4.3 \\
& Industry           & 40 & 2581.1 & 15.9 & 4.0 \\
\bottomrule
\end{tabular*}

\vskip -0.1in
\end{table*}

\subsection{Data Overview}

\paragraph{Comprehensive Domain Coverage.}  \dataset comprises a meticulously curated collection of 400 questions categorized into 5 domains, including law, finance, healthcare, natural science and industry (\cref{tab:omb_stats} and \cref{fig:examples2}). 
These questions are tailored to the actual needs of different positions, workflows and constraints, which is critical for testing agents in a wide array of real-world scenarios, thereby facilitating a comprehensive and robust evaluation. Examples are depicted in \cref{fig:omb_examples} for reference.


\paragraph{Workflow-Realistic Difficulty and High Economic Value Problems.}
 The benchmark focuses on developing real-world agentic tasks across various vertical domains, ensuring that the task design closely matches the actual needs and workflows of the industry, thereby effectively evaluating the decision-making and task execution capabilities of agents in complex professional environments.

\dataset tasks are constructed to simulate professional workflows which not only require the agent to perform deep reasoning and real-time information integration, but also to clearly distinguish the practical capabilities and economic value of different agents through their judgments and insights.
Unlike the exam-style questions of existing benchmarks \cite{rein2024gpqa,phan2026benchmark}, many tasks in \dataset are intentionally structured so that a correct final answer is inseparable from a correct process: precise retrieval, traceable justification, and careful constraint satisfaction. 


\paragraph{Rubrics-based Grading with Negative Penalty.} 
Rubrics in \dataset are primarily determined case-by-case across domains and hand-crafted by domain experts. It ensures the rubrics are tailored and precise, moving beyond generic scoring to incorporate domain-specific logic, factual accuracy, and analytical depth. 
Each effective rubric evaluation consists of four fundamental elements:
\begin{itemize}[leftmargin=2.5em]
    \item \textbf{Rubrics:} Evaluation criteria, traits, and dimensions that determine response correctness.
    \item \textbf{Rubric Weights:} Gains or penalties reflecting the relative importance of each rubric.
    \item \textbf{Rubric Tags:} Categories classifying criteria by required capability.
    \item \textbf{Sources \& Citations:} References to verified external sources for factual accuracy.
\end{itemize}

It is worth highlighting that we propose a \textbf{negative rubric scoring} mechanism with asymmetric rubric weights ranging from -20 to 10 to better align with realistic domain requirements. The negative rubrics are mostly around behavioral violations of industrial norms or professional conduct, harmful generation, hallucination, and lapses in expected competencies such as instruction following.
 

\paragraph{Diverse Agentic Capabilities Assessment.} 
For each rubric, we associate a rubric tag indicating the agentic capability required to finish the task. 
\textbf{(1) Web Search:} This evaluates the agentic ability to locate and extract factual information, entity facts, and authoritative definitions from retrieved sources. It primarily assesses the information retrieval capability, focusing on questions of ``what''.
\textbf{(2) Reasoning:} This aspect examines how agents make causal attributions, coherent trend judgments, and logical deductions, addressing questions of ``why'' and ``how''. \textbf{(3) Verbalization:} This evaluates the organization, language style, and readability of the generation, ensuring clear, professional, and appropriate communication.
\textbf{(4) Instruction Following:} Compliance with instructions is critical for reliability in professional tasks. This assesses the adherence of agents to explicit constraints, rules, and guidance, ensuring controllable, predictable behaviors that meet user requirements.

\paragraph{Bilinguality Integrated with Local Culture.}
Although there are prior general-purpose agentic benchmarks (\eg, GDPVal \cite{patwardhan2026gdpval}), they operate under a default English or western-centric context, failing to capture significant variations in tasks, regulations, and practical scenarios encountered in different linguistic and cultural environments. Therefore, \benchmark consists of 200 English and 200 Chinese instances. The Chinese (CN) set, however, is not a direct translation in parallelism to the Global set, but a purpose-built collection that incorporates tasks specific to the context of the China Mainland. This includes local regulations and industrial standards (\eg, Chinese Cybersecurity Law, healthcare insurance policies, and accounting standards) and cultural nuances in problems. 


\definecolor{econcolor1}{RGB}{255,253,248} 
\definecolor{econcolor2}{RGB}{255,250,235} 
\definecolor{econcolor3}{RGB}{255,242,210} 
\definecolor{econcolor4}{RGB}{250,230,180} 

\definecolor{scoregray}{RGB}{240,240,240}

\definecolor{myposgreen}{RGB}{34,139,34}
\definecolor{mynegred}{RGB}{200,50,50}
\begin{table}[t]
\small
\centering
\caption{Comparison of Economic Value (\$USD), Expert Score (\%), and Pass Rate(\%) on \textbf{Global set}. The sign of \WebLogo denotes enabling the usage of web search tool, and the same holds for Table \ref{tab:cn_model_scores} and \ref{tab:en_cn_rubric}.}\label{tab:en_model_scores}
\setlength{\tabcolsep}{6pt}{
\begin{tabular}{l|cc|cc|cc}
\toprule
\multirowcell{2.6}{\textbf{Model / Agentic System}} & \multicolumn{2}{c|}{\textbf{Economic Value}} & \multicolumn{2}{c|}{\textbf{Expert Score}} & \multicolumn{2}{c}{\textbf{Pass Rate}} \\
\cmidrule(lr){2-3}\cmidrule(lr){4-5}\cmidrule(lr){6-7}
 & \textbf{Vanilla} & \textbf{w/ \WebLogo} & \textbf{Vanilla} & \textbf{w/ \WebLogo} & \textbf{Vanilla} & \textbf{w/ \WebLogo} \\
\midrule
\textsc{Claude-Opus-4.6} & \cellcolor{econcolor4} 439.2k & \cellcolor{econcolor4} 483.8k $_{\textcolor{myposgreen}{\uparrow 44.6}}$ &  55.0 &  63.0 $_{\textcolor{myposgreen}{\uparrow 8.1}}$ &  36.5 &  43.5 $_{\textcolor{myposgreen}{\uparrow 7.0}}$ \\
\textsc{GPT-5.4-High} & \cellcolor{econcolor4} 330.9k & \cellcolor{econcolor4} 365.5k $_{\textcolor{myposgreen}{\uparrow 34.6}}$ &  55.0 &  59.2 $_{\textcolor{myposgreen}{\uparrow 4.2}}$ &  31.5 &  38.0 $_{\textcolor{myposgreen}{\uparrow 6.5}}$ \\

\textsc{Qwen3.5-Plus} & \cellcolor{econcolor4} 256.0k & \cellcolor{econcolor4} 263.7k $_{\textcolor{myposgreen}{\uparrow 7.8}}$ &  47.6 &  49.0 $_{\textcolor{myposgreen}{\uparrow 1.4}}$ &  22.5 &  23.5 $_{\textcolor{myposgreen}{\uparrow 1.0}}$ \\
\textsc{GPT-5.2-High} & \cellcolor{econcolor4} 236.6k & \cellcolor{econcolor4} 335.9k $_{\textcolor{myposgreen}{\uparrow 99.3}}$ &  47.6 &  56.9 $_{\textcolor{myposgreen}{\uparrow 9.3}}$ &  23.0 &  32.0 $_{\textcolor{myposgreen}{\uparrow 9.0}}$ \\
\textsc{Kimi-K2.5} & \cellcolor{econcolor3} 203.3k & \cellcolor{econcolor3} 162.7k $_{\textcolor{mynegred}{\downarrow 40.6}}$ &  45.1 &  41.2 $_{\textcolor{mynegred}{\downarrow 3.9}}$ &  17.5 &  17.5 $_{\textcolor{myposgreen}{\uparrow 0.0}}$ \\
\textsc{GPT-5.3-Codex} & \cellcolor{econcolor3} 201.1k & \cellcolor{econcolor4} 317.8k $_{\textcolor{myposgreen}{\uparrow 116.7}}$ &  48.1 &  50.6 $_{\textcolor{myposgreen}{\uparrow 2.6}}$ &  18.5 &  29.0 $_{\textcolor{myposgreen}{\uparrow 10.5}}$ \\
\textsc{Gemini-3-Pro-Preview} & \cellcolor{econcolor3} 187.3k & \cellcolor{econcolor4} 345.3k $_{\textcolor{myposgreen}{\uparrow 158.0}}$ &  43.7 &  52.8 $_{\textcolor{myposgreen}{\uparrow 9.1}}$ &  15.5 &  28.5 $_{\textcolor{myposgreen}{\uparrow 13.0}}$ \\
\textsc{DeepSeek-V3.2-Speciale} & \cellcolor{econcolor3} 154.9k & \cellcolor{econcolor2} 118.0k $_{\textcolor{mynegred}{\downarrow 37.0}}$ &  40.5 &  38.6 $_{\textcolor{mynegred}{\downarrow 1.8}}$ &  16.0 &  13.0 $_{\textcolor{mynegred}{\downarrow 3.0}}$ \\
\textsc{Grok-4} & \cellcolor{econcolor3} 136.3k & \cellcolor{econcolor3} 204.6k $_{\textcolor{myposgreen}{\uparrow 68.3}}$ &  38.6 &  45.9 $_{\textcolor{myposgreen}{\uparrow 7.2}}$ &  12.0 &  18.5 $_{\textcolor{myposgreen}{\uparrow 6.5}}$ \\
\textsc{Doubao-Seed-2.0-Pro} & \cellcolor{econcolor2} 123.9k & \cellcolor{econcolor4} 330.7k $_{\textcolor{myposgreen}{\uparrow 206.9}}$ &  39.4 &  51.8 $_{\textcolor{myposgreen}{\uparrow 12.4}}$ &  13.5 &  28.5 $_{\textcolor{myposgreen}{\uparrow 15.0}}$ \\
\textsc{GLM-5} & \cellcolor{econcolor2} 107.8k & \cellcolor{econcolor3} 177.9k $_{\textcolor{myposgreen}{\uparrow 70.1}}$ &  38.5 &  41.1 $_{\textcolor{myposgreen}{\uparrow 2.6}}$ &  9.5 &  16.5 $_{\textcolor{myposgreen}{\uparrow 7.0}}$ \\
\textsc{MiniMax-M2.1} & \cellcolor{econcolor2} 85.6k & \cellcolor{econcolor1} 46.6k $_{\textcolor{mynegred}{\downarrow 39.0}}$ &  34.9 &  33.9 $_{\textcolor{mynegred}{\downarrow 1.0}}$ &  8.0 &  5.5 $_{\textcolor{mynegred}{\downarrow 2.5}}$ \\
\textsc{Ling-2.5-1T} & \cellcolor{econcolor2} 79.5k & \cellcolor{econcolor1} 50.0k $_{\textcolor{mynegred}{\downarrow 29.5}}$ &  33.1 &  31.2 $_{\textcolor{mynegred}{\downarrow 1.9}}$ &  7.5 &  6.5 $_{\textcolor{mynegred}{\downarrow 1.0}}$ \\
\textsc{Step-3.5-Flash} & \cellcolor{econcolor1} 68.6k & \cellcolor{econcolor1} 46.9k $_{\textcolor{mynegred}{\downarrow 21.7}}$ &  34.4 &  33.0 $_{\textcolor{mynegred}{\downarrow 1.4}}$ &  7.0 &  5.0 $_{\textcolor{mynegred}{\downarrow 2.0}}$ \\
\textsc{Hunyuan-2.0-Thinking} & \cellcolor{econcolor1} 68.0k & \cellcolor{econcolor1} 25.6k $_{\textcolor{mynegred}{\downarrow 42.3}}$ &  34.7 &  30.2 $_{\textcolor{mynegred}{\downarrow 4.5}}$ &  8.5 &  3.0 $_{\textcolor{mynegred}{\downarrow 5.5}}$ \\
\textsc{MiniMax-M2.5} & \cellcolor{econcolor1} 37.9k & \cellcolor{econcolor2} 82.6k $_{\textcolor{myposgreen}{\uparrow 44.8}}$ &  29.5 &  31.5 $_{\textcolor{myposgreen}{\uparrow 2.0}}$ &  4.0 &  8.0 $_{\textcolor{myposgreen}{\uparrow 4.0}}$ \\

\midrule
\multicolumn{7}{l}{\textbf{\textit{Deep Research Agents}}} \\
\midrule
\textsc{o3-DeepResearch} & \multicolumn{2}{c|}{\cellcolor{econcolor3} 181.7k} & \multicolumn{2}{c|}{ 46.3} & \multicolumn{2}{c}{ 17.0} \\
\textsc{Sonar-DeepResearch} & \multicolumn{2}{c|}{\cellcolor{econcolor2} 122.8k} & \multicolumn{2}{c|}{ 42.3} & \multicolumn{2}{c}{ 13.5} \\
\textsc{o4-Mini-DeepResearch} & \multicolumn{2}{c|}{\cellcolor{econcolor2} 112.6k} & \multicolumn{2}{c|}{ 37.7} & \multicolumn{2}{c}{ 10.5} \\
\bottomrule
\end{tabular}
}%
\end{table}

\section{Benchmarking Frontier Agents on \benchmark}

\subsection{Evaluated Systems}
In this section, we present the evaluation results for \dataset, encompassing a total of 37 models and agents, which are grouped into three categories according to their characteristics:
\begin{itemize}[leftmargin=*, topsep=1pt, noitemsep]
    \item \textbf{Vanilla Models (17 models):} The vanilla models represent open-source and proprietary models evaluated without tool usage, including 
    Claude-Opus-4.6 \cite{anthropic2026claudeopus46},
    DeepSeek-V3.2-Speciale \cite{liu2025deepseek},
    Doubao-Seed-2.0-Pro \cite{bytedance2026seed20},
    Gemini-3-Pro-Preview \cite{deepmind_gemini3pro_2025},
    Gemini-3.1-Pro-Preview\cite{geminiteam2026gemini31pro},
    GLM-5 \cite{Zai_GLM5_2026},
    GPT-5.2-High \cite{openai_gpt5.2_2025},
    GPT-5.3-Codex \cite{openai2026gpt53codex},
    GPT-5.4-High \cite{openai2026gpt54},
    Grok-4 \cite{xai_grok4_2025}, 
    Hunyuan-2.0-Thinking \cite{tencent2025hunyuan20},
    Kimi-K2.5 \cite{moonshot_kimi-k2-5_2026}, 
    Ling-2.5-1T \cite{inclusionai2026ling251t},
    MiniMax-M2.1 \cite{minimax_m2.1_github_2025}, 
    MiniMax-M2.5 \cite{minimax2026m25},
    Qwen3.5-Plus \cite{qwen2026qwen35}, and Step-3.5-Flash \cite{stepfun2026step35flash}.
    
    \item \textbf{Web Search Agents (vanilla models equipped with web search tools, 17 agents):} This category evaluates vanilla models listed above equipped with web search tool calling for external information retrieval, \eg, 
    Claude-Opus-4.6 (Web Search) \cite{anthropic2026claudeopus46}.
    
    \item \textbf{Deep Research Agents (3 agents):} These are specialized systems for long-horizon research, including 
    o3-DeepResearch \cite{openai_deep-research_2025}, 
    o4-Mini-DeepResearch \cite{openai_deep-research_2025}, and Sonar-DeepResearch \cite{perplexity2025sonardeepresearch}.
\end{itemize}

\subsection{Evaluation Metrics}

\paragraph{Expert Score.}
It is introduced to assess the degree to which model generations fulfill professional criteria across all questions. For a question set $Q$, the normalized Expert Score is defined as
\begin{equation}
\text{Expert Score}(Q) =
\frac{1}{|Q|}\sum_{q\in Q}
\max\!\left(0, \frac{\displaystyle\sum_{r \,\in\, R_q} s_r}{\displaystyle\sum_{r \,\in\, R_q^{+}} w_r}\right),
\end{equation}
where $R_q$ denotes the rubric set for question $q$, $R_q^+$ the subset with positive weights, $s_r$ the score of rubric $r$, and $w_r$ its predefined weight. Question scores are clipped to $[0,1]$ before aggregation.

\paragraph{Pass Rate.}
We also define Pass Rate, a binary success metric indicating whether a generation meets a minimum professional standard via calculating the average Expert Score across all questions $Q$:
\begin{equation}
\text{Pass Rate}(Q) = \frac{1}{|Q|}\sum_{q\in Q}\mathds{1}\left[\text{Expert Score}(q)\ge 0.7\right] ,
\end{equation}
where $\mathds{1}[\cdot]$ denotes the indicator function based on the given binary scoring criterion.

\paragraph{Economic Value.}
The economic value of an agent is the cumulative value of all passed questions:
\begin{equation}
\text{Economic Value}(Q) =
\sum_{q \in Q} V_q \cdot \mathds{1}\left[\text{Expert Score}(q) \ge 0.7\right],
\end{equation}
where $V_q$ denotes the estimated economic value of question $q$. \textbf{Specifically, we provide a detailed approach and guideline of reproducibility for economic value calculation in Appendix~\ref{app:eco_value_calculation}.} We follow \citet{dou2026cl} to use 0.7 as the cliff threshold in our default setting.

\definecolor{econcolor1}{RGB}{255,253,248} 
\definecolor{econcolor2}{RGB}{255,250,235} 
\definecolor{econcolor3}{RGB}{255,242,210} 
\definecolor{econcolor4}{RGB}{250,230,180} 

\definecolor{scoregray}{RGB}{240,240,240}

\definecolor{myposgreen}{RGB}{34,139,34}
\definecolor{mynegred}{RGB}{200,50,50}
\begin{table}[t]
\small
\centering
\caption{Comparison of Economic Value (\textyen CNY), Expert Score (\%), and Pass Rate (\%) on \textbf{CN set}.}\label{tab:cn_model_scores}
\setlength{\tabcolsep}{6pt}{
\begin{tabular}{l|cc|cc|cc}
\toprule
\multirowcell{2.6}{\textbf{Model / Agentic System}} & \multicolumn{2}{c|}{\textbf{Economic Value}} & \multicolumn{2}{c|}{\textbf{Expert Score}} & \multicolumn{2}{c}{\textbf{Pass Rate}} \\
\cmidrule(lr){2-3}\cmidrule(lr){4-5}\cmidrule(lr){6-7}
 & \textbf{Vanilla} & \textbf{w/ \WebLogo} & \textbf{Vanilla} & \textbf{w/ \WebLogo} & \textbf{Vanilla} & \textbf{w/ \WebLogo} \\
\midrule
\textsc{Claude-Opus-4.6} & \cellcolor{econcolor4} 350.3k & \cellcolor{econcolor4} 470.2k $_{\textcolor{myposgreen}{\uparrow 119.9}}$ &  55.8 &  64.5 $_{\textcolor{myposgreen}{\uparrow 8.7}}$ &  35.0 &  48.5 $_{\textcolor{myposgreen}{\uparrow 13.5}}$ \\
\textsc{GPT-5.4-High} & \cellcolor{econcolor4} 339.5k & \cellcolor{econcolor4} 305.4k $_{\textcolor{mynegred}{\downarrow 34.1}}$ &  56.2 &  58.7 $_{\textcolor{myposgreen}{\uparrow 2.5}}$ &  35.5 &  35.5 $_{\textcolor{myposgreen}{\uparrow 0.0}}$ \\
\textsc{Qwen3.5-Plus} & \cellcolor{econcolor4} 304.9k & \cellcolor{econcolor4} 308.1k $_{\textcolor{myposgreen}{\uparrow 3.2}}$ &  51.4 &  52.1 $_{\textcolor{myposgreen}{\uparrow 0.6}}$ &  28.0 &  28.0 $_{\textcolor{myposgreen}{\uparrow 0.0}}$ \\
\textsc{GPT-5.2-High} & \cellcolor{econcolor4} 226.9k & \cellcolor{econcolor4} 294.0k $_{\textcolor{myposgreen}{\uparrow 67.1}}$ &  47.1 &  52.6 $_{\textcolor{myposgreen}{\uparrow 5.5}}$ &  22.0 &  28.0 $_{\textcolor{myposgreen}{\uparrow 6.0}}$ \\
\textsc{Gemini-3-Pro-Preview} & \cellcolor{econcolor3} 202.9k & \cellcolor{econcolor4} 332.5k $_{\textcolor{myposgreen}{\uparrow 129.7}}$ &  46.0 &  54.0 $_{\textcolor{myposgreen}{\uparrow 8.0}}$ &  20.0 &  30.0 $_{\textcolor{myposgreen}{\uparrow 10.0}}$ \\
\textsc{Kimi-K2.5} & \cellcolor{econcolor3} 196.5k & \cellcolor{econcolor2} 126.0k $_{\textcolor{mynegred}{\downarrow 70.4}}$ &  45.4 &  43.7 $_{\textcolor{mynegred}{\downarrow 1.7}}$ &  19.5 &  15.0 $_{\textcolor{mynegred}{\downarrow 4.5}}$ \\
\textsc{GPT-5.3-Codex} & \cellcolor{econcolor3} 178.4k & \cellcolor{econcolor3} 201.1k $_{\textcolor{myposgreen}{\uparrow 22.7}}$ &  46.9 &  50.9 $_{\textcolor{myposgreen}{\uparrow 4.1}}$ &  19.0 &  23.0 $_{\textcolor{myposgreen}{\uparrow 4.0}}$ \\
\textsc{DeepSeek-V3.2-Speciale} & \cellcolor{econcolor3} 139.4k & \cellcolor{econcolor3} 155.8k $_{\textcolor{myposgreen}{\uparrow 16.4}}$ &  41.1 &  44.4 $_{\textcolor{myposgreen}{\uparrow 3.3}}$ &  14.0 &  16.5 $_{\textcolor{myposgreen}{\uparrow 2.5}}$ \\
\textsc{GLM-5} & \cellcolor{econcolor3} 138.9k & \cellcolor{econcolor3} 136.3k $_{\textcolor{mynegred}{\downarrow 2.6}}$ &  40.7 &  41.1 $_{\textcolor{myposgreen}{\uparrow 0.4}}$ &  13.0 &  11.0 $_{\textcolor{mynegred}{\downarrow 2.0}}$ \\
\textsc{Ling-2.5-1T} & \cellcolor{econcolor2} 119.8k & \cellcolor{econcolor2} 101.8k $_{\textcolor{mynegred}{\downarrow 17.9}}$ &  36.7 &  35.6 $_{\textcolor{mynegred}{\downarrow 1.1}}$ &  11.0 &  11.5 $_{\textcolor{myposgreen}{\uparrow 0.5}}$ \\
\textsc{Doubao-Seed-2.0-Pro} & \cellcolor{econcolor2} 110.8k & \cellcolor{econcolor4} 255.9k $_{\textcolor{myposgreen}{\uparrow 145.1}}$ &  40.6 &  50.5 $_{\textcolor{myposgreen}{\uparrow 9.8}}$ &  11.5 &  25.0 $_{\textcolor{myposgreen}{\uparrow 13.5}}$ \\
\textsc{MiniMax-M2.1} & \cellcolor{econcolor2} 106.6k & \cellcolor{econcolor1} 67.2k $_{\textcolor{mynegred}{\downarrow 39.4}}$ &  35.8 &  36.7 $_{\textcolor{myposgreen}{\uparrow 1.0}}$ &  10.0 &  6.0 $_{\textcolor{mynegred}{\downarrow 4.0}}$ \\
\textsc{Grok-4} & \cellcolor{econcolor2} 104.2k & \cellcolor{econcolor2} 99.4k $_{\textcolor{mynegred}{\downarrow 4.8}}$ &  39.8 &  39.6 $_{\textcolor{mynegred}{\downarrow 0.2}}$ &  11.0 &  11.5 $_{\textcolor{myposgreen}{\uparrow 0.5}}$ \\
\textsc{Hunyuan-2.0-Thinking} & \cellcolor{econcolor1} 83.0k & \cellcolor{econcolor1} 53.8k $_{\textcolor{mynegred}{\downarrow 29.2}}$ &  36.3 &  32.8 $_{\textcolor{mynegred}{\downarrow 3.4}}$ &  9.5 &  6.5 $_{\textcolor{mynegred}{\downarrow 3.0}}$ \\
\textsc{MiniMax-M2.5} & \cellcolor{econcolor1} 70.3k & \cellcolor{econcolor1} 51.7k $_{\textcolor{mynegred}{\downarrow 18.7}}$ &  33.4 &  32.1 $_{\textcolor{mynegred}{\downarrow 1.3}}$ &  7.5 &  7.5 $_{\textcolor{myposgreen}{\uparrow 0.0}}$ \\
\textsc{Step-3.5-Flash} & \cellcolor{econcolor1} 52.6k & \cellcolor{econcolor1} 68.7k $_{\textcolor{myposgreen}{\uparrow 16.2}}$ &  36.2 &  35.6 $_{\textcolor{mynegred}{\downarrow 0.6}}$ &  6.5 &  6.5 $_{\textcolor{myposgreen}{\uparrow 0.0}}$ \\

\midrule
\multicolumn{7}{l}{\textbf{\textit{Deep Research Agents}}} \\
\midrule
\textsc{o3-DeepResearch} & \multicolumn{2}{c|}{\cellcolor{econcolor3} 149.2k} & \multicolumn{2}{c|}{ 41.7} & \multicolumn{2}{c}{ 16.0} \\
\textsc{Sonar-DeepResearch} & \multicolumn{2}{c|}{\cellcolor{econcolor2} 129.5k} & \multicolumn{2}{c|}{ 40.2} & \multicolumn{2}{c}{ 13.5} \\
\textsc{o4-Mini-DeepResearch} & \multicolumn{2}{c|}{\cellcolor{econcolor2} 84.7k} & \multicolumn{2}{c|}{ 36.0} & \multicolumn{2}{c}{ 9.0} \\
\bottomrule
\end{tabular}
}%
\end{table}

\definecolor{scoregreen}{RGB}{70,130,200}
\definecolor{scorepeach}{RGB}{120,180,240}
\definecolor{scorepink}{RGB}{200,230,255}
\definecolor{scorered}{RGB}{240,248,255}
\definecolor{scoregray}{RGB}{240,240,240}
\definecolor{txtongreen}{RGB}{255,255,255}
\definecolor{txtonpeach}{RGB}{50,50,50}
\definecolor{txtonpink}{RGB}{50,50,50}
\definecolor{txtonred}{RGB}{50,50,50}
\definecolor{txtongray}{RGB}{50,50,50}
\definecolor{myposgreen}{RGB}{34,139,34}
\definecolor{mynegred}{RGB}{200,50,50}
\begin{table}[t]
\centering
\caption{Expert Score aggregated by rubric type, averaged over Global and CN subsets. FI: Factual Information. AR: Analytical Reasoning. IF: Instructions Following. SF: Structure and Formatting.}
\label{tab:en_cn_rubric}
\resizebox{\textwidth}{!}{%
\begin{tabular}{l|cc|cc|cc|cc}
\toprule
\multirow{2.6}{*}{\textbf{Model / Agentic System}} & \multicolumn{2}{c|}{\textbf{FI}} & \multicolumn{2}{c|}{\textbf{AR}} & \multicolumn{2}{c|}{\textbf{IF}} & \multicolumn{2}{c}{\textbf{SF}} \\
\cmidrule(lr){2-3}\cmidrule(lr){4-5}\cmidrule(lr){6-7}\cmidrule(lr){8-9}
 & \textbf{Vanilla} & \textbf{w/ \WebLogo} & \textbf{Vanilla} & \textbf{w/ \WebLogo} & \textbf{Vanilla} & \textbf{w/ \WebLogo} & \textbf{Vanilla} & \textbf{w/ \WebLogo} \\
\midrule
\textsc{GPT-5.2-High} &  \textcolor{black}{47.4} &  \textcolor{black}{59.4}\,\,$_{\textcolor{myposgreen}{\uparrow 12.0}}$ &  \textcolor{black}{52.3} &  \textcolor{black}{56.4}\,\,$_{\textcolor{myposgreen}{\uparrow 4.1}}$ &  \textcolor{black}{66.0} &  \textcolor{black}{72.1}\,\,$_{\textcolor{myposgreen}{\uparrow 6.1}}$ &  \textcolor{black}{82.7} &  \textcolor{black}{85.6}\,\,$_{\textcolor{myposgreen}{\uparrow 2.9}}$ \\
\textsc{Qwen3.5-Plus} &  \textcolor{black}{50.3} &  \textcolor{black}{50.0}\,\,$_{\textcolor{mynegred}{\downarrow 0.3}}$ &  \textcolor{black}{58.9} &  \textcolor{black}{59.9}\,\,$_{\textcolor{myposgreen}{\uparrow 1.0}}$ &  \textcolor{black}{73.2} &  \textcolor{black}{74.6}\,\,$_{\textcolor{myposgreen}{\uparrow 1.4}}$ &  \textcolor{black}{86.4} &  \textcolor{black}{88.8}\,\,$_{\textcolor{myposgreen}{\uparrow 2.4}}$ \\
\textsc{Claude-Opus-4.6} &  \textcolor{black}{54.0} &  \textcolor{black}{66.5}\,\,$_{\textcolor{myposgreen}{\uparrow 12.5}}$ &  \textcolor{black}{61.3} &  \textcolor{black}{73.0}\,\,$_{\textcolor{myposgreen}{\uparrow 11.7}}$ &  \textcolor{black}{74.6} &  \textcolor{black}{82.1}\,\,$_{\textcolor{myposgreen}{\uparrow 7.5}}$ &  \textcolor{black}{88.9} &  \textcolor{black}{90.0}\,\,$_{\textcolor{myposgreen}{\uparrow 1.1}}$ \\
\textsc{GPT-5.4-High} &  \textcolor{black}{53.5} &  \textcolor{black}{63.5}\,\,$_{\textcolor{myposgreen}{\uparrow 10.0}}$ &  \textcolor{black}{60.8} &  \textcolor{black}{61.6}\,\,$_{\textcolor{myposgreen}{\uparrow 0.8}}$ &  \textcolor{black}{71.5} &  \textcolor{black}{73.9}\,\,$_{\textcolor{myposgreen}{\uparrow 2.4}}$ &  \textcolor{black}{86.4} &  \textcolor{black}{84.5}\,\,$_{\textcolor{mynegred}{\downarrow 1.9}}$ \\

\textsc{Gemini-3.1-Pro-Preview} &  \textcolor{black}{49.8} &  \textcolor{black}{46.7}\,\,$_{\textcolor{mynegred}{\downarrow 3.1}}$ &  \textcolor{black}{51.9} &  \textcolor{black}{50.8}\,\,$_{\textcolor{mynegred}{\downarrow 1.1}}$ &  \textcolor{black}{70.6} &  \textcolor{black}{67.9}\,\,$_{\textcolor{mynegred}{\downarrow 2.7}}$ &  \textcolor{black}{82.7} &  \textcolor{black}{82.7}\,\,$_{\textcolor{myposgreen}{\uparrow 0.0}}$ \\
\textsc{GPT-5.3-Codex} &  \textcolor{black}{48.5} &  \textcolor{black}{58.4}\,\,$_{\textcolor{myposgreen}{\uparrow 9.9}}$ &  \textcolor{black}{51.1} &  \textcolor{black}{52.0}\,\,$_{\textcolor{myposgreen}{\uparrow 0.9}}$ &  \textcolor{black}{67.5} &  \textcolor{black}{65.9}\,\,$_{\textcolor{mynegred}{\downarrow 1.6}}$ &  \textcolor{black}{85.0} &  \textcolor{black}{83.6}\,\,$_{\textcolor{mynegred}{\downarrow 1.4}}$ \\
\textsc{Kimi-K2.5} &  \textcolor{black}{46.5} &  \textcolor{black}{44.1}\,\,$_{\textcolor{mynegred}{\downarrow 2.4}}$ &  \textcolor{black}{51.1} &  \textcolor{black}{48.1}\,\,$_{\textcolor{mynegred}{\downarrow 3.0}}$ &  \textcolor{black}{70.1} &  \textcolor{black}{63.4}\,\,$_{\textcolor{mynegred}{\downarrow 6.7}}$ &  \textcolor{black}{85.0} &  \textcolor{black}{80.6}\,\,$_{\textcolor{mynegred}{\downarrow 4.4}}$ \\
\textsc{Gemini-3-Pro-Preview} &  \textcolor{black}{45.8} &  \textcolor{black}{56.7}\,\,$_{\textcolor{myposgreen}{\uparrow 10.9}}$ &  \textcolor{black}{49.2} &  \textcolor{black}{60.7}\,\,$_{\textcolor{myposgreen}{\uparrow 11.5}}$ &  \textcolor{black}{66.9} &  \textcolor{black}{73.1}\,\,$_{\textcolor{myposgreen}{\uparrow 6.2}}$ &  \textcolor{black}{82.8} &  \textcolor{black}{86.8}\,\,$_{\textcolor{myposgreen}{\uparrow 4.0}}$ \\
\textsc{Grok-4} &  \textcolor{black}{41.6} &  \textcolor{black}{48.5}\,\,$_{\textcolor{myposgreen}{\uparrow 6.9}}$ &  \textcolor{black}{44.7} &  \textcolor{black}{46.7}\,\,$_{\textcolor{myposgreen}{\uparrow 2.0}}$ &  \textcolor{black}{65.2} &  \textcolor{black}{62.7}\,\,$_{\textcolor{mynegred}{\downarrow 2.5}}$ &  \textcolor{black}{83.0} &  \textcolor{black}{86.5}\,\,$_{\textcolor{myposgreen}{\uparrow 3.5}}$ \\
\textsc{Doubao-Seed-2.0-Pro} &  \textcolor{black}{42.3} &  \textcolor{black}{53.6}\,\,$_{\textcolor{myposgreen}{\uparrow 11.3}}$ &  \textcolor{black}{45.3} &  \textcolor{black}{58.8}\,\,$_{\textcolor{myposgreen}{\uparrow 13.5}}$ &  \textcolor{black}{62.3} &  \textcolor{black}{72.7}\,\,$_{\textcolor{myposgreen}{\uparrow 10.4}}$ &  \textcolor{black}{83.6} &  \textcolor{black}{86.8}\,\,$_{\textcolor{myposgreen}{\uparrow 3.2}}$ \\
\textsc{DeepSeek-V3.2-Speciale} &  \textcolor{black}{43.7} &  \textcolor{black}{43.9}\,\,$_{\textcolor{myposgreen}{\uparrow 0.2}}$ &  \textcolor{black}{45.4} &  \textcolor{black}{46.8}\,\,$_{\textcolor{myposgreen}{\uparrow 1.4}}$ &  \textcolor{black}{62.8} &  \textcolor{black}{60.2}\,\,$_{\textcolor{mynegred}{\downarrow 2.6}}$ &  \textcolor{black}{79.4} &  \textcolor{black}{78.8}\,\,$_{\textcolor{mynegred}{\downarrow 0.6}}$ \\
\textsc{GLM-5} &  \textcolor{black}{41.8} &  \textcolor{black}{40.8}\,\,$_{\textcolor{mynegred}{\downarrow 1.0}}$ &  \textcolor{black}{45.1} &  \textcolor{black}{47.7}\,\,$_{\textcolor{myposgreen}{\uparrow 2.6}}$ &  \textcolor{black}{65.3} &  \textcolor{black}{65.7}\,\,$_{\textcolor{myposgreen}{\uparrow 0.4}}$ &  \textcolor{black}{79.4} &  \textcolor{black}{82.7}\,\,$_{\textcolor{myposgreen}{\uparrow 3.3}}$ \\
\textsc{Hunyuan-2.0-Thinking} &  \textcolor{black}{39.1} &  \textcolor{black}{36.3}\,\,$_{\textcolor{mynegred}{\downarrow 2.8}}$ &  \textcolor{black}{40.2} &  \textcolor{black}{36.5}\,\,$_{\textcolor{mynegred}{\downarrow 3.7}}$ &  \textcolor{black}{63.5} &  \textcolor{black}{60.9}\,\,$_{\textcolor{mynegred}{\downarrow 2.6}}$ &  \textcolor{black}{85.0} &  \textcolor{black}{79.8}\,\,$_{\textcolor{mynegred}{\downarrow 5.2}}$ \\
\textsc{MiniMax-M2.1} &  \textcolor{black}{39.3} &  \textcolor{black}{38.3}\,\,$_{\textcolor{mynegred}{\downarrow 1.0}}$ &  \textcolor{black}{41.6} &  \textcolor{black}{42.4}\,\,$_{\textcolor{myposgreen}{\uparrow 0.8}}$ &  \textcolor{black}{60.4} &  \textcolor{black}{57.4}\,\,$_{\textcolor{mynegred}{\downarrow 3.0}}$ &  \textcolor{black}{82.6} &  \textcolor{black}{81.8}\,\,$_{\textcolor{mynegred}{\downarrow 0.8}}$ \\
\textsc{Step-3.5-Flash} &  \textcolor{black}{36.2} &  \textcolor{black}{38.4}\,\,$_{\textcolor{myposgreen}{\uparrow 2.2}}$ &  \textcolor{black}{42.5} &  \textcolor{black}{41.6}\,\,$_{\textcolor{mynegred}{\downarrow 0.9}}$ &  \textcolor{black}{62.3} &  \textcolor{black}{56.1}\,\,$_{\textcolor{mynegred}{\downarrow 6.2}}$ &  \textcolor{black}{82.6} &  \textcolor{black}{79.2}\,\,$_{\textcolor{mynegred}{\downarrow 3.4}}$ \\
\textsc{Ling-2.5-1T} &  \textcolor{black}{37.6} &  \textcolor{black}{36.5}\,\,$_{\textcolor{mynegred}{\downarrow 1.1}}$ &  \textcolor{black}{41.6} &  \textcolor{black}{38.7}\,\,$_{\textcolor{mynegred}{\downarrow 2.9}}$ &  \textcolor{black}{59.3} &  \textcolor{black}{55.9}\,\,$_{\textcolor{mynegred}{\downarrow 3.4}}$ &  \textcolor{black}{79.3} &  \textcolor{black}{74.4}\,\,$_{\textcolor{mynegred}{\downarrow 4.9}}$ \\
\textsc{MiniMax-M2.5} &  \textcolor{black}{34.9} &  \textcolor{black}{35.3}\,\,$_{\textcolor{myposgreen}{\uparrow 0.4}}$ &  \textcolor{black}{39.1} &  \textcolor{black}{35.8}\,\,$_{\textcolor{mynegred}{\downarrow 3.3}}$ &  \textcolor{black}{58.2} &  \textcolor{black}{55.4}\,\,$_{\textcolor{mynegred}{\downarrow 2.8}}$ &  \textcolor{black}{83.0} &  \textcolor{black}{71.5}\,\,$_{\textcolor{mynegred}{\downarrow 11.5}}$ \\
\midrule
\multicolumn{9}{l}{\textbf{\textit{Deep Research Agents}}} \\
\midrule
\textsc{Sonar-DeepResearch} & \multicolumn{2}{c|}{ \textcolor{black}{47.2}} & \multicolumn{2}{c|}{ \textcolor{black}{49.9}} & \multicolumn{2}{c|}{ \textcolor{black}{60.7}} & \multicolumn{2}{c}{ \textcolor{black}{80.7}} \\
\textsc{o3-DeepResearch} & \multicolumn{2}{c|}{ \textcolor{black}{47.9}} & \multicolumn{2}{c|}{ \textcolor{black}{50.2}} & \multicolumn{2}{c|}{ \textcolor{black}{59.2}} & \multicolumn{2}{c}{ \textcolor{black}{76.5}} \\
\textsc{o4-Mini-DeepResearch} & \multicolumn{2}{c|}{ \textcolor{black}{42.9}} & \multicolumn{2}{c|}{ \textcolor{black}{42.3}} & \multicolumn{2}{c|}{ \textcolor{black}{58.1}} & \multicolumn{2}{c}{ \textcolor{black}{71.7}} \\
\bottomrule
\end{tabular}
}%
\end{table}

\subsection{Main Results}
Tables \ref{tab:en_model_scores}, \ref{tab:cn_model_scores}, \ref{tab:en_domains}, and \ref{tab:cn_domains} present the comparison for agents evaluated in the Global and CN subsets, respectively. From these empirical results, several key observations emerge and are listed below.

\paragraph{The Strong Get Stronger: Web Search Rewards Capable Models.} 
In both Global and CN subsets, \textsc{Claude-Opus-4.6} leads among vanilla models and further extends its advantage when web search is enabled, suggesting that it mainly benefits models capable of converting retrieved evidence into reliable decisions. Stronger models excel at forming precise queries, identifying authoritative sources, resolving conflicts, and integrating the retrieved facts with task constraints. Weaker models tend to treat search results as raw context and are more easily misled by noisy evidence. For example, \textsc{Hunyuan-2.0} and \textsc{Step-3.5-Flash} shows similar declines after enabling web search, indicating that retrieval can hurt performance when models lack the ability to select and reconcile evidence.

\paragraph{Partial Credit, Low Delivery: The Gap Between Scoring and Solving.} 
Although many agents reach moderate Expert Scores (\eg, 45\%--50\%), their Pass Rates often remain low (below 25\%), indicating partial rubric satisfaction rather than full task competency. For instance, in Finance, an agent may extract relevant figures and give a plausible judgment, yet fail by missing a required calculation, using an outdated threshold, or overlooking a key exception. This also explains why Finance remains harder than others for top models: it requires exact numerical reasoning and strict constraint checking, where incomplete solutions receive limited credit, suggesting that difficulty reflects both rubric fullfillness demanding strong domain knowledge.

\paragraph{Global \vs CN Differences Reveal Region-Specific Generalization Gaps.}
Global and CN results expose distinct capability bottlenecks across agents. Top-tier agents remain relatively stable: \textsc{Claude-Opus-4.6} obtains similar vanilla Expert Scores on Global and CN (55.0\% \vs 55.8\%), as does \textsc{GPT-5.4-High} (55.0\% \vs 56.2\%). This suggests that their performance relies more on general evidence understanding and rubric-aligned execution than on a specific regional distribution. By contrast, \textsc{Qwen3.5-Plus} performs better on CN than Global (51.4\% \vs 47.6\%), likely due to stronger coverage of Chinese knowledge and evidence. Overall, cross-region robustness depends not only on agent strength, but also on adapting reasoning to different evidence contexts.


\section{Analyses}
 
\subsection{Performance Analyses across Rubric Type}
To evaluate task performance in detail, we report the per-type scores in Table \ref{tab:en_cn_rubric} and observe that:

A pronounced gap exists between superficial and evidence-centric capabilities: agents consistently outperform in \textit{Structure and Formatting} (SF) and \textit{Instruction Following} (IF) compared to \textit{Factual Information} (FI) and \textit{Analytical Reasoning} (AR), indicating agents can produce well-organized, instruction-compliant solutions while failing to deliver accurate evidence or complete analysis. Web search primarily enhances evidence-centric dimensions for frontier models, notably \textsc{Claude-Opus-4.6} (+12.5\% FI, +11.7\% AR) and \textsc{GPT-5.2-High} (+12.0\% FI, +4.1\% AR), while potentially misleading weaker models through noisy retrieval and hallucination amplification.

Web search also tests whether agents stay anchored to the original task, making IF highly dependent on specific models ranging from +10.4\% for \textsc{Doubao-Seed-2.0-Pro} to -6.7\% for \textsc{Kimi-K2.5}, as some agents shift toward summarizing retrieved content. SF shows little room for improvement and may even weaken as search encourages longer and evidence-heavy responses.



\subsection{Web Search Scaffold Does Matter}
\label{subsec:web_scaffolding_impact}
\begin{wrapfigure}{r}{0.5\textwidth}
    \centering
    \includegraphics[width=1\linewidth]{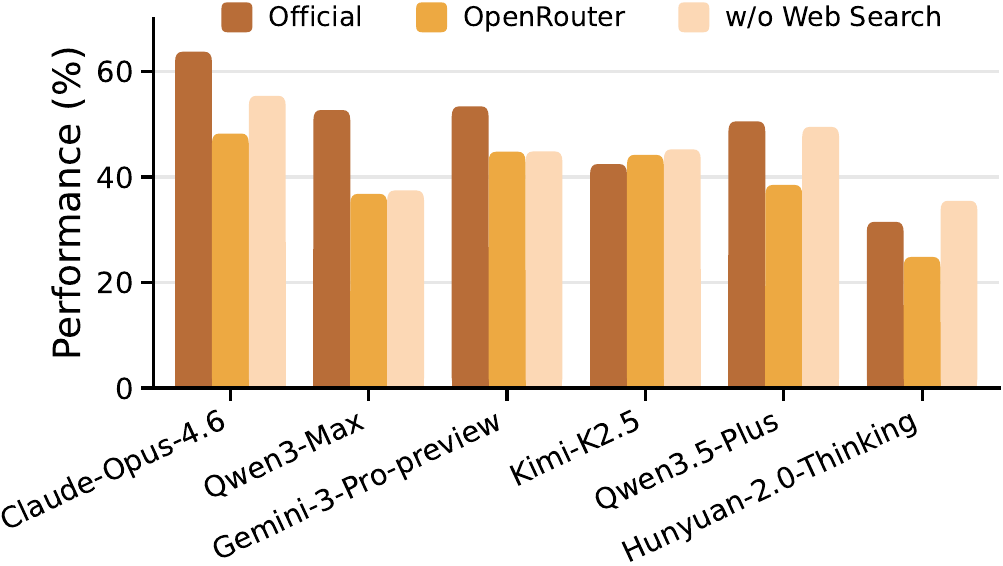}
    \captionsetup{width=1\linewidth}
    \caption{Comparison of web search scaffolds.}
    \label{fig:scaffold_comparsion}
\end{wrapfigure}
To mitigate non-transferability issues in the evaluation and provide insights into the impact of agent scaffold, we investigate three setups: (1) scaffold of official provider (\eg, OpenAI and Anthropic), (2) OpenRouter\footnote{\url{https://openrouter.ai}}, and (3) no scaffold (w/o Web Search). In Figure~\ref{fig:scaffold_comparsion}, official scaffold outperforms OpenRouter in almost all evaluated models. The performance gap between official and OpenRouter scaffold varies considerably across models.
In some cases (\eg, Claude-Opus-4.6), the ``w/o Web Search'' baseline surprisingly surpasses the OpenRouter setup, suggesting that scaffold quality matters more than access to web search alone. 

\subsection{Agents Struggle for Consistent Delivery}
\label{subsec:consistent_delivery}
\begin{wrapfigure}{r}{0.49\textwidth}
    \vspace{-11mm}
    \centering
    \includegraphics[width=1\linewidth]{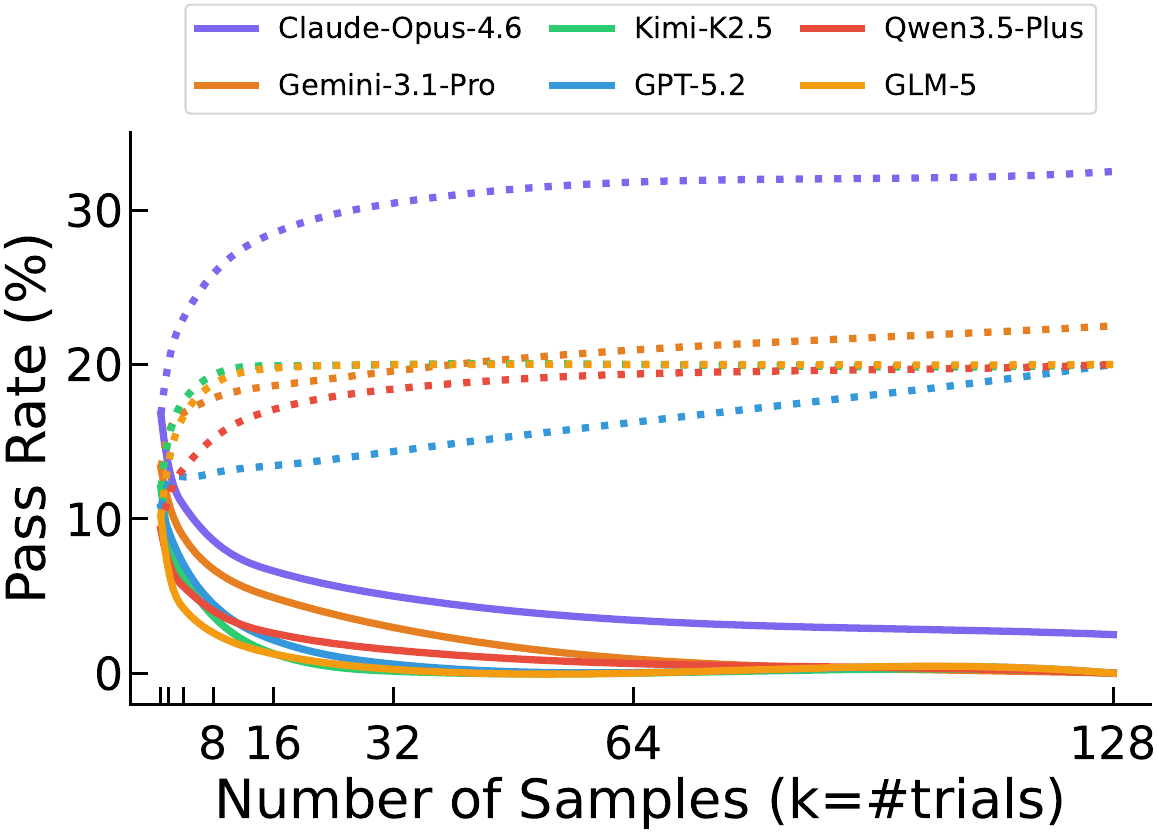}
    \captionsetup{width=1\linewidth}
    \caption{Test-time scalibility on Finance of the Global subset for consistency validation.}
    \label{fig:test_time_scalibility}
\end{wrapfigure}

We report \passatk and \passofk scores with performing parallel test-time scaling experiments on the Finance domain of the Global subset \cite{snell2025scaling}. Overall, \cref {fig:test_time_scalibility} illustrates the test-time scalability of various models on \benchmark. As k increases, the two metrics diverge sharply: \passatk (dotted lines) shows consistent logarithmic gains across all models, with Claude-Opus-4.6 leading and plateauing near $30\%$, while \passofk (solid lines) decays towards zero. While larger sample scales increase the probability of capturing a correct solution, the accompanying rise in output variance degrades reliability, causing consistency to collapse in large-$k$ regimes. 







\subsection{Pareto Optimality of Return on Cost}
\label{subsec:pareto_frontier}
\begin{figure*}[htbp]
    \centering
    \includegraphics[width=\columnwidth]{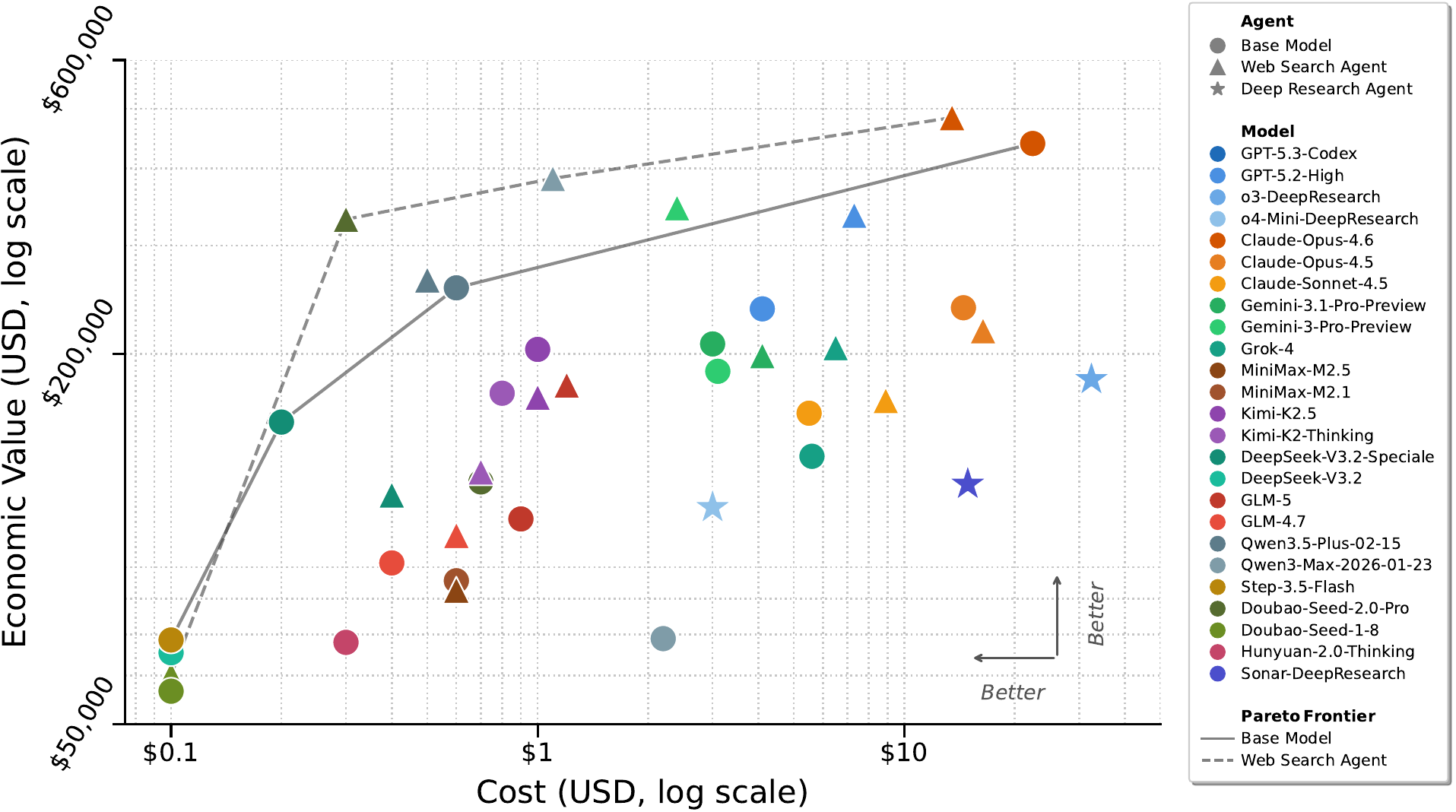} 
    \caption{Pareto frontiers of various agents and models on \benchmark.}
    \label{fig:pareto_frontier}
\end{figure*}

As depicted in \cref{fig:pareto_frontier}, agents or models achieve a pareto-optimal trade-off tendency between the outcome economic value versus the inference cost of agents or models. Specialized web search agents deliver drastically higher economic value than that of the same base model. 
We also notice that smaller models deliver near-frontier deep research performance at a fraction of the cost of the largest top-tier models, making advanced research tools accessible to price-sensitive users in the real scenarios. Based on the further case studies in \cref{subsec:deficiency_taxonomy}, we find that higher token cost does not necessarily contribute to more consistent delivery in those domain specific tasks. 

\subsection{Deficiency Analysis}\label{subsec:deficiency_taxonomy}

We randomly choose 100 trajectories of solutions from different domains for three representative models (GPT-5.2-High, Gemini-3-Pro-preview, and Claude-Opus-4.5). Our qualitative analysis reveals several recurring failure modes in \benchmark as follows: 

\paragraph{Flawed Reasoning with Weak Verification (34\% of Errors).} The model may arrive at a plausible conclusion while failing to validate it against task-specific constraints or edge cases. A common manifestation is the omission of fine-grained rubric requirements that are essential for correctness, such as downstream demand structure, non-applicability of special categories, second-order behavioral changes, quantitative comparison against core business volatility, or explicit compliance preconditions. In this way, the final answer often does not satisfy the decisive conditions of the task. We also find insufficient boundary checking: the model may infer a threshold or cutoff but does not verify minimal integer cases, maximal feasible cases, or adjacent boundary conditions. 

\paragraph{Bottleneck in Domain-Specific Reasoning and Impractical Problem-Solving (23\% of Errors).} The model often demonstrates directionally plausible but insufficiently grounded reasoning: responses are structurally coherent and capture the general topic, but fail to satisfy the task’s concrete requirements. A common pattern is that the model responds to the topic rather than to the question. As for legal tasks, those failures usually lie in inability to apply domain-specific legal norms or precedent-based reasoning.
Models sometimes avoid hallucination by repeatedly emphasizing limited information, lack of access to external sources, or uncertainty, and fail to proceed with the conditional analysis expected by the rubric. This behavior suggests that surface-level coherence is not sufficient for domain-specific tasks.

\paragraph{Grounding and Factual Errors (17\% of Errors).} We frequently observe that factual grounding remains brittle including missing or incorrect task-critical anchors, such as policy dates, applicability conditions, statutory constraints, or technical boundary conditions. In many cases, the response is syntactically well-formed but lacks the specific evidence needed to justify the conclusion. This issue is particularly problematic in domains such as finance, law, and medicine, where small factual deviations can lead to a quite different reasoning result.  

\paragraph{Instruction Violation (14\% of Errors).} The model may ignore output constraints, replace user-specified methods with more familiar templates, or shift from a requested neutral description to an opinionated recommendation.  In some cases, the model may answer the substance of the question partially correctly while still failing this category because it ignored a required format, omitted mandated evidence, or did not follow the requested procedural approach. Such violations are especially harmful in benchmark settings that rely on strict rubric compliance. Overall, these failures indicate that improving domain-specific performance requires not only better general reasoning, but also stronger grounding, more robust verification, and tighter alignment with task-specific instructions.

\paragraph{Fact Hallucination and Refusal (10\% of Errors).} We also observe that the model either refuses to answer when information is incomplete or fills the gap with plausible but unsupported details. The former leads to overly conservative outputs that underutilize information available in the prompt; the latter introduces hallucinated facts, including numerical estimates or assumed future disclosures. Both behaviors reduce utility in settings where conditional reasoning is required. Ideally, the model should distinguish between missing information that invalidates a claim and missing information that still permits a constrained, conditional answer.

\paragraph{Calculation Errors (2\% of Errors).} The model can often perform isolated arithmetic steps, but struggles with comparisons, ratios, and relative impact assessment. As a result, it may report absolute changes without translating them into normalized quantities or effects, making it difficult to assess materiality. Relatively, the model may state that an effect is limited without providing the numerical comparison needed to support that claim. This limitation is especially salient in tasks requiring threshold analysis, sensitivity estimation, or comparison against a baseline.
\section{Related Work}

\paragraph{Agentic Benchmarks.}
Recent works like xbench target professional productivity tasks~\cite{chen2025xbench}, while SWE-bench measures software engineering capability through GitHub issue resolution~\cite{jimenez2024swe} and $\tau$-bench evaluate conversational interactions under policy constraints~\cite{yao2025tau,barres2025tau}. Vending-Bench and FutureX examine long-horizon coherence and future prediction in dynamic scenarios~\cite{backlund2025vending,zeng2025futurex}. Compared to \benchmark, prior testbeds such as GDPVal \cite{patwardhan2026gdpval}, RLI \cite{mazeika2025remote}, and APEX-Agents \cite{vidgen2026apex} target economically valuable tasks, measuring performance through proxies merely rather than directly quantifying the real-world earnings an agent is capable of delivering with reproducible evaluations. \textbf{Unlike GDPVal, RLI, and APEX-Agents, \benchmark focus more on consistent delivery (\S\ref{subsec:consistent_delivery}), pareto optimality (\S\ref{subsec:pareto_frontier}), and diachronical analyses (\S\ref{subsec:diachronical_analysis}) for realistic grounding.} 



\paragraph{Reality-grounded Evaluations.}
A complementary direction evaluates agents in real-world environments where outcomes are determined by external reality. 
LiveTradeBench formalizes this paradigm using streaming financial markets, showing that leaderboard performance does not reliably translate into real trading outcomes~\cite{yu2025livetradebench}. 
Public competitions such as Alpha Arena further demonstrate the high variance and regime sensitivity of agent performance in dynamic markets~\cite{nof1ai2026alphaarena}. 
Beyond finance, Lab-Bench evaluates agents in real scientific workflows involving experimental data and literature analysis~\cite{laurent2024lab}. 
While these assessments offer strong ecological validity, they sacrifice diagnostic resolution: failures are difficult to attribute to specific reasoning, retrieval, and tool execution issues.



\section{Conclusion}

In this work, we introduce \dataset{}, a testbed aimed at bridging the gap between static exam-style evaluations and the high-stakes demands of professional deployment. By integrating expert-curated workflows with a rubric-based assessment focusing on logical coherence, factual grounding, and professional compliance, \dataset{} sets a rigorous standard for evaluating the economic readiness of language agents. 
\dataset{} emphasizes the need to shift the evaluation focus from superficial correctness to a framework that prioritizes genuine decision-making capabilities for agent optimization.





\bibliography{custom}
\bibliographystyle{abbrvnat}

\newpage
\appendix

\clearpage
\renewcommand \thepart{}
\renewcommand \partname{}
\part{Appendix}
\section*{Table of Contents}
\setcounter{tocdepth}{2}
\renewcommand{\contentsname}{Appendix Contents}
\startcontents[appendix]  
\printcontents[appendix]{}{1}{}

\clearpage

\section{Authors}\label{app:authors}

\subsection{Data Contributors}
\textit{We sincerely appreciate all experts for their invaluable data and feedback in advancing AI. Listed below are major contributors who agreed to be named {(}alphabetical by surname{)}}:
\\
\\
Mingyang Che, Jiarui Chen, Hanqi Li, Muzhen Liu, Hang Lu, Pan Wang, Shuo Wang, Wenzhou Wang, Chenyi Xie, Zhenjie Ye, Xia Yue, Jiayi Zhang, Yanru Zhang, Yongmou Zhao, Zishen Zhou.

\section{Economic Value Audit of \dataset }\label{app:eco_value_calculation}

\subsection{Calculation Method for Reproducibility}



\paragraph{Formulation.} The economic value of \benchmark is estimated by summing the value of each question (\cf Eq. \ref{eq:value_compute}) from senior experts' required time cost and hourly wage listed in \cref{tab:eco_values}:
\begin{equation}
\label{eq:value_compute}
V_{\text{\benchmark}} = \sum_{q \in Q} V_q
= \sum_{q \in Q} T_q^{\text{ExpertCost}} \times W_q^{\text{HourlyWage}} .
\end{equation}

We provide a detailed table of task values (\cf Table \ref{tab:task_value_list}) for calculating or reproducing the economic value results for any tested agent or model on \benchmark.

\paragraph{Expert Cost.}  To ensure our valuation is grounded in reality, we gather time estimates for each task creation from 2 to 3 senior experts in their respective domains. This collaborative approach minimizes subjective bias, allowing us to derive a reliable mean estimation for task completion times. 

\paragraph{Wage Anchoring.} We developed a robust and multi-faceted wage reference that considers regional and sectoral nuances. For instance, in the United States, we utilized the latest Occupational Employment and Wage Statistics (OEWS) from the U.S. Bureau of Labor Statistics, complementing it with industry-specific reports for high-income roles, such as financial analysts\footnote{\url{https://www.biglawinvestor.com/biglaw-salary-scale}} and medical specialists~\cite{mgma2024, biglaw2024}. In China, we tailored our estimates to reflect the economic realities of tier-1 cities, drawing data from recent labor market wage guidelines.

All calculations are standardized to an hourly basis using a typical work year of 2{,}080 hours ($52$ weeks\ $\times 5$ days\ $\times 8$ hours). We infer the implied hourly wage by dividing the reported annual salary by 2{,}080. To make compensation figures comparable across sources, we apply an overhead multiplier of $1.3$ \emph{only} when a source reports wages and salaries (\ie, wage-only figures). This adjustment is motivated by the BLS \emph{Employer Costs for Employee Compensation} (ECEC) release for June 2025 (released in 12, Sep, 2025): total employer compensation for civilian workers averaged \$48.05 per hour worked, comprising \$33.02 in wages and salaries and \$15.03 in benefits. Benefits therefore
account for about $31\%$ of total compensation \(\left(\frac{15.03}{48.05} \approx 31\%\right)\)\footnote{\url{https://www.bls.gov/news.release/ecec.nr0.html}}. For sources that already report total compensation, we do not apply any multiplier.

\paragraph{Calculation for Reproduction.}
We provide the detailed list of economic value of each task in (\$) USD or (\textyen) CNY unit depicted in Table \ref{tab:eco_values} for reproducing the results in \benchmark.

\clearpage
\newpage

\subsection{Task Economic Value List}
\label{subsec:task_list}

\small
\begin{longtable}{l l l l l}

  \caption{Task economic value of \$OneMillion-Bench} \label{tab:task_value_list} \\
  \toprule
  \textbf{Task UUID} & \textbf{Domain} & \textbf{CNY} & \textbf{USD} & \textbf{Subset} \\
  \midrule
  \endfirsthead

  \caption*{\textit{Table \ref*{tab:task_value_list} (continued)}} \\
  \toprule
  \textbf{UUID} & \textbf{Domain} & \textbf{CNY} & \textbf{USD} & \textbf{Subset} \\
  \midrule
  \endhead

  \midrule
  \multicolumn{5}{r}{\textit{Continued on next page\ldots}} \\
  \endfoot

  \bottomrule
  \endlastfoot

  f4cdab18-7636-4416-bcc0-9f40705687a5 & Economics and Finance & 5,793 & -- & cn \\
  9709dc9b-ea0a-4e14-af06-3c636798b242 & Economics and Finance & 9,837 & -- & cn \\
  50cea244-cd7b-4213-94f0-eb199e056eb5 & Economics and Finance & 6,664 & -- & cn \\
  5e9d346b-e174-4f7c-bc0e-c0ddd3c0bb64 & Economics and Finance & 7,298 & -- & cn \\
  0781ab70-3513-4da6-8032-a327aba51d26 & Economics and Finance & 3,862 & -- & cn \\
  a4d2c2e8-9143-414a-86df-b76d565283c9 & Economics and Finance & 9,202 & -- & cn \\
  c6fa3167-2fad-48c0-9160-62dde6424769 & Economics and Finance & 8,568 & -- & cn \\
  4ab521dd-fcdb-4f61-b8e8-9eaf47fb392d & Economics and Finance & 10,154 & -- & cn \\
  3ef5a591-1158-42c7-8f49-427f7979634a & Economics and Finance & 6,651 & -- & cn \\
  b7e58dda-226a-42a8-8699-f98bbd5f60ad & Economics and Finance & 6,346 & -- & cn \\
  281a940f-16dd-4cff-bfd0-054c426e01bb & Economics and Finance & 5,793 & -- & cn \\
  4a961bc6-a0b7-4f2f-9a00-d5fc3365a19f & Economics and Finance & 11,106 & -- & cn \\
  1826f17c-0c59-44b9-925b-eeea8663429f & Economics and Finance & 11,106 & -- & cn \\
  5c786cc2-d1ed-48b3-8417-07c1910a5f94 & Economics and Finance & 6,222 & -- & cn \\
  5a220c4d-1035-48aa-bc6f-f0f87cee79a8 & Economics and Finance & 4,720 & -- & cn \\
  921ce736-5056-431e-bdab-769b161d3ccf & Economics and Finance & 8,250 & -- & cn \\
  b090fa25-d637-4f96-add1-546c49628b25 & Economics and Finance & 7,298 & -- & cn \\
  472e9e99-f6c2-43df-b9e2-5e3feefaadda & Economics and Finance & 5,077 & -- & cn \\
  98da94c3-3733-4eca-b6c9-daf6929aa8c7 & Economics and Finance & 8,568 & -- & cn \\
  3f500bde-303b-4b22-a11f-871eb1fbd705 & Economics and Finance & 9,520 & -- & cn \\
  14e8f749-4d45-4fef-b9cf-c47788bfc908 & Economics and Finance & 8,885 & -- & cn \\
  d08cefc6-dcb2-433b-88a4-e4ed3e9c19e1 & Economics and Finance & 6,346 & -- & cn \\
  172516cf-4cff-49d3-90e4-30d8cbd285ca & Economics and Finance & 3,218 & -- & cn \\
  6a029f36-71d0-429b-a64c-9ef950d2686a & Economics and Finance & 8,885 & -- & cn \\
  16c841ac-f21b-4983-b48a-7248f3c8eb7b & Economics and Finance & 4,760 & -- & cn \\
  3100747e-ced0-4493-beaa-a37ee2b1ef97 & Economics and Finance & 6,007 & -- & cn \\
  0d503359-64f3-47a8-9911-30baddd5eab7 & Economics and Finance & 4,291 & -- & cn \\
  c1805f6c-c29e-41e8-a1c5-349bb14c113c & Economics and Finance & 11,106 & -- & cn \\
  61e05096-325a-4f3c-b1b4-74a34d0896d9 & Economics and Finance & 9,520 & -- & cn \\
  ec9352fa-56c0-4a2f-b7aa-5cc2b3850f58 & Economics and Finance & 8,885 & -- & cn \\
  034226e4-d37b-4cbf-b57d-74a018123f18 & Economics and Finance & 9,520 & -- & cn \\
  39774cd5-aa36-428c-95fc-020ebc38ad8f & Economics and Finance & 11,106 & -- & cn \\
  21439c3b-52b4-43fd-a4ba-b5304051f9d0 & Economics and Finance & 5,712 & -- & cn \\
  8cfaeaa6-f275-4e8e-bef8-8785855ed244 & Economics and Finance & 9,837 & -- & cn \\
  9f8000ca-9a59-4f06-8107-0e223b65e826 & Economics and Finance & 10,789 & -- & cn \\
  dabd5ce8-8348-4ea8-a697-3fb7376636af & Economics and Finance & 3,218 & -- & cn \\
  a3a7dbe4-5521-4c6b-8a42-c7ee61ded640 & Economics and Finance & 4,291 & -- & cn \\
  bc8b6415-8fcd-4633-b7bb-76466823036b & Economics and Finance & 7,509 & -- & cn \\
  11e6f0de-2dee-4a0a-bb5a-6211deb59bec & Economics and Finance & 3,433 & -- & cn \\
  559f2275-0ebb-4e53-8568-d9b6335bfdff & Economics and Finance & 7,080 & -- & cn \\
  e1bcc059-86a7-4b4d-8e5c-9de507d6ef1c & Economics and Finance & -- & 6,042 & global \\
  aaf8b644-94ea-4a68-8386-a44761be88a7 & Economics and Finance & -- & 6,334 & global \\
  68795f96-1443-44d0-903c-020ca1d32c59 & Economics and Finance & -- & 3,814 & global \\
  b342585e-9665-444b-9f42-4afd346b3955 & Economics and Finance & -- & 4,486 & global \\
  bbdd25c4-f8cc-4208-bc4d-f9a9d56847d7 & Economics and Finance & -- & 4,131 & global \\
  160179a5-f900-4c06-971b-b79229f44ad3 & Economics and Finance & -- & 4,131 & global \\
  d4ed1bc5-eff1-4b41-b787-ec13b010a0cb & Economics and Finance & -- & 4,131 & global \\
  72539acb-922f-4e39-804f-20c2035d3c60 & Economics and Finance & -- & 3,814 & global \\
  68f1fe22-77da-4d5b-886c-8fc4e18ad459 & Economics and Finance & -- & 5,005 & global \\
  0f141139-c43b-486f-a9f9-3699547585d7 & Economics and Finance & -- & 4,277 & global \\
  5e7c850f-2b48-437c-9dab-4e2e164d6373 & Economics and Finance & -- & 3,679 & global \\
  e1b94c86-b6c9-43f6-8251-2e513e5efc52 & Economics and Finance & -- & 3,840 & global \\
  13916e64-de22-4b8a-b790-62d3c1a19740 & Economics and Finance & -- & 5,751 & global \\
  6946a41f-5b87-46fc-bf3a-83ffd2986b24 & Economics and Finance & -- & 5,460 & global \\
  ebdcd944-aa53-4727-9591-3c8ce27bd000 & Economics and Finance & -- & 8,117 & global \\
  38b7dfbd-df19-47a6-9e4e-f31e571d20b6 & Economics and Finance & -- & 4,906 & global \\
  4d15d7a3-4f47-4c4d-a8db-a7c6fe0e75f7 & Economics and Finance & -- & 4,771 & global \\
  2381e458-11f0-4878-a4c4-6442d3cde24a & Economics and Finance & -- & 5,309 & global \\
  8077372b-e72d-43a3-bac4-3ff995040ba1 & Economics and Finance & -- & 4,131 & global \\
  45435d84-9e68-40c9-b2f0-ce69b248c89b & Economics and Finance & -- & 3,259 & global \\
  0d8b4484-e80d-482b-8b4a-65803e60ecb3 & Economics and Finance & -- & 5,606 & global \\
  06150229-a28a-4842-a93a-fa7dfeb9031d & Economics and Finance & -- & 4,423 & global \\
  db77b8bf-04c8-49bc-a8a4-578c41589636 & Economics and Finance & -- & 3,948 & global \\
  f1dcbc6e-133f-4f1b-8b74-9e07f69d74f4 & Economics and Finance & -- & 6,479 & global \\
  ccd2f927-5479-42b8-bb7d-07b16b7c5f8a & Economics and Finance & -- & 3,125 & global \\
  e8c50005-5621-46a4-8f1d-014317bac459 & Economics and Finance & -- & 4,423 & global \\
  df72fe47-0a47-48d0-bf0c-fa39509908df & Economics and Finance & -- & 3,259 & global \\
  6c0bf4ed-e56a-4ca7-b21c-23038468133f & Economics and Finance & -- & 4,486 & global \\
  15b6abcd-80ab-42c9-ab11-aa03d2b216bc & Economics and Finance & -- & 5,040 & global \\
  8002472d-7ccd-4970-8e01-67e9a7cf793f & Economics and Finance & -- & 3,125 & global \\
  15ae0a17-0b9a-4f19-b57c-27b76cc1e8c8 & Economics and Finance & -- & 3,814 & global \\
  67c5d15f-65d3-45ba-9df8-debcc08e0a47 & Economics and Finance & -- & 3,410 & global \\
  972f6b74-3e6a-41eb-a7c8-b8e388999d8f & Economics and Finance & -- & 4,859 & global \\
  f0cfd1a0-efbf-43b9-826e-98ea95fee9d8 & Economics and Finance & -- & 6,188 & global \\
  fbe57877-ec38-4cda-bd95-fb5c773662c4 & Economics and Finance & -- & 4,351 & global \\
  5b070842-8ff7-484b-8ea7-98c3a19fe739 & Economics and Finance & -- & 5,169 & global \\
  ea45e6d4-d05f-4372-9587-c77a38d6f250 & Economics and Finance & -- & 4,568 & global \\
  7c5ca57b-b6f3-4b6b-85f3-73c60dd0878c & Economics and Finance & -- & 3,986 & global \\
  ad1240ff-6ede-4794-82af-0e2d31e487a9 & Economics and Finance & -- & 4,131 & global \\
  d73d983f-e531-4bc9-8a17-9ace27450c09 & Economics and Finance & -- & 3,948 & global \\
  14c80191-6712-46fb-a48e-0c3e4e714c7a & Law & 6,180 & -- & cn \\
  4a5571ee-41f4-4bab-aa0c-8220ae6b7191 & Law & 3,675 & -- & cn \\
  98eb28f3-ec92-4a07-aafa-9997f96955aa & Law & 4,343 & -- & cn \\
  c902d3e0-c8a4-4610-90ec-b4e93da13ab9 & Law & 3,629 & -- & cn \\
  4f44b9e3-06f8-4e04-8983-1cbec827fb7a & Law & 4,510 & -- & cn \\
  d3ea340a-8fcf-4cbf-9532-fd50e4ced0ac & Law & 5,512 & -- & cn \\
  9aea07f8-60c1-4ea4-bde3-d1562fc1aa1b & Law & 3,508 & -- & cn \\
  d4bf75d1-4e83-4f87-94f0-a60c44c8c16f & Law & 5,679 & -- & cn \\
  1b355e8f-25c3-44c5-8d76-55b791d686e4 & Law & 3,007 & -- & cn \\
  a12269fe-0ae3-403f-8e4a-ec1278568307 & Law & 6,180 & -- & cn \\
  a7a67d1c-d7fa-4e8e-9141-01f0b4649e01 & Law & 4,666 & -- & cn \\
  86af7cd8-a582-4035-b0d2-788292786c90 & Law & 4,510 & -- & cn \\
  4cca073f-5e11-40db-9a5f-8909144a2beb & Law & 3,341 & -- & cn \\
  746cfd85-18e9-4667-aa63-a61f88022f7f & Law & 6,514 & -- & cn \\
  bd130497-af77-4163-a49c-ac195d164685 & Law & 4,510 & -- & cn \\
  5931cbd0-aa90-4e0d-8b37-2fac30b6c5a4 & Law & 6,347 & -- & cn \\
  022812c3-d191-4ab5-82d7-cd4cbc8019f0 & Law & 5,512 & -- & cn \\
  bf9888e6-f5ee-4d00-ab88-7209753d7d05 & Law & 5,011 & -- & cn \\
  079fbcc9-3520-407e-83fa-623378b2f36c & Law & 5,679 & -- & cn \\
  dc8277a4-e3c9-45e8-a942-fe1f80273776 & Law & 3,174 & -- & cn \\
  43a0f47c-21eb-4323-8eeb-a7da3e641758 & Law & 4,343 & -- & cn \\
  4656b5dc-73de-405c-bccb-fcc77bf63cc1 & Law & 4,510 & -- & cn \\
  fba1cdc6-e08d-440d-be8a-36f85a06fb70 & Law & 6,180 & -- & cn \\
  5b7d77f0-4f75-4579-8338-3a9be4aeeaa9 & Law & 4,925 & -- & cn \\
  907e27a2-8a94-4ac5-8686-7f2d839497f5 & Law & 3,675 & -- & cn \\
  0397f35a-5e9e-4281-a74f-217f4df2786d & Law & 4,343 & -- & cn \\
  d0b5291f-ed30-4e32-aca6-b380f1748e35 & Law & 4,009 & -- & cn \\
  64c29e74-abce-41b1-b275-22c055e5a6da & Law & 5,846 & -- & cn \\
  f63e490e-1205-4417-8325-d0b64811bbc6 & Law & 5,512 & -- & cn \\
  0f0eac70-bb29-4f45-9423-7d8dc8fd4ce7 & Law & 6,514 & -- & cn \\
  3ac266ca-c193-48fe-85c0-f4e9522961ce & Law & 4,176 & -- & cn \\
  dac34e8c-217c-447d-8694-6dbe34c0746c & Law & 4,677 & -- & cn \\
  7fc7f2ab-d56f-438a-b9ae-424323f96e0c & Law & 5,011 & -- & cn \\
  611387e2-7a44-4ee5-add7-2563a45b5264 & Law & 3,110 & -- & cn \\
  46b5480b-5b9a-4d4c-b115-45bf60b616f0 & Law & 4,677 & -- & cn \\
  448d98aa-660d-498a-8213-213f1065ff96 & Law & 5,178 & -- & cn \\
  d7d9214f-ac7c-487d-852f-20845caa413a & Law & 3,370 & -- & cn \\
  dd459add-fe84-4de2-b177-f470389454a6 & Law & 3,842 & -- & cn \\
  0d3acaab-f7de-454b-bba9-fe4694acc77c & Law & 5,011 & -- & cn \\
  8124b83e-9ee9-4a99-b703-28bb492a3688 & Law & 4,510 & -- & cn \\
  7da100c7-4054-419b-8226-c3e97dce0b54 & Law & -- & 12,302 & global \\
  23d9a97d-aba9-4dd4-9a1f-35acde6c735f & Law & -- & 12,302 & global \\
  749e0b9b-eeed-4f57-a8ad-2759f4a268d9 & Law & -- & 7,796 & global \\
  707c4831-8137-45a6-b889-3396695f8205 & Law & -- & 9,239 & global \\
  d01ba3e0-4620-4baa-8b26-5d6597fbd7f0 & Law & -- & 7,189 & global \\
  76fc9645-cb52-41f4-9971-21e67773b1f4 & Law & -- & 5,543 & global \\
  3b763792-7c5c-42ac-81d7-b2b2d31309d3 & Law & -- & 4,911 & global \\
  97dc696f-bb3b-4476-b818-f42a42c0f192 & Law & -- & 7,391 & global \\
  61ee7292-5e3b-4bce-bd76-c97e0d4ffad2 & Law & -- & 6,961 & global \\
  2d015f0d-4564-4a26-b5d4-fd2114fb3161 & Law & -- & 7,189 & global \\
  926c1afb-e115-4e4d-831a-f857afc6a5c1 & Law & -- & 5,431 & global \\
  92ecbb65-9f51-47f9-b918-d6ab932c70bc & Law & -- & 9,442 & global \\
  1bd5358d-f6ed-47cb-8a02-d303e2e1407d & Law & -- & 6,961 & global \\
  386cc233-cdf0-43cd-9614-c8e436e00ead & Law & -- & 7,999 & global \\
  d78a6473-ab98-4cb4-bdfe-c46ed5b67a08 & Law & -- & 3,764 & global \\
  1070d2a9-1576-4e8a-aa25-d01d2f37e477 & Law & -- & 12,302 & global \\
  dd3dca51-0b81-419c-ad75-c5c3ed7544d7 & Law & -- & 8,201 & global \\
  7ad023af-e1fc-44e4-afc3-b04098ddb324 & Law & -- & 5,341 & global \\
  42943063-a1b0-419b-a62f-0ee09c4d5785 & Law & -- & 7,511 & global \\
  d728baca-5d5d-47fe-882d-dbc6e1de9f55 & Law & -- & 7,999 & global \\
  68c0e9e4-4e01-4137-b62e-c523a3e0116e & Law & -- & 9,011 & global \\
  33e7a536-ac9f-465d-b34d-76b29f3512c0 & Law & -- & 8,201 & global \\
  420150e9-93c1-4d5e-9760-966ebb66daf4 & Law & -- & 8,404 & global \\
  dbbd68e9-9a20-42ff-ba8a-87218345f1cc & Law & -- & 6,961 & global \\
  136d58d2-bd74-4049-93fd-3df23cbdeb91 & Law & -- & 9,442 & global \\
  6faea514-a56c-4aca-9414-593c34a1b7b1 & Law & -- & 11,289 & global \\
  6183a0ce-401d-4763-a960-2509c2384765 & Law & -- & 10,252 & global \\
  96c86547-d48b-4ca5-8303-8be965d4ebed & Law & -- & 7,999 & global \\
  a47b504a-b3f6-4e21-8fe4-30512c1db407 & Law & -- & 3,627 & global \\
  ff42d583-5064-4cb3-8ecc-ab562d56c060 & Law & -- & 6,758 & global \\
  588ca9fa-84b9-4ed8-962a-65b86e0500b5 & Law & -- & 4,708 & global \\
  dc5967e8-18bb-4b4f-8c50-b0ce6eb2c87b & Law & -- & 9,442 & global \\
  bd0cfce5-9056-4697-a35e-bb6e13865a22 & Law & -- & 5,543 & global \\
  e3b60852-b6ba-4eed-a2f0-e8ae2a364bc9 & Law & -- & 7,796 & global \\
  6f9511f6-951a-4c97-ac10-694c0d201748 & Law & -- & 4,506 & global \\
  02f114bf-c1ba-4d21-b8cc-c1c31ab495f9 & Law & -- & 7,796 & global \\
  24804afe-da39-4fd7-908f-56e21ab03b9c & Law & -- & 12,302 & global \\
  580310a9-bb84-401f-96bc-55a1eaa5447c & Law & -- & 7,796 & global \\
  41af7f51-32e2-4a97-b4dc-fb2ea631ebdc & Law & -- & 3,902 & global \\
  330c4240-c6d4-4830-b1a1-788da3f69591 & Law & -- & 5,138 & global \\
  443c4bff-5502-4658-91c4-0f5b26a4d4f0 & Healthcare and Medicine & -- & 6,067 & global \\
  68922ab6-13c7-422c-9e0c-2429a4fafd7b & Healthcare and Medicine & -- & 7,660 & global \\
  ec20a6b0-7320-454e-8398-8f382a89efd2 & Healthcare and Medicine & -- & 6,716 & global \\
  c437a033-7f62-473c-810d-35ed79d60df6 & Healthcare and Medicine & -- & 11,622 & global \\
  a4119dbd-d5aa-4b1f-a215-d2f1b7285165 & Healthcare and Medicine & -- & 7,320 & global \\
  3ecca542-c7a0-456d-8897-6c85f9732d11 & Healthcare and Medicine & -- & 7,660 & global \\
  5cc4e65d-8984-4352-8dc4-f707c9da5eb1 & Healthcare and Medicine & -- & 9,773 & global \\
  d9a3a37b-2fde-46f3-90c6-48a6d2215322 & Healthcare and Medicine & -- & 10,376 & global \\
  df0b46a6-087d-4d4c-804e-dfbe7fc2f67f & Healthcare and Medicine & -- & 11,018 & global \\
  2b54a7d1-2d11-45ff-985e-5ce390ec0e1c & Healthcare and Medicine & -- & 9,169 & global \\
  669ccd28-509f-4e5b-a3c6-f4cc7d8930cf & Healthcare and Medicine & -- & 10,376 & global \\
  6a94f803-014f-4323-9c44-2bb80a59e19f & Healthcare and Medicine & -- & 4,658 & global \\
  c1c2c26f-0384-4f3c-82a3-5482b7009d56 & Healthcare and Medicine & -- & 11,018 & global \\
  f003d40a-1190-409d-841e-84da045d24cc & Healthcare and Medicine & -- & 10,074 & global \\
  31fe657a-50a8-46c8-8186-f6b8da611584 & Healthcare and Medicine & -- & 7,660 & global \\
  f223b123-621c-466f-b9fc-8e0cbe2785bd & Healthcare and Medicine & -- & 9,471 & global \\
  92ed8a85-f03b-458c-8b5a-9fea360ed649 & Healthcare and Medicine & -- & 9,471 & global \\
  e1b922a9-ccd0-4a44-ae27-1d51c6cae4f2 & Healthcare and Medicine & -- & 10,376 & global \\
  d1aa95fa-4b01-4bde-a289-bb91bf489f79 & Healthcare and Medicine & -- & 6,298 & global \\
  a6424187-9d45-4659-8a10-2f0f5e4524d4 & Healthcare and Medicine & -- & 12,225 & global \\
  85a7c05e-2235-4d0c-ba8c-a2cca33ab401 & Healthcare and Medicine & -- & 7,660 & global \\
  77b68a7d-c72f-4c21-88d6-36917feb5e85 & Healthcare and Medicine & -- & 5,811 & global \\
  d8bd8164-b21f-4f6b-8ec6-7d058c514412 & Healthcare and Medicine & -- & 7,660 & global \\
  0f464232-4413-4972-9e19-021d7d793161 & Healthcare and Medicine & -- & 8,565 & global \\
  82664b2b-6471-4f66-a496-e35a67a18d40 & Healthcare and Medicine & -- & 9,773 & global \\
  4b2c9234-229a-416b-9993-77b03b11d05e & Healthcare and Medicine & -- & 8,867 & global \\
  7bcf6aef-2235-45cd-8b02-02e8d9a00a39 & Healthcare and Medicine & -- & 4,428 & global \\
  bb0a8654-2760-4fa0-9fa6-0e047c0d8d71 & Healthcare and Medicine & -- & 11,923 & global \\
  4e906852-19f1-4380-9fc8-7f39af4ad3ca & Healthcare and Medicine & -- & 8,867 & global \\
  308de99f-a853-45a6-be11-67be5b113333 & Healthcare and Medicine & -- & 8,397 & global \\
  26212219-9a4a-4afb-9cbe-819dbf0f73bf & Healthcare and Medicine & -- & 4,888 & global \\
  af865a52-51d3-4d3e-b170-d8d249e57f95 & Healthcare and Medicine & -- & 8,263 & global \\
  e0921259-ac28-410d-94ad-6e795cd3250d & Healthcare and Medicine & -- & 4,888 & global \\
  5a2689ad-fbc9-45cd-b35f-2bdd5a33fb84 & Healthcare and Medicine & -- & 6,716 & global \\
  3fc30be9-8aff-4c7e-afa3-d4c60c7b170b & Healthcare and Medicine & -- & 11,622 & global \\
  9fe8853f-42c5-41d4-8622-79e5b6a9a753 & Healthcare and Medicine & -- & 8,263 & global \\
  d19a66bd-6552-4d0f-9574-6128f4f7ff05 & Healthcare and Medicine & -- & 5,119 & global \\
  96d762f1-aaee-473d-a092-1024ce63872e & Healthcare and Medicine & -- & 6,067 & global \\
  01995bb2-e019-4195-844e-4012250583c9 & Healthcare and Medicine & -- & 6,113 & global \\
  7db19fe4-4164-45d0-947e-61bef3496e75 & Healthcare and Medicine & -- & 4,658 & global \\
  673490a8-a490-4809-8a7c-f2950685c331 & Healthcare and Medicine & 3,050 & -- & cn \\
  86b11c02-b600-44ff-9806-176a723cdddf & Healthcare and Medicine & 4,051 & -- & cn \\
  f6d9a7e6-6d20-4f89-9937-1c9ba7836e73 & Healthcare and Medicine & 3,552 & -- & cn \\
  d99d37b8-14d6-4df3-ada7-f5b3721747cb & Healthcare and Medicine & 6,146 & -- & cn \\
  4c394bc9-4f72-4b9d-b5ec-7b26968c8d68 & Healthcare and Medicine & 3,871 & -- & cn \\
  7b7fe40e-e5d7-4cc0-887b-3c1d216bee06 & Healthcare and Medicine & 4,051 & -- & cn \\
  2008063e-3dd5-4b76-831f-f06ef4e7d739 & Healthcare and Medicine & 5,168 & -- & cn \\
  611c260e-93ef-4391-bbdd-0beda5f267de & Healthcare and Medicine & 5,487 & -- & cn \\
  c6b234bb-2142-4b26-993d-a0c2984a3943 & Healthcare and Medicine & 5,826 & -- & cn \\
  dcfde4cd-fcb9-4390-94ea-3b8016158250 & Healthcare and Medicine & 4,849 & -- & cn \\
  7cc331b2-297b-4311-ba03-a5ed0a27996b & Healthcare and Medicine & 5,487 & -- & cn \\
  c91c56f9-e432-484a-aaeb-867c7143d4bd & Healthcare and Medicine & 2,342 & -- & cn \\
  908f40e1-2d85-4e57-9320-cfc6fc450037 & Healthcare and Medicine & 5,826 & -- & cn \\
  eb90ce6e-8a53-4fc4-8b71-4fc2f6913cdc & Healthcare and Medicine & 5,328 & -- & cn \\
  ce8ed8a0-8312-4464-ac08-cd33095eb019 & Healthcare and Medicine & 4,051 & -- & cn \\
  81bf36b9-d2b6-40e6-9f4e-870fa0212b31 & Healthcare and Medicine & 5,008 & -- & cn \\
  44847f8f-e10a-493b-a867-e505808f510f & Healthcare and Medicine & 5,008 & -- & cn \\
  b6502878-bfac-4be7-9851-d7364dd2ed3c & Healthcare and Medicine & 5,487 & -- & cn \\
  cd123b23-75a5-4619-8b9e-c77c7f5d9672 & Healthcare and Medicine & 3,165 & -- & cn \\
  9a22b80c-9d00-488b-9e1e-7f9579efe2c2 & Healthcare and Medicine & 6,465 & -- & cn \\
  12ec26fa-84c0-4639-b5c5-9e00c62dca0c & Healthcare and Medicine & 4,051 & -- & cn \\
  06be9028-f633-4737-98f3-b523421b1276 & Healthcare and Medicine & 3,073 & -- & cn \\
  1135b8c9-60eb-4cd1-abfa-3d11bd522bb8 & Healthcare and Medicine & 4,051 & -- & cn \\
  653c82d5-3dfa-44f7-b570-ca9ae59aa715 & Healthcare and Medicine & 4,529 & -- & cn \\
  3107f221-d3cc-4401-b0e1-0b2fe36264f4 & Healthcare and Medicine & 5,168 & -- & cn \\
  e71191ec-40f7-4156-b68d-b70754672fc4 & Healthcare and Medicine & 4,689 & -- & cn \\
  ba4bb671-0fcc-4f6d-bb37-ea12aab8e5d8 & Healthcare and Medicine & 2,226 & -- & cn \\
  fe491b1f-b37b-4f37-8e93-509ebcd000e9 & Healthcare and Medicine & 6,305 & -- & cn \\
  8eb9fe93-8198-491c-a0ae-2d59b62b635a & Healthcare and Medicine & 4,689 & -- & cn \\
  f0597e81-d2ba-4069-a6de-8f86d29964e4 & Healthcare and Medicine & 5,826 & -- & cn \\
  81e7a86e-6dfc-4a5b-8fdc-281924e24a45 & Healthcare and Medicine & 3,392 & -- & cn \\
  b625b3c1-a92d-4234-911a-dfab5072dd51 & Healthcare and Medicine & 4,370 & -- & cn \\
  1d156a89-d475-4c7a-8f19-ffc4a8514323 & Healthcare and Medicine & 2,457 & -- & cn \\
  b9616065-9034-471b-8410-d53f231f1b5a & Healthcare and Medicine & 3,552 & -- & cn \\
  4bb63011-2170-4483-b171-2c74d2bf6102 & Healthcare and Medicine & 6,146 & -- & cn \\
  dd646f83-7a44-4e24-8240-f626624fa179 & Healthcare and Medicine & 4,370 & -- & cn \\
  d97af248-9a73-4a5b-8e9b-a406004cb5b2 & Healthcare and Medicine & 2,573 & -- & cn \\
  578de37e-4cff-4436-8c44-d2cdc176d598 & Healthcare and Medicine & 3,050 & -- & cn \\
  de587f91-b1b2-4ec5-aa58-e48b2c9e5fb9 & Healthcare and Medicine & 3,232 & -- & cn \\
  8ab658c4-f8c5-4533-9025-1dbcc2dad0ae & Healthcare and Medicine & 3,232 & -- & cn \\
  865cca91-9fca-4611-9dfe-2de5eec3ca41 & Natural Sciences & -- & 1,340 & global \\
  fe5662f9-a6d6-4aad-9449-5156e61b9cec & Natural Sciences & -- & 2,147 & global \\
  944d060e-e3af-4f44-9ae4-51920e6a7898 & Natural Sciences & -- & 2,971 & global \\
  68bfa996-706b-43e1-8a91-1af695ba5bb5 & Natural Sciences & -- & 1,340 & global \\
  659a8308-83be-434b-9a74-d0d60e72b4d1 & Natural Sciences & -- & 2,391 & global \\
  f5bd1ec1-b1bf-49b2-be47-c2c5ecc5d1df & Natural Sciences & -- & 1,730 & global \\
  32809395-def0-4e6b-afa2-b80a37620384 & Natural Sciences & -- & 3,378 & global \\
  eae3cc5a-49e4-4a07-bc7d-e01ada5a0847 & Natural Sciences & -- & 2,042 & global \\
  47bd8a68-2c91-4ceb-af3c-7f113bf4d213 & Natural Sciences & -- & 2,066 & global \\
  c300a2f7-7b4c-4cf7-8385-6f49834c8a09 & Natural Sciences & -- & 1,340 & global \\
  97a49948-4c98-41ed-9092-1ef88bc09e1f & Natural Sciences & -- & 1,214 & global \\
  228e062e-2dfb-450a-828d-446f5c18bb9a & Natural Sciences & -- & 1,277 & global \\
  3452218f-1d0d-45fc-ac72-a0f0d6fa17e5 & Natural Sciences & -- & 1,853 & global \\
  a7024d80-956d-491c-b1bb-8567c7571999 & Natural Sciences & -- & 1,466 & global \\
  8ca39d59-ba11-48de-8c7a-27a5d4f7935a & Natural Sciences & -- & 1,600 & global \\
  431c7ede-5c91-4a49-b4b3-cc94229822fc & Natural Sciences & -- & 2,428 & global \\
  a6f3f009-2786-4347-8b13-fa7e614695ab & Natural Sciences & -- & 1,979 & global \\
  2daed9dc-b3fa-4724-b0da-470da4e672a7 & Natural Sciences & -- & 1,403 & global \\
  bea6d62b-80ec-4d95-81e8-c3f5952d63d2 & Natural Sciences & -- & 1,853 & global \\
  31777574-0def-4e66-ad92-4274013cca63 & Natural Sciences & -- & 1,151 & global \\
  833a9fb2-537a-42cd-ad0e-86b0df8ca415 & Natural Sciences & -- & 1,916 & global \\
  274f5e8a-056f-4585-b0fb-2564868b793a & Natural Sciences & -- & 2,554 & global \\
  2e32c0a9-9382-4943-8594-309d29187539 & Natural Sciences & -- & 1,214 & global \\
  b2d97709-e4f5-4bba-aee5-6676ca0ecdc2 & Natural Sciences & -- & 1,403 & global \\
  b24a5d38-b539-4653-9cdc-dc3ca4e1f151 & Natural Sciences & -- & 1,916 & global \\
  6bfba637-bd4f-469a-9bc5-6b92a55a3e59 & Natural Sciences & -- & 1,600 & global \\
  bf61d44d-97fe-4365-9fed-8d3ac8440f54 & Natural Sciences & -- & 1,151 & global \\
  585708d8-25c5-4e91-a779-5cf6e6fb1bf2 & Natural Sciences & -- & 2,744 & global \\
  5fabbd4d-2295-4426-87c4-7fb735f4b657 & Natural Sciences & -- & 2,635 & global \\
  404b288d-d5d8-4568-8de4-1f5ecedba486 & Natural Sciences & -- & 1,403 & global \\
  2929294b-54cd-4fe1-8d28-2ae6f7c81e03 & Natural Sciences & -- & 1,466 & global \\
  923f1733-de66-4d8f-a9c3-603a8b4b7158 & Natural Sciences & -- & 1,974 & global \\
  8f8101f1-f874-4058-a523-5f01bbbec0be & Natural Sciences & -- & 1,663 & global \\
  3227a346-1115-4c4d-a0b6-bc5cd3ea914f & Natural Sciences & -- & 2,491 & global \\
  3cbe5faf-28b2-457c-bcd7-af179f5aec83 & Natural Sciences & -- & 2,491 & global \\
  6e9131aa-dfd5-41e5-a2cf-d01041e895f5 & Natural Sciences & -- & 1,340 & global \\
  dc4db810-7b0e-4110-bfb1-d53471680291 & Natural Sciences & -- & 1,600 & global \\
  e92ecabf-a10a-46e1-98b2-a15d1ae073ac & Natural Sciences & -- & 3,516 & global \\
  dd20f5a3-670b-4fef-96f4-15d5a4186492 & Natural Sciences & -- & 2,890 & global \\
  d4f2f740-c9b8-46f5-859d-b9e0fdb84f79 & Natural Sciences & -- & 1,340 & global \\
  fd46a64b-90d3-481f-86da-8820d6cc7a38 & Natural Sciences & 1,751 & -- & cn \\
  644d838a-bfeb-4043-b595-748527157415 & Natural Sciences & 3,485 & -- & cn \\
  3e2cc4d7-0a4b-4862-bcf8-905535bf1829 & Natural Sciences & 4,823 & -- & cn \\
  8e8151b8-1e45-411f-ab02-41614bb7400e & Natural Sciences & 1,751 & -- & cn \\
  6ca999e2-cdd0-4691-89d1-7534f5b564ce & Natural Sciences & 3,882 & -- & cn \\
  d2b65bf0-3872-4d24-8e52-1dc386fca06d & Natural Sciences & 2,808 & -- & cn \\
  dac5282c-448c-44f3-9f76-48e823852d17 & Natural Sciences & 5,484 & -- & cn \\
  ee952885-9c32-4085-ac10-aa1536ade2ef & Natural Sciences & 2,668 & -- & cn \\
  d042f45e-f76a-4e39-ad94-bfd68a67e578 & Natural Sciences & 3,353 & -- & cn \\
  2f5d1c57-c071-4b73-a226-d146d8b59c68 & Natural Sciences & 1,751 & -- & cn \\
  b4d5950c-18f3-42f6-8f9b-6867c9949136 & Natural Sciences & 1,586 & -- & cn \\
  059e1e9e-6e3e-4518-b880-b59173f1480e & Natural Sciences & 1,669 & -- & cn \\
  438aaf2a-eab2-40eb-b3d6-1ca4368585b8 & Natural Sciences & 2,420 & -- & cn \\
  2c32b379-8c60-4648-97e5-c897dc46a623 & Natural Sciences & 1,916 & -- & cn \\
  5ce6255b-3526-4916-9681-fcb5d6810da0 & Natural Sciences & 2,091 & -- & cn \\
  6591d711-404f-48d0-ac7f-10f7094caeaf & Natural Sciences & 3,172 & -- & cn \\
  546dc745-d460-41a1-a9a0-0ecc446a31d6 & Natural Sciences & 2,585 & -- & cn \\
  cd276d89-416c-4490-b688-3c79424c9699 & Natural Sciences & 1,833 & -- & cn \\
  76ea552b-e0f8-46f8-9203-e75903181843 & Natural Sciences & 2,420 & -- & cn \\
  611d8974-aed7-45a1-b201-b1d2b380f82e & Natural Sciences & 1,504 & -- & cn \\
  3604da43-a774-4bd1-ae74-9d99a72354b1 & Natural Sciences & 2,503 & -- & cn \\
  810bf406-5b65-4f47-b3f4-c62e507a08ee & Natural Sciences & 4,146 & -- & cn \\
  c3c93440-64ac-401e-8aa3-a0027dce2b6a & Natural Sciences & 1,586 & -- & cn \\
  07167000-3548-4fd7-ac80-0e12d54ac72f & Natural Sciences & 1,833 & -- & cn \\
  e73d9f41-3ccf-45e6-a387-e00472bfedd7 & Natural Sciences & 2,503 & -- & cn \\
  d78cd986-1823-40b4-bf9d-931726a5d526 & Natural Sciences & 2,091 & -- & cn \\
  938165db-06c0-428b-bdda-1f14ac25dbff & Natural Sciences & 1,504 & -- & cn \\
  44f67f0a-d62b-4ff0-b985-4a45d84e8911 & Natural Sciences & 3,584 & -- & cn \\
  1b32355f-cd10-4e1f-aa8e-28cf4b5f2848 & Natural Sciences & 4,278 & -- & cn \\
  949ac7a6-6b7e-46a1-8af1-c2963432b236 & Natural Sciences & 1,833 & -- & cn \\
  66bf0a68-2d9b-457c-8817-19e13401cd2d & Natural Sciences & 1,916 & -- & cn \\
  280f0df8-7242-426a-91ac-5d61f1b48832 & Natural Sciences & 3,204 & -- & cn \\
  d1c4b15f-9721-4827-840c-c6510a32b8dd & Natural Sciences & 2,173 & -- & cn \\
  2bc64a8e-1210-4dfd-aa30-ba02eceedd5a & Natural Sciences & 3,255 & -- & cn \\
  8622c778-80b1-442c-8997-de6e2405dd59 & Natural Sciences & 3,255 & -- & cn \\
  1fc999bd-c4e5-4d04-b638-bfbb866d508f & Natural Sciences & 1,751 & -- & cn \\
  6de37f4f-8e50-4b1e-b5d2-8dbbcea8c3ca & Natural Sciences & 2,091 & -- & cn \\
  1dba23ad-c86b-4e82-a0e0-fa6c3d649538 & Natural Sciences & 4,594 & -- & cn \\
  54e0d81e-67ec-4370-b886-9864ba328add & Natural Sciences & 4,691 & -- & cn \\
  457f5ca4-ba92-42b2-a82e-c4e6678e4f04 & Natural Sciences & 1,751 & -- & cn \\
  9c2672f2-430f-44cc-823b-2784f3a84a2b & Industry & -- & 5,285 & global \\
  36d1050a-72a1-4210-94c8-cd654bc25132 & Industry & -- & 2,469 & global \\
  44b07f0a-d6c2-4b74-bfdd-2c7b5a87e1c7 & Industry & -- & 3,645 & global \\
  54a18a27-264b-42c7-9762-c030dc97bfe1 & Industry & -- & 3,165 & global \\
  56cc66ac-2bde-4ce0-9c9e-8c816026020c & Industry & -- & 2,469 & global \\
  f0bc2660-fdb2-4e05-9766-e8db285d52e2 & Industry & -- & 1,458 & global \\
  64b0f439-36d8-4fa1-8e81-81d37447cccf & Industry & -- & 3,350 & global \\
  6c34d48e-cc4e-4193-be63-a1094f541fe8 & Industry & -- & 3,969 & global \\
  bba510cb-38c5-4c70-88e3-14780a0491c6 & Industry & -- & 4,860 & global \\
  ee809a60-d4e2-4514-a3b5-175eb591f21e & Industry & -- & 4,125 & global \\
  f0bfdae8-df07-4c23-8ac0-08abcd0329be & Industry & -- & 4,108 & global \\
  89055e11-d0a5-4e7b-8d15-e1002f4cae54 & Industry & -- & 2,259 & global \\
  302703db-b1c9-43e1-9684-c524139d6545 & Industry & -- & 1,588 & global \\
  fad912f4-b576-4cd0-b659-9530ee97ade2 & Industry & -- & 2,382 & global \\
  2b4583f1-f332-43c6-999e-9166e6d2c6e5 & Industry & -- & 4,860 & global \\
  591d5f51-ed05-4c50-8abd-7a79c1333497 & Industry & -- & 2,670 & global \\
  58a83321-05c2-4d66-86d0-1fbca0dc4741 & Industry & -- & 2,382 & global \\
  d6357459-83e7-4bb6-8cbd-d2ff97c46a2f & Industry & -- & 2,110 & global \\
  0d785b81-fc88-438c-964b-e4c8256f9f42 & Industry & -- & 1,827 & global \\
  550f71ad-94fe-40a0-9688-6acad415e987 & Industry & -- & 3,089 & global \\
  5e785915-8448-461e-b663-9a8c9c777c4b & Industry & -- & 1,746 & global \\
  e2487a94-a128-4b67-9ed9-3fb056e28d5b & Industry & -- & 3,263 & global \\
  6b4ea30a-e9f1-4a30-a3bc-c6c006354ade & Industry & -- & 4,850 & global \\
  d46c9edd-b2e7-41f0-83a9-faf219653650 & Industry & -- & 3,089 & global \\
  6267d153-72b8-45ef-81a5-d88db387096c & Industry & -- & 2,295 & global \\
  8276aec4-036f-459b-8770-2f15f693edd2 & Industry & -- & 4,260 & global \\
  54246069-1446-41a1-8d17-89a0e5c52dc0 & Industry & -- & 1,849 & global \\
  0defb49c-e4f5-451b-9821-807d77e8c8b3 & Industry & -- & 1,530 & global \\
  025bf5b1-99d0-4d08-872e-5c4c7a77c5dd & Industry & -- & 4,500 & global \\
  c68fc03a-3fdc-4cca-824f-f14db3730027 & Industry & -- & 2,475 & global \\
  1fb6bbf5-394d-49c7-ae64-bc374b0fcc3b & Industry & -- & 1,849 & global \\
  dcc3dbb7-0cbb-4f75-ae06-0efbb8987ad0 & Industry & -- & 1,827 & global \\
  71f76252-1db0-41f8-93bf-338a89f17341 & Industry & -- & 1,827 & global \\
  d6344516-e4f6-42b3-a2a6-f1329ed16dbf & Industry & -- & 1,458 & global \\
  364cfadc-3764-4ffd-a2c2-2688ae800950 & Industry & -- & 3,645 & global \\
  1dbd3e38-1c81-4cf1-94eb-d1d24f9a2d49 & Industry & -- & 2,382 & global \\
  6129a42f-a16c-4a28-8912-c60332e70406 & Industry & -- & 2,295 & global \\
  deabcd1e-64ae-4659-82ac-646b517cdfc2 & Industry & -- & 2,951 & global \\
  b0cc7823-86a6-4a3e-acdd-0d20af89806e & Industry & -- & 2,106 & global \\
  fd1f36d1-c29a-4077-91f4-b0d23cb38f5b & Industry & -- & 1,899 & global \\
  ebb94405-b0af-4485-8efa-c95c3971a7b1 & Industry & 9,020 & -- & cn \\
  42f2f642-deb1-46c6-a47e-b78c51b2cbcc & Industry & 4,213 & -- & cn \\
  a961442b-491d-469e-80e6-37c6b0301563 & Industry & 4,401 & -- & cn \\
  c2a430c0-7f98-4ab5-af84-93824c790d72 & Industry & 3,821 & -- & cn \\
  3aae3e0f-2c68-4575-86e4-015a146bc88e & Industry & 4,213 & -- & cn \\
  7c72e1b6-0eab-4e02-9a67-0661576dd212 & Industry & 1,242 & -- & cn \\
  2ddd40b6-f110-47f4-ae3d-3fee5b01716f & Industry & 5,716 & -- & cn \\
  da515c84-7adc-4290-9807-67c26de850d5 & Industry & 6,774 & -- & cn \\
  b5f5aecd-7ae0-4e6e-97ec-8dce1101d9bf & Industry & 5,868 & -- & cn \\
  520cdb68-0103-4ee3-80b5-0466a392a408 & Industry & 4,980 & -- & cn \\
  ea67f32d-c827-438f-951e-17206a6f196c & Industry & 3,637 & -- & cn \\
  b29b17f5-d321-494e-a7fa-c4ec4fdb9a4f & Industry & 1,924 & -- & cn \\
  226b82d8-e406-4f1d-bd92-cb0adfb5374f & Industry & 2,710 & -- & cn \\
  088a6376-95c1-4b3b-a45e-4a5d19ea4b6e & Industry & 4,064 & -- & cn \\
  ab801e6d-7974-4cc6-b63a-36ba0faa0575 & Industry & 5,868 & -- & cn \\
  d29e29db-affb-494c-b5b7-1b193cff24bd & Industry & 3,224 & -- & cn \\
  94e8ae5c-1bf8-49bc-bd40-f774644cc7f1 & Industry & 4,064 & -- & cn \\
  7e0ab98e-379e-4b1e-acff-3ff37bffaf32 & Industry & 3,600 & -- & cn \\
  6fd07669-8d90-4ace-81ea-9a1ec49cf9aa & Industry & 1,556 & -- & cn \\
  39f36658-5239-4536-a9cd-d0270b652111 & Industry & 5,271 & -- & cn \\
  38b981dc-4f80-40ee-b568-0ed6bf129f62 & Industry & 1,487 & -- & cn \\
  b7f472d0-1733-45f6-86f2-04d050be121c & Industry & 5,568 & -- & cn \\
  c4c00b09-7166-4d8c-b03d-273a9b43da6f & Industry & 8,277 & -- & cn \\
  84ec9ed0-abae-45d9-a303-935b47f23954 & Industry & 5,271 & -- & cn \\
  467db57d-6c16-468f-b97b-cc33c94891dd & Industry & 3,916 & -- & cn \\
  b0d00294-81d8-453f-b8e1-76543c490717 & Industry & 5,143 & -- & cn \\
  e79e0354-4848-4896-8c61-47ae21566204 & Industry & 3,155 & -- & cn \\
  d6b99b4e-c7f9-4cab-8064-e782ad87115f & Industry & 1,303 & -- & cn \\
  3c8f83a2-96db-42b2-812d-643cfb1d99e4 & Industry & 5,433 & -- & cn \\
  cce48b66-20eb-4d4a-9dfb-fff4e75968f8 & Industry & 2,108 & -- & cn \\
  c037de9f-eb16-452e-832d-0af01f0ff50e & Industry & 3,155 & -- & cn \\
  4bd30355-9179-4db2-b19f-acfe08b71833 & Industry & 1,556 & -- & cn \\
  a708701f-f2c5-40a6-b14d-d2f77f8924c9 & Industry & 1,556 & -- & cn \\
  e771ab18-f120-4c54-8450-aa2367b71bf6 & Industry & 1,242 & -- & cn \\
  c545314e-f070-43c7-bec7-c426687d27cf & Industry & 4,401 & -- & cn \\
  9a88315b-27e8-4263-a73a-4016befe01a4 & Industry & 4,064 & -- & cn \\
  af4dcf04-29fe-4ba3-8d2f-d1bdae0406ab & Industry & 3,916 & -- & cn \\
  eb7638ac-035f-4b6c-a207-ebc6804aca3a & Industry & 2,613 & -- & cn \\
  34b542fb-9418-4485-a235-67c9f6eb505f & Industry & 1,865 & -- & cn \\
  bea3bcd5-4ef3-47f2-b563-7300ff506a2a & Industry & 1,617 & -- & cn \\

\end{longtable}
\section{Category Distribution of \dataset}
\label{category}
We identified a total of \textbf{92} unique third-level subdomain tags in \dataset in five high-level domains. The non-synonymous tags across languages are kept as separate entries

\paragraph{Healthcare and Medicine (25 tags):}
\begin{itemize}
  \item Cell and Gene Therapy (CGT)
  \item Gastrointestinal Surgery
  \item Pathophysiology
  \item Hepatobiliary and Pancreatic Surgery
  \item Cardiothoracic Surgery
  \item Thyroid and Breast Surgery
  \item Pathology
  \item Obstetrics and Gynecology
  \item Rheumatology and Immunology
  \item Biopharmaceuticals (Macromolecules)
  \item Orthopedics
  \item Molecular Biology
  \item Ophthalmology
  \item Medical Genetics
  \item Dentistry
  \item Emergency and Critical Care Medicine
  \item Cardiovascular Medicine
  \item Neurology
  \item Medical Laboratory Science
  \item Respiratory Medicine
  \item Immunology
  \item Endocrinology and Metabolic Diseases
  \item Oncology
  \item Nephrology
  \item Neurosurgery
\end{itemize}

\paragraph{Industry (12 tags):}
\begin{itemize}
  \item Semiconductors
  \item Telecommunications
  \item Machine Learning
  \item Backend Development
  \item Databases and Data Engineering
  \item Civil Engineering
  \item Chemical Engineering and Materials
  \item Architectural Design
  \item Systems / Embedded / 3D Rendering
  \item Frontend Development
  \item Machinery and Automation
  \item Materials
\end{itemize}

\paragraph{Law (19 tags):}
\begin{itemize}
  \item Criminal Defense
  \item Contract Disputes
  \item Restructuring / Financing / M\&A
  \item Corporate Governance
  \item Data Compliance/Cybersecurity
  \item Financial Regulation
  \item Government Regulation
  \item Copyright
  \item Law-Other
  \item Trademark
  \item Patent
  \item Labor Law
  \item Marriage/Inheritance
  \item Criminal Compliance
  \item Securities \& Listing (IPO) 
  \item Funds \& Asset Management
  \item Public International Law
  \item Private International Law
  \item Tort Disputes
\end{itemize}

\paragraph{Economics and Finance (18 tags):}
\begin{itemize}
  \item Equities
  \item Bonds
  \item Mergers \& Acquisitions
  \item Management/Consulting/Business Analysis
  \item Risk Management
  \item VC/PE
  \item Accounting
  \item Investing-Other
  \item Equity Financing
  \item Commodities
  \item Consumer Finance 
  \item Corporate Banking 
  \item Other
  \item Cryptocurrency
  \item Derivatives
  \item Macroeconomics
  \item Life Insurance
  \item Quantitative
\end{itemize}

\paragraph{Natural Science (18 tags):}
\begin{itemize}
  \item Condensed Matter Physics
  \item Physics-Other
  \item Organic Chemistry
  \item Inorganic Chemistry
  \item Materials Chemistry
  \item Physical and Theoretical Chemistry
  \item Classical Physics
  \item Quantum Physics
  \item Applied Mathematics
  \item Mathematics-Other
  \item Microbiology
  \item Molecular Biology
  \item Molecular and Cell Biology
  \item Cell Biology
  \item Biochemistry
  \item Genetics and Genomics
  \item Ecology
  \item Other
\end{itemize}

\begin{figure*}[htbp]
    \centering
    \includegraphics[width=0.98\columnwidth]{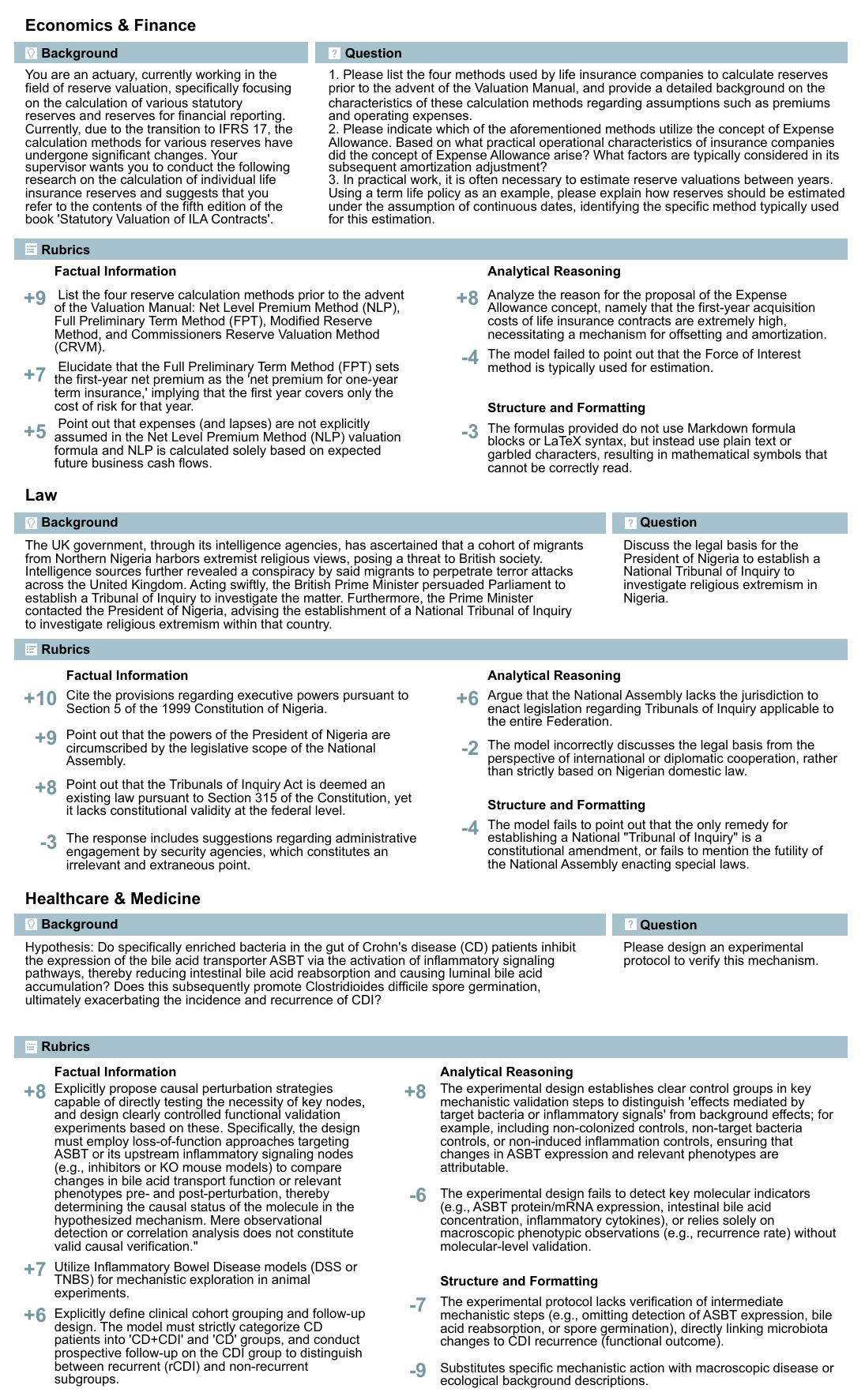} 
    \caption{Sample cases of different domains with varying rubric weights and tags.}
    \label{fig:omb_examples}
\end{figure*}

\section{Further Analyses}\label{app:analysis}

\subsection{Diachronical Analysis}
\label{subsec:diachronical_analysis}

\begin{wrapfigure}{r}{0.5\textwidth}
    \centering
    \includegraphics[width=1\linewidth]{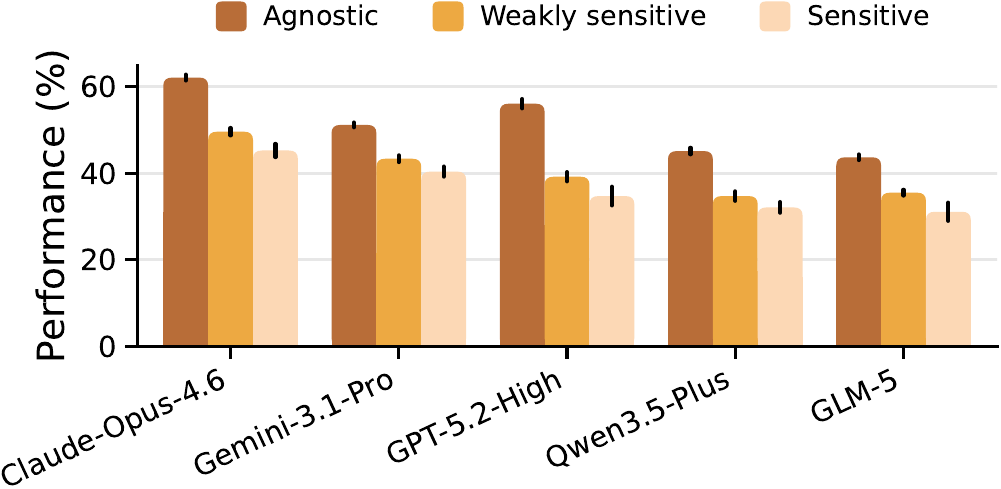}
    \captionsetup{width=1\linewidth}
    \caption{Diachronical analysis (\ie, time-sensitivity analysis) on \benchmark.}
    \label{fig:diachronical_comparison}
\end{wrapfigure}

As depicted in Figure~\ref{fig:diachronical_comparison}, all models consistently achieve higher performance on temporally agnostic questions than on weakly sensitive and sensitive ones, revealing that temporal sensitivity poses a genuine and progressive challenge. The performance drop from agnostic to sensitive is most pronounced for Claude-Opus-4.6 and GPT-5.2-High ($\sim$15\% to 20\%), suggesting these models may over-rely on temporal anchoring cues when reasoning. Notably, Qwen3.5-Plus and GLM-5 exhibit narrower gaps across categories, but at the cost of lower overall performance, indicating limited temporal reasoning capacity rather than robustness. These findings suggest that strong performance and resilience to temporal sensitivity are not always correlated, a nuance worth considering when selecting agents for real-world, time-evolving information tasks.

\subsection{Correlation Analysis of Model Judge \texorpdfstring{\vs}{vs} Human Judge}
\label{subsec:correlation_analysis}

\begin{figure*}[htbp]
    \centering
    \includegraphics[width=\columnwidth]{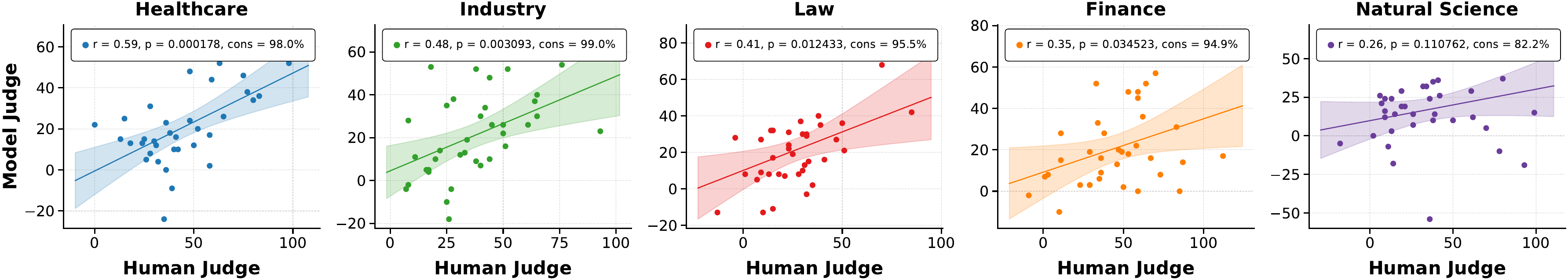} 
    \caption{Correlation of Expert Score between model judge and human judge.}
    \label{fig:correlation}
\end{figure*}

As shown in Figure \ref{fig:correlation}, we investigate the correlation between the model judge and human judge. We employ the Pearson coefficient \cite{pearson1896vii} and pairwise consistency to measure the correlation degree of each domain subset in \benchmark.

Overall, the model judge demonstrates moderate-to-strong agreement with human judges across most domains, though alignment varies notably by field. \textbf{Healthcare} exhibits the strongest correlation ($r = 0.59$, $p < 0.001$, cons $= 98.0\%$), suggesting that model judgments are most reliable in domains with well-established, objective clinical criteria. \textbf{Industry} and \textbf{Law} follow with $r = 0.48$ ($p < 0.01$) and $r = 0.41$ ($p < 0.05$), respectively, both achieving high pairwise consistency ($\geq 95.5\%$). \textbf{Finance} yields a weaker yet still significant correlation ($r = 0.35$, $p < 0.05$, cons $= 94.9\%$). In contrast, \textbf{Natural Science} shows the lowest alignment ($r = 0.26$, $p = 0.11$, cons $= 82.2\%$), with the correlation failing to reach statistical significance, indicating that expert scientific judgment is the hardest for models to approximate, likely due to the greater subjectivity and depth of domain expertise required. These results suggest that while model judges can serve as reliable proxies in most professional domains, caution is warranted in highly specialized scientific contexts.

\subsection{Robustness to the cliff threshold}\label{subsec:cliff_effect}
In order to ablation the cliff threshold 0.7 used in the economic value, we added a sensitivity analysis of main results using  rankings varies from 0.6 to 1.0 in \cref{tab:rankings}.

\begin{table}[htbp]
\centering
\renewcommand{\arraystretch}{1.35}
\setlength{\tabcolsep}{10pt}
\caption{Model ranking results ($\Delta$) using various cliff thresholds.}\label{tab:rankings}
\resizebox{\textwidth}{!}{%
\begin{tabular}{lccccc}
\toprule
\textbf{Model / Agentic System} & \textbf{Rank@0.6} & \textbf{Rank@0.7} & \textbf{Rank@0.8} & \textbf{Rank@0.9} & \textbf{Rank@1.0} \\
\midrule
Claude-Opus-4.6           & 1  & 1  & 1  & 1  & 1  \\
GPT-5.4-High              & 2  & 2  & 2  & 2  & 3 (+1) \\
Qwen3.5-Plus              & 3  & 3  & 3  & 4 (+1) & 2 (-1) \\
GPT-5.2-High              & 4  & 4  & 4  & 3 (+1) & 4  \\
Kimi-K2.5                 & 5  & 5  & 6 (+1) & 7 (+2) & 6 (+1) \\
GPT-5.3-Codex             & 6  & 6  & 5 (-1) & 5 (-1) & 5 (-1) \\
Gemini-3-Pro-Preview      & 7  & 7  & 7  & 6 (-1) & 8 (+1) \\
DeepSeek-V3.2-Speciale    & 8  & 8  & 9 (+1) & 10 (+2) & 9 (+1) \\
Grok-4                    & 9  & 9  & 8 (-1) & 8 (-1) & 7 (-2) \\
Doubao-Seed-2.0-Pro       & 10 & 10 & 10 & 9 (-1) & 10 \\
\bottomrule
\end{tabular}%
}
\label{tab:rankings_more}
\end{table}

From the table, we observe that the sequential rankings of the top 10 LLMs/Agents are highly consistent with varying thresholds. We argue that, compared to absolute scores, we provide multiple metrics such as Economic Value, Expert Score, and Pass Rate to reflect model capabilities from different dimensions. At the same time, the partial order of model rankings deserves more attention than absolute scores. The cutoff of 0.7 is sufficient to demonstrate that the results in the paper are credible and reliable.

\section{Evaluation Details}\label{app:Evaluation}
\subsection{Prompt}
As depicted in Figure \ref{fig:prompt}, we list the prompt template adopted for all evaluations in this work.

\begin{figure}
\begin{tcolorbox}[colback=LightGray!5!white,colframe=DarkGray,title=Prompt for Rubric-based Evaluation in \benchmark,coltitle=white,fonttitle=\bfseries]
Your response should be in the following format:\\[1em]
\textbf{Explanation:} \{your explanation for your answer choice\}\\[1em]
\textbf{Answer:} \{your chosen answer\}\\[1em]
\textbf{Confidence:} \{your confidence score between 0\% and 100\% for your answer.\}
\end{tcolorbox}
\caption{Prompt template rubric-based evaluation.}
\label{fig:prompt}
\end{figure}

\subsection{Sampling Parameters}
\begin{table}[ht]
\small
\centering
\caption{Sampling parameters adopted in the models evaluation against the \benchmark.}
\label{tab:model_configs}
\begin{tabular}{llll}
\toprule
\textbf{Model} & \textbf{Search Provider} & \textbf{Thinking Effort} & \textbf{Misc. Parameters} \\
\midrule
DeepSeek-V3.2 & OpenRouter & [On by default] & Temperature = 0 \\
 & & & Max new tokens = 65,536 \\
\midrule
DeepSeek-V3.2-Speciale & OpenRouter & [On by default] & Temperature = 0 \\
 & & & Max new tokens = 65,536 \\
\midrule
MiniMax-M2.1 & OpenRouter & [On by default] & Temperature = 0 \\
 & & & Max new tokens = 65,536 \\
\midrule
MiniMax-M2.5 & OpenRouter & [On by default] & Temperature = 0 \\
 & & & Max new tokens = 196,608 \\
\midrule
Qwen3-235B-A22B-2507 & OpenRouter & [On by default] & Temperature = 0 \\
 & & & Max new tokens = 262,100 \\
\midrule
Hunyuan-2.0-Thinking & Tencent & [On by default] & Temperature = 0 \\
 & & & Max new tokens = 32,768 \\
\midrule
Step-3.5-Flash & OpenRouter & [On by default] & Temperature = 0 \\
 & & & Max new tokens = 256,000 \\
\midrule
Doubao-Seed-2.0-Pro-260215 & ByteDance & High & Temperature = 0 \\
 & & & Max new tokens = 128,000 \\
\midrule
Doubao-Seed-1.8-251228 & ByteDance & High & Temperature = 0 \\
 & & & Max new tokens = 64,000 \\
\midrule
GLM-4.7 & OpenRouter & [On by default] & Temperature = 0 \\
 & & & Max new tokens = 131,072 \\
\midrule
GLM-5 & OpenRouter & [On by default] & Temperature = 0 \\
 & & & Max new tokens = 131,072 \\
\midrule
Kimi-K2-Thinking & OpenRouter & [On by default] & Temperature = 0 \\
 & & & Max new tokens = 262,144 \\
\midrule
Kimi-K2.5 & OpenRouter & High & Temperature = 0 \\
 & & & Max new tokens = 262,144 \\
\midrule
GPT-5.4-High & OpenAI & High & Temperature = 0 \\
 & & & Max new tokens = 128,000 \\
\midrule
GPT-5.2-High & OpenAI & High & Temperature = 0 \\
 & & & Max new tokens = 128,000 \\
\midrule
GPT-5.3-Codex & OpenAI & High & Temperature = 0 \\
 & & & Max new tokens = 128,000 \\
\midrule
Claude-Opus-4.5 & Anthropic & High & Temperature = 0 \\
 & & & Max new tokens = 64,000 \\
\midrule
Claude-Opus-4.6 & Anthropic & High & Temperature = 0 \\
 & & & Max new tokens = 128,000 \\
\midrule
Claude-Sonnet-4.5 & Anthropic & High & Temperature = 0 \\
 & & & Max new tokens = 64,000 \\
\midrule
Claude-Sonnet-4.5 & Anthropic & High & Temperature = 0 \\
 & & & Max new tokens = 64,000 \\
\midrule
Gemini-3-Pro-preview & Google & High & Temperature = 0 \\
 & & & Max new tokens = 65,500 \\
\midrule
Grok-4 & xAI & [On by default] & Temperature = 0 \\
 & & & Max new tokens = 256,000 \\
\midrule
Qwen3-Max-2026-01-23 & Alibaba & High & Temperature = 0 \\
 & & & Max new tokens = 32,768 \\
\midrule
Qwen3.5-Plus-2026-02-15 & Alibaba & High & Temperature = 0 \\
 & & & Max new tokens = 65,536 \\
\bottomrule
\end{tabular}
\end{table}
As depicted in \cref{tab:model_configs}, we report the sampling parameters adopted in the evaluation procedure.

\subsection{Evaluation Cost Estimation}
\label{subsec:eval_cost}
We estimate the evaluation cost using the following equation:

\begin{equation}
\label{eq:cost_compute}
C_{\text{eval}} = \alpha \left( c_{\text{in}} \cdot T_{\text{in}} + c_{\text{out}} \cdot T_{\text{out}} + c_{\text{tool}} \cdot N_{\text{tool}} + c_{\text{cache}} \cdot T_{\text{cache}} + c_{\text{infra}} \cdot t_{\text{eval}} \right)
\end{equation}

\noindent where $c_{\text{in}}, T_{\text{in}}$, $c_{\text{out}}, T_{\text{out}}$, and $c_{\text{tool}}, N_{\text{tool}}$ are the unit costs and quantities for input tokens, output tokens, and tool calls (\eg, web search), respectively\footnote{Token and tool call pricing are queried via the OpenRouter cost and stats API: \url{https://openrouter.ai/docs/api/reference/overview\#querying-cost-and-stats}.}. $c_{\text{cache}}, T_{\text{cache}}$ denote the cost and volume of cached tokens (\eg, prompt cache reads and writes), and $c_{\text{infra}}, t_{\text{eval}}$ denote the infrastructure cost rate and total wall-clock evaluation time, capturing compute and storage overhead beyond API billing. The scaling factor $\alpha \geq 1$ provides a buffer against underestimation due to retries, rate-limit handling, and other incidental overhead not captured by the individual billing terms; in practice we set $\alpha = 1.2$.



\definecolor{econcolor1}{RGB}{250,253,255} 
\definecolor{econcolor2}{RGB}{235,247,255} 
\definecolor{econcolor3}{RGB}{210,235,255} 
\definecolor{econcolor4}{RGB}{180,220,250} 

\definecolor{scoregray}{RGB}{240,240,240}

\definecolor{myposgreen}{RGB}{34,139,34}
\definecolor{mynegred}{RGB}{200,50,50}
\begin{table}[t]
\centering
\caption{Comparison of Expert Score across different domains on the \textbf{Global evaluation set}.}\label{tab:en_domains}
\resizebox{\textwidth}{!}{%
\begin{tabular}{l|cc|cc|cc|cc|cc}
\toprule
\multirowcell{2.6}{\textbf{Model / Agentic System}} & \multicolumn{2}{c|}{\textbf{Finance}} & \multicolumn{2}{c|}{\textbf{Healthcare}} & \multicolumn{2}{c|}{\textbf{Industry}} & \multicolumn{2}{c|}{\textbf{Law}} & \multicolumn{2}{c}{\textbf{Natural Science}} \\
\cmidrule(lr){2-3}\cmidrule(lr){4-5}\cmidrule(lr){6-7}\cmidrule(lr){8-9}\cmidrule(lr){10-11}
 & \textbf{Vanilla} & \textbf{Search} & \textbf{Vanilla} & \textbf{Search} & \textbf{Vanilla} & \textbf{Search} & \textbf{Vanilla} & \textbf{Search} & \textbf{Vanilla} & \textbf{Search} \\
\midrule
\textsc{GPT-5.4-High} &  42.0 &  68.8 $_{\textcolor{myposgreen}{\uparrow 26.8}}$ &  63.6 &  53.9 $_{\textcolor{mynegred}{\downarrow 9.7}}$ &  54.7 &  58.2 $_{\textcolor{myposgreen}{\uparrow 3.5}}$ &  55.2 &  55.6 $_{\textcolor{myposgreen}{\uparrow 0.5}}$ &  59.4 &  59.4 $_{\textcolor{mynegred}{\downarrow 0.0}}$ \\
\textsc{Claude-Opus-4.6} &  42.1 &  55.5 $_{\textcolor{myposgreen}{\uparrow 13.4}}$ &  68.4 &  73.7 $_{\textcolor{myposgreen}{\uparrow 5.3}}$ &  58.9 &  67.5 $_{\textcolor{myposgreen}{\uparrow 8.7}}$ &  53.0 &  59.5 $_{\textcolor{myposgreen}{\uparrow 6.5}}$ &  52.4 &  58.9 $_{\textcolor{myposgreen}{\uparrow 6.5}}$ \\
\textsc{GPT-5.3-Codex} &  36.2 &  63.2 $_{\textcolor{myposgreen}{\uparrow 27.1}}$ &  53.2 &  48.8 $_{\textcolor{mynegred}{\downarrow 4.4}}$ &  50.4 &  48.2 $_{\textcolor{mynegred}{\downarrow 2.1}}$ &  46.7 &  50.8 $_{\textcolor{myposgreen}{\uparrow 4.0}}$ &  53.8 &  42.0 $_{\textcolor{mynegred}{\downarrow 11.9}}$ \\
\textsc{Qwen3.5-Plus} &  30.7 &  31.3 $_{\textcolor{myposgreen}{\uparrow 0.6}}$ &  59.3 &  60.9 $_{\textcolor{myposgreen}{\uparrow 1.6}}$ &  50.8 &  52.8 $_{\textcolor{myposgreen}{\uparrow 2.0}}$ &  49.7 &  48.6 $_{\textcolor{mynegred}{\downarrow 1.1}}$ &  47.5 &  51.1 $_{\textcolor{myposgreen}{\uparrow 3.6}}$ \\
\textsc{GPT-5.2-High} &  29.3 &  60.7 $_{\textcolor{myposgreen}{\uparrow 31.4}}$ &  59.5 &  59.3 $_{\textcolor{mynegred}{\downarrow 0.2}}$ &  50.1 &  52.5 $_{\textcolor{myposgreen}{\uparrow 2.4}}$ &  48.8 &  57.0 $_{\textcolor{myposgreen}{\uparrow 8.2}}$ &  50.0 &  54.7 $_{\textcolor{myposgreen}{\uparrow 4.7}}$ \\
\textsc{Kimi-K2.5} &  34.6 &  31.6 $_{\textcolor{mynegred}{\downarrow 3.0}}$ &  54.8 &  48.1 $_{\textcolor{mynegred}{\downarrow 6.7}}$ &  46.0 &  44.4 $_{\textcolor{mynegred}{\downarrow 1.5}}$ &  43.8 &  36.9 $_{\textcolor{mynegred}{\downarrow 6.8}}$ &  46.4 &  44.8 $_{\textcolor{mynegred}{\downarrow 1.6}}$ \\
\textsc{Gemini-3-Pro-Preview} &  35.8 &  47.6 $_{\textcolor{myposgreen}{\uparrow 11.8}}$ &  52.3 &  61.7 $_{\textcolor{myposgreen}{\uparrow 9.3}}$ &  42.0 &  50.4 $_{\textcolor{myposgreen}{\uparrow 8.4}}$ &  47.7 &  50.7 $_{\textcolor{myposgreen}{\uparrow 3.0}}$ &  40.6 &  53.6 $_{\textcolor{myposgreen}{\uparrow 13.0}}$ \\
\textsc{DeepSeek-V3.2-Speciale} &  27.6 &  28.5 $_{\textcolor{myposgreen}{\uparrow 1.0}}$ &  48.9 &  45.7 $_{\textcolor{mynegred}{\downarrow 3.1}}$ &  43.2 &  37.1 $_{\textcolor{mynegred}{\downarrow 6.1}}$ &  38.3 &  36.5 $_{\textcolor{mynegred}{\downarrow 1.8}}$ &  44.3 &  45.3 $_{\textcolor{myposgreen}{\uparrow 1.0}}$ \\
\textsc{Doubao-Seed-2.0-Pro} &  25.9 &  35.8 $_{\textcolor{myposgreen}{\uparrow 9.9}}$ &  51.7 &  64.3 $_{\textcolor{myposgreen}{\uparrow 12.6}}$ &  38.9 &  52.6 $_{\textcolor{myposgreen}{\uparrow 13.7}}$ &  35.4 &  55.0 $_{\textcolor{myposgreen}{\uparrow 19.6}}$ &  45.2 &  51.3 $_{\textcolor{myposgreen}{\uparrow 6.1}}$ \\
\textsc{Grok-4} &  27.3 &  51.4 $_{\textcolor{myposgreen}{\uparrow 24.2}}$ &  45.3 &  47.5 $_{\textcolor{myposgreen}{\uparrow 2.1}}$ &  39.0 &  38.5 $_{\textcolor{mynegred}{\downarrow 0.5}}$ &  39.5 &  46.4 $_{\textcolor{myposgreen}{\uparrow 6.8}}$ &  42.0 &  45.6 $_{\textcolor{myposgreen}{\uparrow 3.6}}$ \\
\textsc{GLM-5} &  30.9 &  34.5 $_{\textcolor{myposgreen}{\uparrow 3.7}}$ &  47.7 &  51.9 $_{\textcolor{myposgreen}{\uparrow 4.2}}$ &  36.0 &  39.1 $_{\textcolor{myposgreen}{\uparrow 3.1}}$ &  38.0 &  39.6 $_{\textcolor{myposgreen}{\uparrow 1.6}}$ &  39.9 &  40.5 $_{\textcolor{myposgreen}{\uparrow 0.6}}$ \\
\textsc{MiniMax-M2.1} &  19.9 &  25.4 $_{\textcolor{myposgreen}{\uparrow 5.5}}$ &  46.7 &  41.4 $_{\textcolor{mynegred}{\downarrow 5.2}}$ &  39.6 &  34.6 $_{\textcolor{mynegred}{\downarrow 5.0}}$ &  31.4 &  30.8 $_{\textcolor{mynegred}{\downarrow 0.6}}$ &  36.7 &  37.3 $_{\textcolor{myposgreen}{\uparrow 0.6}}$ \\
\textsc{Hunyuan-2.0} &  25.0 &  22.6 $_{\textcolor{mynegred}{\downarrow 2.4}}$ &  46.4 &  36.7 $_{\textcolor{mynegred}{\downarrow 9.8}}$ &  34.6 &  38.3 $_{\textcolor{myposgreen}{\uparrow 3.8}}$ &  28.7 &  24.9 $_{\textcolor{mynegred}{\downarrow 3.8}}$ &  38.7 &  28.4 $_{\textcolor{mynegred}{\downarrow 10.3}}$ \\
\textsc{Step-3.5-Flash} &  23.5 &  25.9 $_{\textcolor{myposgreen}{\uparrow 2.4}}$ &  45.3 &  43.6 $_{\textcolor{mynegred}{\downarrow 1.7}}$ &  37.9 &  27.8 $_{\textcolor{mynegred}{\downarrow 10.1}}$ &  29.2 &  29.1 $_{\textcolor{mynegred}{\downarrow 0.1}}$ &  36.0 &  38.3 $_{\textcolor{myposgreen}{\uparrow 2.4}}$ \\
\textsc{Ling-2.5-1T} &  24.2 &  22.9 $_{\textcolor{mynegred}{\downarrow 1.3}}$ &  43.5 &  40.8 $_{\textcolor{mynegred}{\downarrow 2.7}}$ &  37.9 &  36.9 $_{\textcolor{mynegred}{\downarrow 1.0}}$ &  23.8 &  25.0 $_{\textcolor{myposgreen}{\uparrow 1.1}}$ &  36.2 &  30.5 $_{\textcolor{mynegred}{\downarrow 5.7}}$ \\
\textsc{MiniMax-M2.5} &  19.9 &  16.7 $_{\textcolor{mynegred}{\downarrow 3.3}}$ &  33.1 &  38.8 $_{\textcolor{myposgreen}{\uparrow 5.7}}$ &  36.4 &  39.3 $_{\textcolor{myposgreen}{\uparrow 2.9}}$ &  27.5 &  26.5 $_{\textcolor{mynegred}{\downarrow 1.0}}$ &  30.5 &  36.2 $_{\textcolor{myposgreen}{\uparrow 5.7}}$ \\
\midrule
\multicolumn{11}{l}{\textbf{\textit{Deep Research Agents}}} \\
\midrule
\textsc{o3-DeepResearch} & \multicolumn{2}{c|}{ 49.4} & \multicolumn{2}{c|}{ 50.9} & \multicolumn{2}{c|}{ 43.9} & \multicolumn{2}{c|}{ 50.7} & \multicolumn{2}{c}{ 36.5} \\
\textsc{Sonar-DeepResearch} & \multicolumn{2}{c|}{ 45.5} & \multicolumn{2}{c|}{ 42.7} & \multicolumn{2}{c|}{ 43.8} & \multicolumn{2}{c|}{ 41.6} & \multicolumn{2}{c}{ 38.0} \\
\textsc{o4-Mini-DeepResearch} & \multicolumn{2}{c|}{ 40.3} & \multicolumn{2}{c|}{ 40.3} & \multicolumn{2}{c|}{ 34.3} & \multicolumn{2}{c|}{ 36.4} & \multicolumn{2}{c}{ 37.1} \\
\bottomrule
\end{tabular}
}%
\end{table}


\definecolor{econcolor1}{RGB}{250,253,255} 
\definecolor{econcolor2}{RGB}{235,247,255} 
\definecolor{econcolor3}{RGB}{210,235,255} 
\definecolor{econcolor4}{RGB}{180,220,250} 

\definecolor{scoregray}{RGB}{240,240,240}

\definecolor{myposgreen}{RGB}{34,139,34}
\definecolor{mynegred}{RGB}{200,50,50}
\begin{table}[t]
\centering
\caption{Comparison of Expert Score across different domains on the \textbf{CN evaluation set}.}\label{tab:cn_domains}
\resizebox{\textwidth}{!}{%
\begin{tabular}{l|cc|cc|cc|cc|cc}
\toprule
\multirowcell{2.6}{\textbf{Model / Agentic System}} & \multicolumn{2}{c|}{\textbf{Finance}} & \multicolumn{2}{c|}{\textbf{Healthcare}} & \multicolumn{2}{c|}{\textbf{Industry}} & \multicolumn{2}{c|}{\textbf{Law}} & \multicolumn{2}{c}{\textbf{Natural Science}} \\
\cmidrule(lr){2-3}\cmidrule(lr){4-5}\cmidrule(lr){6-7}\cmidrule(lr){8-9}\cmidrule(lr){10-11}
 & \textbf{Vanilla} & \textbf{Search} & \textbf{Vanilla} & \textbf{Search} & \textbf{Vanilla} & \textbf{Search} & \textbf{Vanilla} & \textbf{Search} & \textbf{Vanilla} & \textbf{Search} \\
\midrule
\textsc{GPT-5.4-High} &  53.6 &  63.3 $_{\textcolor{myposgreen}{\uparrow 9.7}}$ &  67.3 &  57.1 $_{\textcolor{mynegred}{\downarrow 10.1}}$ &  55.8 &  57.5 $_{\textcolor{myposgreen}{\uparrow 1.7}}$ &  43.9 &  53.6 $_{\textcolor{myposgreen}{\uparrow 9.7}}$ &  60.6 &  62.1 $_{\textcolor{myposgreen}{\uparrow 1.5}}$ \\
\textsc{Claude-Opus-4.6} &  46.9 &  60.9 $_{\textcolor{myposgreen}{\uparrow 14.0}}$ &  68.4 &  70.6 $_{\textcolor{myposgreen}{\uparrow 2.2}}$ &  59.0 &  69.7 $_{\textcolor{myposgreen}{\uparrow 10.7}}$ &  49.4 &  60.9 $_{\textcolor{myposgreen}{\uparrow 11.4}}$ &  55.5 &  60.7 $_{\textcolor{myposgreen}{\uparrow 5.2}}$ \\
\textsc{Qwen3.5-Plus} &  48.4 &  46.9 $_{\textcolor{mynegred}{\downarrow 1.4}}$ &  60.1 &  61.7 $_{\textcolor{myposgreen}{\uparrow 1.5}}$ &  50.2 &  49.0 $_{\textcolor{mynegred}{\downarrow 1.2}}$ &  48.9 &  56.1 $_{\textcolor{myposgreen}{\uparrow 7.2}}$ &  49.6 &  46.7 $_{\textcolor{mynegred}{\downarrow 2.9}}$ \\
\textsc{GPT-5.2-High} &  41.4 &  56.1 $_{\textcolor{myposgreen}{\uparrow 14.8}}$ &  57.0 &  60.8 $_{\textcolor{myposgreen}{\uparrow 3.8}}$ &  46.4 &  50.3 $_{\textcolor{myposgreen}{\uparrow 4.0}}$ &  35.0 &  42.5 $_{\textcolor{myposgreen}{\uparrow 7.4}}$ &  55.8 &  53.3 $_{\textcolor{mynegred}{\downarrow 2.5}}$ \\
\textsc{GPT-5.3-Codex} &  45.6 &  54.7 $_{\textcolor{myposgreen}{\uparrow 9.1}}$ &  57.1 &  54.9 $_{\textcolor{mynegred}{\downarrow 2.2}}$ &  45.6 &  54.6 $_{\textcolor{myposgreen}{\uparrow 9.0}}$ &  37.0 &  41.2 $_{\textcolor{myposgreen}{\uparrow 4.3}}$ &  49.1 &  49.2 $_{\textcolor{myposgreen}{\uparrow 0.1}}$ \\
\textsc{Gemini-3-Pro-Preview} &  46.4 &  56.4 $_{\textcolor{myposgreen}{\uparrow 10.0}}$ &  53.9 &  60.3 $_{\textcolor{myposgreen}{\uparrow 6.5}}$ &  42.5 &  50.9 $_{\textcolor{myposgreen}{\uparrow 8.4}}$ &  40.5 &  54.7 $_{\textcolor{myposgreen}{\uparrow 14.2}}$ &  46.7 &  47.6 $_{\textcolor{myposgreen}{\uparrow 0.9}}$ \\
\textsc{Kimi-K2.5} &  48.8 &  42.5 $_{\textcolor{mynegred}{\downarrow 6.3}}$ &  53.0 &  46.6 $_{\textcolor{mynegred}{\downarrow 6.5}}$ &  39.2 &  46.9 $_{\textcolor{myposgreen}{\uparrow 7.6}}$ &  40.3 &  40.1 $_{\textcolor{mynegred}{\downarrow 0.3}}$ &  45.4 &  42.3 $_{\textcolor{mynegred}{\downarrow 3.1}}$ \\
\textsc{DeepSeek-V3.2-Speciale} &  37.1 &  44.0 $_{\textcolor{myposgreen}{\uparrow 6.9}}$ &  49.9 &  49.5 $_{\textcolor{mynegred}{\downarrow 0.3}}$ &  39.6 &  42.1 $_{\textcolor{myposgreen}{\uparrow 2.5}}$ &  31.0 &  43.2 $_{\textcolor{myposgreen}{\uparrow 12.1}}$ &  48.1 &  43.1 $_{\textcolor{mynegred}{\downarrow 5.0}}$ \\
\textsc{GLM-5} &  39.5 &  37.4 $_{\textcolor{mynegred}{\downarrow 2.1}}$ &  48.5 &  48.3 $_{\textcolor{mynegred}{\downarrow 0.1}}$ &  38.3 &  38.6 $_{\textcolor{myposgreen}{\uparrow 0.2}}$ &  37.1 &  39.2 $_{\textcolor{myposgreen}{\uparrow 2.1}}$ &  40.2 &  42.1 $_{\textcolor{myposgreen}{\uparrow 1.9}}$ \\
\textsc{Doubao-Seed-2.0-Pro} &  36.6 &  48.6 $_{\textcolor{myposgreen}{\uparrow 12.0}}$ &  43.9 &  54.4 $_{\textcolor{myposgreen}{\uparrow 10.6}}$ &  43.7 &  50.7 $_{\textcolor{myposgreen}{\uparrow 7.0}}$ &  37.8 &  51.3 $_{\textcolor{myposgreen}{\uparrow 13.5}}$ &  41.3 &  47.4 $_{\textcolor{myposgreen}{\uparrow 6.1}}$ \\
\textsc{Grok-4} &  37.3 &  43.5 $_{\textcolor{myposgreen}{\uparrow 6.2}}$ &  45.5 &  45.2 $_{\textcolor{mynegred}{\downarrow 0.3}}$ &  39.0 &  36.8 $_{\textcolor{mynegred}{\downarrow 2.2}}$ &  35.9 &  30.7 $_{\textcolor{mynegred}{\downarrow 5.2}}$ &  41.1 &  41.8 $_{\textcolor{myposgreen}{\uparrow 0.6}}$ \\
\textsc{Ling-2.5-1T} &  29.5 &  33.6 $_{\textcolor{myposgreen}{\uparrow 4.1}}$ &  46.4 &  42.8 $_{\textcolor{mynegred}{\downarrow 3.6}}$ &  40.3 &  38.9 $_{\textcolor{mynegred}{\downarrow 1.4}}$ &  31.6 &  26.3 $_{\textcolor{mynegred}{\downarrow 5.4}}$ &  35.8 &  36.4 $_{\textcolor{myposgreen}{\uparrow 0.5}}$ \\
\textsc{Hunyuan-2.0} &  36.1 &  36.6 $_{\textcolor{myposgreen}{\uparrow 0.5}}$ &  37.1 &  33.2 $_{\textcolor{mynegred}{\downarrow 3.9}}$ &  41.1 &  34.1 $_{\textcolor{mynegred}{\downarrow 7.0}}$ &  35.2 &  32.1 $_{\textcolor{mynegred}{\downarrow 3.2}}$ &  32.0 &  28.3 $_{\textcolor{mynegred}{\downarrow 3.7}}$ \\
\textsc{Step-3.5-Flash} &  38.5 &  33.8 $_{\textcolor{mynegred}{\downarrow 4.7}}$ &  44.2 &  50.1 $_{\textcolor{myposgreen}{\uparrow 5.9}}$ &  36.1 &  30.3 $_{\textcolor{mynegred}{\downarrow 5.8}}$ &  28.4 &  28.0 $_{\textcolor{mynegred}{\downarrow 0.4}}$ &  33.8 &  36.1 $_{\textcolor{myposgreen}{\uparrow 2.3}}$ \\
\textsc{MiniMax-M2.1} &  31.6 &  35.9 $_{\textcolor{myposgreen}{\uparrow 4.4}}$ &  46.2 &  45.1 $_{\textcolor{mynegred}{\downarrow 1.0}}$ &  43.3 &  38.0 $_{\textcolor{mynegred}{\downarrow 5.3}}$ &  27.6 &  30.2 $_{\textcolor{myposgreen}{\uparrow 2.6}}$ &  30.2 &  34.4 $_{\textcolor{myposgreen}{\uparrow 4.2}}$ \\
\textsc{MiniMax-M2.5} &  29.4 &  29.8 $_{\textcolor{myposgreen}{\uparrow 0.4}}$ &  40.3 &  34.2 $_{\textcolor{mynegred}{\downarrow 6.1}}$ &  39.4 &  34.8 $_{\textcolor{mynegred}{\downarrow 4.6}}$ &  25.0 &  25.2 $_{\textcolor{myposgreen}{\uparrow 0.1}}$ &  33.0 &  36.5 $_{\textcolor{myposgreen}{\uparrow 3.5}}$ \\
\midrule
\multicolumn{11}{l}{\textbf{\textit{Deep Research Agents}}} \\
\midrule
\textsc{o3-DeepResearch} & \multicolumn{2}{c|}{ 47.4} & \multicolumn{2}{c|}{ 46.4} & \multicolumn{2}{c|}{ 40.1} & \multicolumn{2}{c|}{ 34.9} & \multicolumn{2}{c}{ 39.5} \\
\textsc{Sonar-DeepResearch} & \multicolumn{2}{c|}{ 39.1} & \multicolumn{2}{c|}{ 42.0} & \multicolumn{2}{c|}{ 43.1} & \multicolumn{2}{c|}{ 39.2} & \multicolumn{2}{c}{ 37.7} \\
\textsc{o4-Mini-DeepResearch} & \multicolumn{2}{c|}{ 39.5} & \multicolumn{2}{c|}{ 39.8} & \multicolumn{2}{c|}{ 34.7} & \multicolumn{2}{c|}{ 27.7} & \multicolumn{2}{c}{ 38.3} \\
\bottomrule
\end{tabular}
}%
\end{table}

\begin{table}[htbp]
\caption{Economic value (\$USD) of \dataset by domain and regional subsets.}
\label{tab:eco_values}
\centering
\small
\setlength{\tabcolsep}{5pt}
\begin{tabular}{l c l c c c c}
\toprule
\textbf{Domain} & \textbf{Count} & \textbf{Subset}  & \textbf{Avg. Time (h)} & \textbf{Avg. Wage (/h)} & \textbf{Avg. Value/Q} & \textbf{Total Value} \\
\midrule

\multirow{2}{*}{Finance} & \multirow{2}{*}{80} & CN ({\textyen})        & 23.5 & 312.6 & 7,410.8 & 296,432 \\
                         &                      & Global (\$)   & 26.1 & 175.7 & 4,593.2 & 183,726 \\
\addlinespace

\multirow{2}{*}{Law}     & \multirow{2}{*}{80} & CN ({\textyen})        & 26.2 & 180.4 & 4,722.5 & 188,901 \\
                         &                      & Global (\$)   & 31.4 & 243.0 & 7,666.1 & 306,646 \\
\addlinespace

\multirow{2}{*}{Healthcare} & \multirow{2}{*}{80} & CN ({\textyen})      & 22.9 & 189.9 & 4,379.9 & 175,197 \\
                            &                      & Global (\$) & 22.9 & 354.9 & 8,188.9 & 327,557 \\
\addlinespace

\multirow{2}{*}{Natural Science} & \multirow{2}{*}{80} & CN ({\textyen})   & 22.4 & 118.5 & 2,687.3 & 107,493 \\
                                &                      & Global (\$) & 22.4 & 84.6  & 1,907.0 & 76,280  \\
\addlinespace

\multirow{2}{*}{Industry} & \multirow{2}{*}{80} & CN ({\textyen})      & 24.5 & 152.1 & 3,845.2 & 153,809 \\
                          &                      & Global (\$) & 24.5 & 114.9 & 2,854.0 & 114,161 \\
\midrule
\multicolumn{5}{l}{Total (Total Value)} & \multicolumn{1}{r}{CN ({\textyen})} & \textbf{921,832} \\
\multicolumn{5}{l}{}                    & \multicolumn{1}{r}{Global (\$)}     & \textbf{1,008,370} \\

\bottomrule
\end{tabular}
\end{table}
\section{Case Examples}\label{case_example}

\subsection{Economics and Finance}
{\small

\begin{CaseBox}
\textbf{Background:} \\
Today is December 20, 2025. Yesterday (December 19), the Bank of Japan (BoJ) announced an interest rate hike during its Monetary Policy Meeting, raising the policy rate from 0.50\% to 0.75\%. However, market performance was unexpected: the Japanese Yen did not strengthen following the hike; instead, it depreciated significantly from around 155 prior to the meeting, rapidly breaching the 157 mark. Looking back at the entirety of 2025, the Yen exchange rate experienced multiple rounds of volatility. Despite the Bank of Japan releasing hawkish signals multiple times throughout the year and Japan's current account surplus remaining high, the Yen failed to enter a sustained appreciation trend. \par
\leavevmode

\textbf{Question:} \\
Task Requirements: \par
Review the phased characteristics of the Yen's depreciation cycle throughout 2025, and attempt to delineate the volatility cycles based on key policy milestones and macroeconomic data. \par
Conduct a deep analysis of why the Yen exchange rate exhibited a snapback depreciation and fell below 157 after yesterday's (December 19) rate hike to 0.75\%---a thirty-year high---and identify the underlying market logic. \par
Research and integrate changes in Japan's current account structure in 2025 (specifically the primary income surplus and the services trade deficit) to analyze their long-term suppressive effect on the Yen exchange rate. \par
Provide a strategic outlook for the Yen's trend in 2026 and identify core variables that may trigger a trend reversion or trend reversal for the Yen. \par

\BlackBar{Rubrics}

\textbf{1. Analytical Reasoning} \\
(+8) Divide the 2025 Yen trend into at least 3 distinct phases, with each phase containing: (1) time interval, (2) dominant driving factors, (3) characteristics of exchange rate movement. \\
(+6) Accurately cite at least 2 of the following core data points: (1) policy rate hike from 0.50\% to 0.75\%, (2) pre-meeting exchange rate approx. 155, (3) post-meeting rate breaching 157, (4) this represents a thirty-year high. \\
(+9) Explicitly state that the real interest rate remains negative after the Japanese rate hike, and provide at least one of the following: (1) specific calculation (e.g., 0.75\% - 3.0\% $\approx$ -2.25\%), (2) comparison with U.S. real interest rates, (3) explanation of why negative real interest rates fail to support the exchange rate. \\
(+7) Analyze the dovish signals from BoJ Governor Kazuo Ueda's post-meeting remarks, including at least 2 of the following: (1) emphasis that ``financial conditions remain accommodative,'' (2) lack of a clear path for future hikes, (3) market interpretation as ``buy the rumor, sell the fact.'' \\
(+8) Explain the behavior of Carry Trades post-December 19, including: (1) the low-cost characteristic of the Yen as a funding currency, (2) the logic of arbitrage position re-entry after the hike, (3) mention of strengthening cross pairs (e.g., EUR/JPY, AUD/JPY). \\
(+8) Point out the structural weaknesses in the composition of Japan's current account surplus, encompassing the following 2 points: (1) primary income surplus is largely reinvested overseas, creating no demand for FX conversion, (2) services trade deficit (especially the digital services deficit) constitutes structural, inelastic Yen-selling flow. \\
(+8) Provide specific data on Japan's digital services deficit (e.g., approx. 6--6.6 trillion Yen), or detail its composition (cloud services, advertising, software licensing, etc.). \\
(+8) Analyze the short-term negative shock of capital repatriation, noting: (1) rising JGB yields attract Japanese investors to reduce holdings of overseas assets, (2) this process causes a short-term increase in Yen supply and a demand mismatch, (3) it only supports the Yen in the medium-to-long term. \\
(-7) Heavy reliance on technical analysis terminology (e.g., Fibonacci retracement, moving average golden cross, RSI oversold) while lacking fundamental logic support. \\
(+7) When analyzing the different phases of the Yen trend, continuously compare the U.S.--Japan real interest rate spread; the method must include: (1) Japan real rate formula: policy rate - core CPI, (2) U.S. real rate formula: nominal rate (10-year Treasury or Fed Funds Rate) - core PCE/CPI, (3) explicitly state the direction (positive for the U.S., negative for Japan) and magnitude (hundreds of basis points) of the spread. \\
(+6) List key BoJ meeting dates and resolutions in 2025, such as: Jan 24 (hike to 0.5\%), Jun 17 (JGB purchase tapering path), Sep 19 (ETF disposition), Oct 30 (split vote), Dec 19 (hike to 0.75\%). \\
(+6) Cite Shunto wage growth data for 2024 and 2025 are $>5\%$, and conclude based on this that while 2026 wage growth will remain historically high, it will fall below 5\% due to the marginal slowdown in inflation. \\
(+5) Analyze how Japan's aging population affects long-term capital flow patterns, noting: aging $\to$ pension funds liquidating overseas assets $\to$ structural shift from ``capital exporting nation'' to ``capital repatriating nation.'' \\
(-4) Heavy use of second-hand phrasing such as ``reportedly,'' ``market rumors,'' or ``analysts expect,'' without direct citation of official BoJ documents, Ministry of Finance statements, or primary economic data reports. \\
(-5) Failure to analyze the deeper impact of the 2 votes supporting a hike (7--2 split) during the October BoJ meeting on market expectations and central bank credibility (e.g., ``split vote increases policy uncertainty,'' ``implies rising internal hawkish power''). \\
(-5) Simplistically attributing the post-hike depreciation to ``market priced in'' or ``buy the rumor, sell the fact'' without deeply analyzing multiple mechanisms such as real interest rates, policy credibility, and capital flows. \\

\end{CaseBox}

\begin{CaseBox}
\BlackBar{Response}

You referenced ``today = Dec 20, 2025,'' but as of my system date Feb 18, 2026, the Dec 19, 2025 BoJ meeting is a past event. Below I treat 2025 as a completed year, and the Dec 19, 2025 hike (0.50\% $\to$ 0.75\%) as the late-cycle shock you described. (investing.com) \par

\leavevmode

\textbf{1) USD/JPY in 2025: phased volatility / ``depreciation-cycle'' mapping by milestones} \par
A useful way to segment 2025 is: (A) early-year USD stress $\to$ JPY rally, then (B) mid/late-year re-widening of ``JPY funding'' logic as BoJ normalization stayed slow and Japan's services outflows remained structural. \par

\leavevmode

\textbf{Phase 1 --- Jan: ``BoJ hikes, but JPY still a funding currency''} \par
\textbf{Policy milestone:} Jan 24, 2025 BoJ hiked 0.25\% $\to$ 0.50\%. (apnews.com) \par
\textbf{Market character:} Brief JPY support, but the market quickly reverted to rate-differential + carry framing (JPY still low yield vs. G10). \par

\leavevmode

\textbf{Phase 2 --- Late Mar--Apr: ``Tariff/Fed-independence risk $\to$ USD de-risking $\to$ JPY safe-haven bid''} \par
\textbf{Macro/policy catalyst:} Escalation of US tariff / policy uncertainty and Fed independence concerns triggered ``sell USD'' hedging. \par
\textbf{Price action marker:} USD/JPY broke $\sim$140 (JPY strongest area of 2025). (bloomberg.com) \par
\textbf{Interpretation:} This was less ``Japan strength'' and more global USD risk premium rising; JPY benefited as a liquid hedge. \par

\leavevmode

\textbf{Phase 3 --- May--Aug: ``Mean reversion into a range; JPY becomes tethered to Fed-cut pricing''} \par
\textbf{Macro marker:} As risk sentiment stabilized intermittently, carry demand returned and USD/JPY drifted back up from the April lows. \par
\textbf{Range framing:} Research commentary described range trading and high sensitivity to Fed cut expectations rather than Japan's own data. (mufgresearch.com) \par

\leavevmode

\textbf{Phase 4 --- Sep--Oct: ``BoJ holds at 0.50\% (dissents), market punishes slow normalization''} \par
\textbf{Policy milestone:} BoJ repeatedly held at $\sim$0.50\%, with dissenters favoring hikes. \par
\textbf{Price action marker:} After one such hold, the yen weakened sharply (example: $\sim$153 per USD reported). (ft.com) \par
\textbf{Interpretation:} The market concluded the BoJ reaction function was still highly asymmetric (very gradual hikes, cautious guidance). \par

\leavevmode

\textbf{Phase 5 --- Dec: ``Hike delivered, but guidance disappoints $\to$ snapback depreciation''} \par
\textbf{Policy milestone:} Dec 19, 2025 BoJ hiked 0.50\% $\to$ 0.75\% (30-year high). (apnews.com) \par
\textbf{Price action marker:} Yen weakened; USD/JPY spiked through 157 (Reuters cited $\sim$157.67 intraday). (investing.com) \par

\textbf{Important contextual point:} Japan's MoF \textbf{did not support JPY with actual intervention operations during 2025} (multiple monthly windows show JPY 0). That removed a key ``hard downside limiter'' for USD/JPY breakouts. (mof.go.jp) \par

\leavevmode

\textbf{2) Why USD/JPY snapbacked above 157 after the Dec 19 hike: the market logic} \par
The apparent paradox (``hawkish action, weaker currency'') is common when the marginal information in the decision is less hawkish than positioning implied. \par

\textbf{(1) Priced-in hike + sell the fact} \par
Reuters explicitly noted the move was well-telegraphed, and selling intensified after Ueda stayed vague on timing/pace of future hikes. (investing.com) \par
Mechanics: pre-meeting investors build JPY-long / USDJPY-short expectations; post-meeting if the statement/presser does not upgrade the expected path, those longs unwind fast. \par

\textbf{(2) The path matters more than the level: forward rates did not reprice hawkishly enough} \par
Multiple reports emphasized the BoJ was unclear about the ``neutral destination'' and still described conditions as accommodative / real rates deeply negative. (barrons.com) \par
So the market read the hike as one step, not a regime of fast normalization, leaving terminal BoJ rates and speed lower than hoped by yen bulls. \par

\textbf{(3) US--Japan rate differentials still favored USD even after Fed cuts} \par
After the Fed's Dec 2025 decision, the fed funds target range was 3.50\%--3.75\%, far above 0.75\% in Japan. (federalreserve.gov) \par
That spread keeps the carry math attractive unless the market believes the BoJ will keep hiking or the Fed will cut faster. \par

\textbf{(4) Cross-yen dynamics + risk-on tone} \par
Reuters reported not only USDJPY strength but also record/major highs in EUR/JPY, CHF/JPY, GBP/JPY on the day---classic broad JPY selling rather than a USD-only story. (investing.com) \par
This usually aligns with risk assets holding up and investors maintaining/adding JPY-funded exposures. \par

\textbf{(5) Fiscal risk channel: higher yields can be ``bad JPY'' if interpreted as supply/fiscal stress} \par
FT coverage around the post-hike period highlighted concerns about bond yields and future issuance tied to stimulus, i.e., a risk premium problem rather than a growth-positive normalization. (ft.com) \par

\leavevmode

\textbf{3) Japan's 2025 current-account structure shifts---and why they can suppress JPY long-term} \par

\textbf{2025 CY headline structure (official BoP)} \par
For calendar year 2025 (preliminary), Japan ran: \par
\begin{itemize}[leftmargin=1.2em]
  \item Current account surplus: JPY 31.8799 tn
  \item Primary income surplus: JPY 41.5903 tn
  \item Services balance: -JPY 3.3928 tn
  \item Goods balance: -JPY 0.8487 tn (deficit narrowed a lot vs 2024)
\end{itemize}
(mof.go.jp) \par

So you are right that ``current account is strong,'' but the FX impact depends on conversion behavior and offsetting structural outflows. \par

\textbf{(A) Primary income surplus $\neq$ immediate spot JPY buying} \par
The primary income table shows large components that often do not translate into immediate repatriation: big direct investment income, including reinvested earnings (kept abroad by corporates). (mof.go.jp) \par
FX implication: Even if Japan earns huge overseas income, if it is retained overseas, reinvested, or hedged, it does not create the spot-market yen demand that would force sustained appreciation. \par

\textbf{(B) Services deficit widened, and its composition matters} \par
The services deficit deepened in 2025 (-JPY 3.3928 tn). (mof.go.jp) \par
Inside services, 2025 shows: travel surplus +JPY 6.3429 tn (strong inbound tourism support), but large structural deficits in ``modern'' services, notably telecom/computer/info services (-JPY 2.4740 tn) and other business services (-JPY 5.9443 tn), plus insurance \& pension services (-JPY 3.3137 tn). (mof.go.jp) \par
Why this can suppress JPY over time: These service categories are recurring payments (subscriptions, cloud, advertising, consulting, reinsurance, etc.). They are less sensitive to short-term FX moves and can behave like a structural demand for foreign currency, diluting the ``surplus $\Rightarrow$ stronger currency'' intuition. \par

\leavevmode

\textbf{4) Strategic outlook for 2026: what could trigger reversal vs reversion?} \par

\textbf{Where the market started 2026} \par
USD/JPY in early 2026 traded roughly 152--159, and was around the low-153 area in mid-Feb 2026 (per published daily history). (poundsterlinglive.com) \par

\textbf{Base case (range-to-lower USD/JPY; gradual JPY recovery)} \par
A reasonable base case is JPY modestly stronger over 2026 if: (1) BoJ keeps hiking (or credibly signals a path), not just one-off moves, (2) the Fed continues easing (narrowing the US--Japan spread), and (3) global risk sentiment is not persistently risk-on carry. \par
Recent reporting suggested BoJ signaled additional hikes likely in 2026 even when holding 0.75\%. (ft.com) \par

\textbf{What would trigger a trend reversion (JPY strengthening sustainably)} \par
Core variables to watch: \par
\begin{itemize}[leftmargin=1.2em]
  \item BoJ reaction function turns symmetric: clear guidance on pace and a believable estimate of ``neutral'' (reduces the chance of another ``hike but dovish'' selloff). (barrons.com)
  \item Japan wage/inflation persistence (Shunto + services inflation): if wage growth sustains and BoJ can hike without choking demand, JPY can re-rate higher.
  \item Fed cuts faster than currently discounted: a sharper US slowdown compressing the front-end spread is the cleanest ``JPY bull'' macro setup. (federalreserve.gov)
  \item Risk-off / deleveraging shock: any global volatility event that forces carry unwind tends to produce abrupt JPY strength (as in April 2025's USD de-risking episode). (bloomberg.com)
  \item Meaningful improvement in the services deficit: e.g., sustained shrinkage in digital/business-services outflows (hard, but it is the structural piece).
\end{itemize}

\textbf{What would trigger a trend reversal against JPY (USD/JPY back toward highs)} \par
\begin{itemize}[leftmargin=1.2em]
  \item BoJ slows/pauses again due to growth or political pressure (market re-anchors on ``JPY funding'').
  \item Japan fiscal stress narrative rises (issuance fears $\to$ higher yields but wider risk premium) rather than ``healthy normalization.'' (ft.com)
  \item Persistent risk-on environment (carry demand stays dominant).
  \item No intervention deterrent even during sharp moves (2025 had JPY 0 operations in multiple windows). (mof.go.jp)
\end{itemize}

If you want, I can turn the 2025 phase section into a timeline table (date $\to$ event $\to$ USD/JPY reaction $\to$ which flow dominated: rates / risk / BoP / positioning) and tailor it to the exact USD/JPY series you use (spot close, Tokyo close, or intraday highs/lows). \par

\end{CaseBox}

}

\subsection{Law}
{\small

\begin{CaseBox}
\textbf{Background:} \\
Case Background: \par
A company incorporated in the United States (hereinafter referred to as the ``Plaintiff'') and a company incorporated in Singapore (hereinafter referred to as the ``Defendant'') entered into a commodity purchase and sale contract in 2024. The Plaintiff acted as the Supplier and the Defendant as the Purchaser. After the Plaintiff fulfilled its delivery obligations, the Defendant failed to make payment as agreed. Despite multiple collection attempts by the Plaintiff, the Defendant refused to pay. Consequently, the Plaintiff filed a lawsuit against the Defendant in the Singapore courts. \par
Upon receiving the notice of action, the Defendant, having failed to retain a copy of the contract, requested a copy of the contract for review from the Plaintiff. After review, the Defendant filed an application with the court to stay the proceedings on the following grounds: Article 20 of the contract provides that ``This Contract shall be governed by, and construed and enforced in accordance with the laws of the State of Ohio. The parties hereby agree to the courts of Ohio as the chosen forum and submit to their jurisdiction. Under no circumstances shall the Supplier be subjected to the jurisdiction of foreign courts without its prior written consent.'' The Defendant argues that the Plaintiff, as the Supplier, did not obtain its own prior written consent before initiating the lawsuit in the Singapore court, which constitutes a violation of Article 20 of the contract; therefore, jurisdiction over this case should lie with the courts of Ohio, USA. \par
The Plaintiff rejects the Defendant's opinion and insists on proceeding with the litigation in Singapore. \par
The Defendant further invokes the doctrine of forum non conveniens, arguing that since the contract chooses the application of the laws of the State of Ohio, it is evident that the Singapore court is a forum non conveniens and should decide not to hear this case. \par
\leavevmode

\textbf{Question:} \\
Acting as the Presiding Judge of the Singapore court in this case, you are required to adjudicate whether this case should continue to be heard in Singapore. Please focus your argumentation on the following issues: \par
1. Is the Plaintiff required to obtain its own written consent? \par
2. Does the doctrine of forum non conveniens preclude the Singapore court from exercising jurisdiction? \par
3. Are there any other facts and reasons to support your analysis? \par

\BlackBar{Rubrics}

\textbf{1. Factual Information} \\
(+2) Correctly distinguished the procedural status of both parties in the response, specifically identifying the Plaintiff as a US company and the Defendant as a Singaporean company. \\[2mm]

\textbf{2. Analytical Reasoning} \\
(+3) Explicitly provided a general conclusion of ``dismissing the Defendant's application for a stay of proceedings'' or ``continuing the trial in Singapore'' at the beginning or end of the response. \\
(+5) Comprehensively enumerated the three core components of Article 20 of the contract (governing law, choice of forum, protection of the Supplier), concluding that Article 20 constitutes a permissive jurisdiction clause rather than an exclusive jurisdiction clause. \\
(+3) Determined that the literal meaning and purpose of Article 20 are to protect the unilateral interest of the Supplier (Plaintiff) in choosing the forum, i.e., determining the clause as an exclusive right conferred upon the Supplier. \\
(+3) Pointed out that the Plaintiff's act of filing a lawsuit in Singapore itself constitutes a waiver of the right requiring ``its own written consent''. \\
(+7) Clarified the definition of the doctrine of forum non conveniens, comprising core elements: the existence of a competent court, that trying the case would cause inconvenience to the parties and the judicial system (or fail to ensure the fair administration of justice), and the existence of a more appropriate alternative forum. \\
(-6) In analyzing the applicability of the Hague Convention on Choice of Court Agreements or choice of court agreements, erroneously stated the contracting states, specifically by claiming that Singapore is not a party to the Hague Convention on Choice of Court Agreements, or that the United States (Ohio) is a party thereto. \\
(-3) Failed to distinguish the independence of governing law from venue/forum, supporting the forum non conveniens principle solely on the grounds of ``applying Ohio law''. \\
(+7) Elucidated that both Singapore and Ohio belong to the common law system; therefore, applying Ohio law presents no substantive difficulty, and the parties have not argued that the application of different legal principles by Ohio courts versus Singapore courts would substantively affect the dispute. \\
(+3) Analyzed that adjudicating in Singapore is more conducive to the enforcement of the judgment, given that the Defendant's assets are likely located in Singapore. \\
(+3) Pointed out that the mere fact that the lex causae (governing law) is a foreign law (Ohio law) is insufficient to constitute grounds for applying the doctrine of forum non conveniens. \\
(+5) Noted the fact that the Defendant did not retain a copy of the contract to infer that Article 20 was not subject to close negotiation between the parties, thereby reducing the influence or weight of said clause. \\
(+4) Clarified that the Singapore court has personal jurisdiction (in personam jurisdiction) over the Defendant. \\
(-2) Erroneously allocated the burden of proof regarding the doctrine of forum non conveniens, arguing that the Plaintiff should prove that Singapore is the appropriate forum. \\[2mm]

\textbf{3. Instructions Following} \\
(+3) Adopted the identity and tone of a Presiding Judge of the Singapore Court for the judicial reasoning; the narrative perspective should be that of the judge (using ``I'', ``the Court'', or avoiding self-reference without using a researcher or third-party tone). \\
(-2) When discussing ``other facts and reasons for argument,'' completely repeated the arguments used in answering the two questions: ``whether the Plaintiff must obtain its own written consent'' and ``whether the forum non conveniens doctrine hinders jurisdiction'' in the previous sections. \\[2mm]

\textbf{4. Structure and Formatting} \\
(+2) The structure of the response must at least contain ``General Conclusion, Issue 1, Issue 2, Issue 3'', where Issues 1, 2, and 3 refer to the analysis of whether the Plaintiff must obtain its own written consent, whether the doctrine of forum non conveniens impedes Singaporean jurisdiction, and whether there are other supporting facts and reasons. \\

\end{CaseBox}

\begin{CaseBox}
\BlackBar{Gemini-3.0-Pro Response}

\textbf{JUDGMENT} \par
\textbf{Case:} Plaintiff (US Supplier) v. Defendant (Singapore Purchaser) \\
\textbf{Forum:} High Court of the Republic of Singapore \\
\textbf{Presiding Judge:} [Assigned Judge] \par
\leavevmode

\textbf{DECISION} \par
The Defendant's application to stay the proceedings is \textbf{dismissed}. This Court retains jurisdiction, and the case shall proceed in Singapore. The reasons for this decision, addressing the specific issues raised regarding the interpretation of the jurisdiction clause and the doctrine of forum non conveniens, are set out below. \par

\leavevmode

\textbf{General Conclusion} \\
The Defendant's application is dismissed. The action will continue in Singapore. \par

\leavevmode

\textbf{Issue 1. Is the Plaintiff required to obtain its own written consent?} \par

\leavevmode

\textbf{No. The Defendant's interpretation of Article 20 is commercially absurd and legally untenable.} \par
The Defendant argues that the Plaintiff (the Supplier) breached Article 20 by failing to obtain its own ``prior written consent'' before filing suit in Singapore. This Court rejects this submission for the following reasons: \par
\begin{itemize}[leftmargin=1.2em]
  \item \textbf{Nature of the clause (benefit of the party):} The sentence in Article 20 stating, ``Under no circumstances shall the Supplier be subjected to the jurisdiction of foreign courts without its prior written consent,'' is an asymmetrical jurisdiction clause designed exclusively for the benefit of the Plaintiff (the Supplier). Its purpose is to protect the Supplier from being dragged into a foreign jurisdiction against its will by the Purchaser. It acts as a shield for the Supplier, not a shackle. \par
  \item \textbf{Waiver by conduct:} A party for whose benefit a contractual provision is inserted is entitled to waive that benefit. By voluntarily commencing proceedings in the Singapore courts, the Plaintiff has unequivocally waived its right to object to this forum. \par
  \item \textbf{Absurdity of ``self-consent'':} It is a logical fallacy to suggest that a legal entity must issue a formal written document to itself granting permission to take an action it has voluntarily decided to undertake. The act of filing the Writ of Summons and Statement of Claim constitutes the clearest possible evidence of the Plaintiff's consent to submit to the jurisdiction of the Singapore courts. To require a separate internal written memo would be an exercise in futility and form over substance. \par
\end{itemize}
Therefore, the Plaintiff was not required to generate a prior written consent to itself to validate the proceedings. \par

\leavevmode

\textbf{Issue 2. Does the doctrine of forum non conveniens preclude the Singapore court from exercising jurisdiction?} \par
\textbf{No. The Singapore Court is the appropriate forum for this dispute.} \par
The Defendant invokes the doctrine of forum non conveniens (based on the principles in \textit{Spiliada Maritime Corp v Cansulex Ltd}, applicable in Singapore), arguing that the choice of Ohio law makes Singapore an inappropriate forum. This argument fails for the following reasons: \par
\begin{itemize}[leftmargin=1.2em]
  \item \textbf{Distinction between choice of law and choice of forum:} The fact that the contract is governed by the laws of the State of Ohio does not automatically render the courts of Ohio the only appropriate forum. It is a common occurrence for Singapore courts to adjudicate disputes governed by foreign laws. The choice of law is a factor to be considered, but it is not decisive. \par
  \item \textbf{Nature of the dispute:} This is a straightforward debt collection case arising from a commodity sale where delivery has occurred, and payment is outstanding. The legal issues likely revolve around contractual performance and payment obligations, which are not so uniquely complex under Ohio law that a Singapore court cannot adjudicate them (potentially with the aid of expert evidence on Ohio law, if strictly necessary). \par
  \item \textbf{Connecting factors:} The Defendant is a company incorporated in Singapore. As such, Singapore is the Defendant's home jurisdiction. The Defendant is clearly present here, and the Singapore courts have personal jurisdiction over the Defendant. Therefore, Singapore is a prima facie natural forum for the dispute. \par
\end{itemize}
Consequently, the Defendant has failed to demonstrate that Ohio is a ``clearly more appropriate'' forum than Singapore for the interests of all parties and the ends of justice. \par

\leavevmode

\textbf{Issue 3. Other facts and reasons supporting this analysis} \par
In addition to the points above, the Court relies on the following factors to deny the stay: \par
\begin{itemize}[leftmargin=1.2em]
  \item \textbf{Enforceability and efficiency:} The ultimate goal of litigation is to obtain an enforceable judgment. Since the Defendant is a Singapore-incorporated entity, its assets are presumably located in Singapore. If this Court were to stay proceedings in favor of Ohio, the Plaintiff would have to litigate in the US, obtain a judgment there, and then return to Singapore to register and enforce that foreign judgment. This would result in unnecessary circularity, delay, and expense. Hearing the case in Singapore allows for a direct judgment and immediate enforcement, serving the ends of justice and judicial economy. \par
  \item \textbf{Burden of proof:} In a case where the Plaintiff sues the Defendant in the Defendant's own jurisdiction (Singapore), the burden on the Defendant to show that the forum is inappropriate is particularly high. The Defendant has offered no evidence regarding the location of witnesses or evidence that would make a trial in Ohio significantly more convenient. \par
  \item \textbf{Interpretation of ``courts of Ohio'':} While Article 20 mentions the parties agree to Ohio courts, it does not use the word ``exclusive'' regarding the Plaintiff's obligations. When combined with the ``consent'' proviso, the clause is best interpreted as: (a) exclusive jurisdiction for suits against the Supplier (unless it consents otherwise), and (b) non-exclusive jurisdiction for suits by the Supplier against the Purchaser. The Plaintiff has exercised its option to sue in the Defendant's domicile. \par
\end{itemize}

\textbf{Conclusion} \par
The application for a stay of proceedings is dismissed with costs awarded to the Plaintiff. The Defendant is ordered to file its Defence within the statutory timeline. \par

\end{CaseBox}

}

\subsection{Healthcare and Medicine}
{\small

\begin{CaseBox}
\textbf{Background:} \\
You are an orthodontist. You have received a 10-year-old female patient whose parents' chief complaint is dental misalignment and ``protrusive maxillary incisors'' (buck teeth), which have gradually worsened over the past 2 years. There is no family history of similar facial features. \par
Upon taking the medical history, it was found that the patient has a habit of nocturnal mouth breathing. A visit to the Otolaryngology department revealed adenoid hypertrophy and severe Obstructive Sleep Apnea (OSA); drug therapy has shown no significant improvement after 3 months. \par
Extraoral examination reveals a convex profile, maxillary protrusion, mandibular retrusion, and a long lower facial third. Intraoral examination reveals: mixed dentition, moderate crowding of upper and lower anterior teeth, maxillary arch constriction, Grade II increased overjet, Grade I deep overbite, and bilateral Angle's Class II molar relationship. \par
Cephalometric analysis shows a distinct concavity on the lower border of the C2 vertebral body, while the lower borders of C3 and C4 are flat. Measurements indicate SNA$=80^\circ$, SNB$=74^\circ$, SN--MP$=32^\circ$, U1--PP$=120^\circ$, and L1--MP$=90^\circ$. \par
After researching online and consulting other clinics, the family has presented the following demands:
\begin{itemize}[leftmargin=1.2em]
  \item Non-extraction treatment only; they claim extraction will cause teeth to become spaced, mobile, and prone to loss later in life, affecting eating;
  \item No adenoidectomy, as they believe it will affect immunity;
  \item Request to wear silicone-based appliances, strictly for night use only, so as not to affect daytime study.
\end{itemize}

\leavevmode

\textbf{Question:} \\
Please provide professional treatment recommendations based on the above information and incorporating the latest clinical guidelines or expert consensus from 2024--2025.

\BlackBar{Rubrics}

\textbf{1. Factual Information} \\
(+5) Diagnose Angle Class II malocclusion based on the distal molar relationship. \\
(+5) Diagnose skeletal Class II malocclusion based on SNA, SNB, and ANB values. \\
(+7) Determine that development is at the CS2 stage based on cervical vertebral morphology, indicating the patient is prior to the peak growth spurt. \\
(+5) Recommend adenoidectomy based on severe Obstructive Sleep Apnea (OSA), inefficacy of drug therapy, and the presence of mouth breathing habits. \\
(+5) Cite the ``Expert Consensus on the Clinical Application of Silicone Oral Functional Trainers (2024)'' to recommend against the use of silicone-based appliances. \\[2mm]

\textbf{2. Analytical Reasoning} \\
(+5) Analyze and identify mouth breathing caused by adenoid hypertrophy as the primary etiology of the patient's dentofacial deformity. \\
(+6) Propose maxillary expansion therapy to resolve the maxillary arch constriction. \\
(+7) Recommend using a mandibular advancement appliance to guide mandibular growth, with Twin Block appliance being the preferred choice. \\
(+5) Recommend full-day wear for the functional appliance, refuting the parents' request for night-only wear, and explain the principles of muscular, skeletal, and dental adaptation. \\
(+5) Propose myofunctional training (e.g., tongue muscle, orbicularis oris muscle training). \\
(+5) Explain the safety of extraction treatment, clarifying that spaces will close and properly executed orthodontic treatment will not cause pathologic tooth mobility or tooth loss. \\
(-5) During expansion therapy, failing to combine a Twin Block appliance with a palatal expander (e.g., jackscrew) for simultaneous expansion and mandibular advancement, and limiting treatment to a single expansion method (rapid/slow) or a single type of expander. \\
(-8) Yielding to the family's unreasonable demands by considering the use of silicone-based appliances for treatment. \\[2mm]

\textbf{3. Instructions Following} \\
(+7) Address the parents' refusal of surgery by pointing out that orthodontic treatment is only suitable after the mouth breathing habit is resolved. \\
(+3) Address the parents' concerns about extraction by explaining the evolutionary trend leading to insufficient bone mass relative to tooth mass. \\
(+5) Explicitly recommend against silicone-based appliances, highlighting disadvantages such as insufficient hardness, poor retention, and lack of skeletal remodeling effects. \\[2mm]

\textbf{4. Structure and Formatting} \\
(+8) Structure the orthodontic treatment plan into two stages: Phase I functional treatment and Phase II finishing and detailing. \\
(-5) The response contains 3 or more errors in the use or translation of professional terminology. \\
(-5) The response uses large, dense blocks of pure text lacking clear subheadings or bullet points, resulting in poor readability. \\[2mm]

\textbf{5. Others} \\
(-7) Omitting the disclaimer that AI only provides suggestions and offline medical consultation is required.

\end{CaseBox}

\begin{CaseBox}
\BlackBar{GPT-5.2-High-Search Response}

\textbf{1) Problem list \& priorities (what matters most first)} \\
\leavevmode

This 10-year-old has (a) severe pediatric OSA with adenoid hypertrophy and chronic mouth breathing, plus (b) skeletal/dental Class II div 1 (mandibular retrusion with proclined maxillary incisors, increased overjet), (c) transverse maxillary constriction, (d) moderate anterior crowding, and a vertical tendency/long lower facial third. \par
From a medical and risk-management standpoint, severe OSA is the first priority because untreated pediatric OSA is linked to neurocognitive/behavioral, cardiovascular and quality-of-life consequences; orthodontics should be coordinated around airway care, not used as a substitute. \par
Your cervical vertebrae maturation finding (concavity only at C2; C3--C4 flat) is consistent with early CVM (pre-peak growth), meaning some growth modification can start now, but the most efficient mandibular ``growth spurt'' window is usually a bit later (often when C3 also shows concavity). This supports a staged (Phase I / Phase II) plan. \par

\leavevmode

\textbf{2) Airway / OSA recommendations (based on 2024--2025 guidance)} \\
\leavevmode

\textbf{A. Re-counsel: surgery is standard first-line for hypertrophy-driven pediatric OSA} \\
Major pediatric guidance continues to recommend adenotonsillar surgery as first-line when hypertrophy is present, and CPAP when surgery is not performed or is ineffective. (publications.aap.org) \\
For severe OSA, ENT guidelines also emphasize careful peri-operative planning/monitoring and that OSA can persist and needs follow-up. (aafp.org) \par
Clinical recommendation: I would formally recommend re-discussion with ENT + pediatric sleep medicine about adenoidectomy (often with tonsil evaluation) as the evidence-based first-line option, with a plan for post-treatment reassessment (repeat PSG or objective testing). \par

\leavevmode

\textbf{B. If the family refuses adenoidectomy: CPAP is the medically appropriate fallback} \\
If adenoidectomy is declined, then CPAP is the most evidence-based non-surgical therapy for significant OSA (though adherence can be difficult). This ``CPAP if surgery not performed'' pathway is explicitly recommended in pediatric guidance. (publications.aap.org) \par
Clinical recommendation: If no surgery, I would refer for CPAP initiation/titration and adherence coaching before relying on orthodontic appliances for an airway problem. \par

\leavevmode

\textbf{C. Orthodontics can be an adjunct for selected airway anatomy (maxillary constriction)} \\
The 2024 American Thoracic Society (ATS) guideline on persistent pediatric OSA states that children with persistent OSA and specific craniofacial features may be candidates for orthodontic/dentofacial orthopedic treatment, and the panel specifically notes that RME is preferred when there is a constricted maxilla (high/narrow palate, often posterior crossbite). (pmc.ncbi.nlm.nih.gov) \par
Also, a 2025 randomized controlled trial comparing RME vs adenotonsillectomy found both improved PSG parameters (including AHI), while adenotonsillectomy produced better symptom/QOL improvement. (pubmed.ncbi.nlm.nih.gov) \par
So: RME is reasonable for this child's narrow maxilla and may help airway metrics---but it is not a guarantee, and for severe OSA it should be adjunctive, not the sole treatment. \par

\leavevmode

\textbf{3) Orthodontic plan (what I would recommend clinically)} \\
\leavevmode

\textbf{Phase I (now; mixed dentition, narrow maxilla, trauma-risk overjet)} \\
\leavevmode

\textbf{1. Rapid Maxillary Expansion (RME) with a fixed expander (Hyrax/Haas-type).} \\
Goals: correct constriction, gain arch width/perimeter to help crowding, improve nasal airflow potential, and create a better foundation for Class II correction. \\
Important: RME is not ``night-only.'' It must be worn 24/7 because it is cemented and works by orthopedic forces over days--weeks, then retained to stabilize. \par

\leavevmode

\textbf{2. Incisor/overjet risk reduction} \\
Because protrusive incisors raise the risk of trauma, I would usually add either:
\begin{itemize}[leftmargin=1.2em]
  \item a short ``$2\times4$'' partial fixed appliance to align/level incisors and manage deep bite, and/or
  \item a functional Class II approach once transverse issues are controlled (see below).
\end{itemize}

\leavevmode

\textbf{3. Class II correction (mandibular retrusion) -- realistic options}
\begin{itemize}[leftmargin=1.2em]
  \item Most predictable: a fixed functional appliance (e.g., Herbst-type) later in Phase I/early Phase II---minimal daytime compliance burden.
  \item Removable functional (Twin-Block-type): workable if worn enough hours.
\end{itemize}
Evidence on wear time is mixed; an RCT found part-time wear ($\sim$9 h/day) could achieve similar overjet reduction to full-time in 12 months (pubmed.ncbi.nlm.nih.gov), but other work using compliance indicators suggests greater skeletal effect with longer daily wear (pubmed.ncbi.nlm.nih.gov). \\
Bottom line: ``night-only'' may improve overjet some, but it is less reliable for meaningful skeletal correction and stability. \par

\leavevmode

\textbf{Phase II (permanent dentition, around/after growth peak)} \\
Comprehensive braces/aligners as appropriate to:
\begin{itemize}[leftmargin=1.2em]
  \item finalize Class II correction (as needed),
  \item fully relieve crowding and coordinate arches,
  \item optimize incisor inclination and overbite.
\end{itemize}

\leavevmode

\textbf{4) Addressing the family's three demands (with professional boundaries)}

\leavevmode

\textbf{Demand 1: ``Non-extraction only; extractions cause spacing, mobility, tooth loss later.''} \\
Professional response:
\begin{itemize}[leftmargin=1.2em]
  \item Extractions do not inherently cause teeth to become loose and fall out later. Long-term tooth loss risk is driven primarily by caries, periodontal disease, and maintenance, not by having had orthodontic extractions. Temporary mobility during orthodontic movement can occur in any orthodontic treatment (extraction or non-extraction) and usually resolves with retention and healthy periodontium.
  \item Spaces after extraction are a treatment-execution/retention issue, not an inevitable biologic consequence. If spaces are properly closed with correct root positioning and then retained, they are usually stable; if spaces reopen, it is managed orthodontically.
\end{itemize}

\leavevmode

\textbf{Demand 2: ``No adenoidectomy; it will affect immunity.''} \\
Professional response:
\begin{itemize}[leftmargin=1.2em]
  \item There is substantial evidence (including systematic review/meta-analysis and broad reviews) indicating no clinically significant long-term impairment of humoral/cellular immunity after tonsil/adenoid surgery in typical children, and that remaining lymphoid tissue can compensate. (pubmed.ncbi.nlm.nih.gov)
  \item At the same time, there are large observational data suggesting associations between removal and higher long-term rates of some respiratory/allergic/infectious diagnoses---this does not prove causation for an individual child, but it is part of balanced counseling. (jamanetwork.com)
  \item For a child with severe OSA, the known morbidity of leaving OSA untreated is typically considered to outweigh the uncertain/variable long-term infection-risk associations---this is why adenotonsillar surgery remains standard first-line when hypertrophy is present. (publications.aap.org)
\end{itemize}

\leavevmode

\textbf{Demand 3: ``Silicone appliance, night-only, not affecting daytime study.''} \\
They are likely referring to prefabricated silicone myofunctional appliances (e.g., trainer-type).

\leavevmode

\textbf{Non-negotiables (professional boundary)}

\begin{itemize}[leftmargin=1.2em]
  \item I cannot ethically recommend night-only silicone orthodontics as the sole management for severe pediatric OSA.
  \item If adenoidectomy remains refused, I would require documentation that the family understands that CPAP is the medically recommended non-surgical treatment, and that orthodontics is adjunctive. (publications.aap.org)
\end{itemize}

\end{CaseBox}

}

\subsection{Natural science}
{\small

\begin{CaseBox}
\textbf{Background:} \\
A Quantum Spin Liquid (QSL) is a state of matter characterized by the absence of long-range magnetic order and the presence of high degrees of quantum entanglement. Experimental research on Herbertsmithite has revealed that its spin correlation behavior transcends the theoretical framework of simple nearest-neighbor models. Although the ground state of Herbertsmithite does not exhibit traditional long-range magnetic ordering at low temperatures and displays behavior similar to short-range Resonating Valence Bond (RVB) states, no spin gap is observed in $Q$--$\omega$ space. This exotic phenomenon necessitates the consideration that the spin liquid state in Herbertsmithite may possess more complex spin correlation patterns. \par

\leavevmode

\textbf{Question:} \\
1. Referencing results from relevant Inelastic Neutron Scattering (INS) experiments, elucidate the basis for determining that spin correlation behavior in Herbertsmithite exceeds the framework of the nearest-neighbor model. \par
2. Analyze the impact of these spin correlations, which transcend nearest-neighbor interactions, on the properties of the Herbertsmithite ground state spin liquid. \par
3. Combining relevant INS data, discuss why the spin correlation pattern in Herbertsmithite does not conform to the simple short-range Resonating Valence Bond (RVB) model, and propose the challenges this phenomenon poses to theoretical models. \par

\BlackBar{Rubrics}

\textbf{1. Analytical Reasoning} \\
(+5) In INS experiments on Herbertsmithite, the observed spin excitation modes exhibit significant discrepancies with the nearest-neighbor singlet model. For example, the experimental energy-integrated structure factor indicates that the spin correlation behavior in Herbertsmithite demonstrates longer-range correlations. \\
(+10) Model deviation of the equal-time structure factor: INS experiments indicate that spin correlations are not confined to nearest neighbors but exhibit longer-range spatial correlations, leading to a reduction in the diffuse nature of scattering peaks. \\
(+5) Supplementary evidence: Although the experimentally measured equal-time structure factor integrated over energies from 1 to 9 meV shows similarity to calculations based on the uncorrelated nearest-neighbor singlet model, the width of the experimental scattering signal in reciprocal space is significantly narrower. \\
(+4) Mention that at specific momentum positions (e.g., (0,2,0) as described in doi: 10.1038/nature11659), the nearest-neighbor singlet model fails to explain the signal contribution at this location, illustrating the existence of non-nearest-neighbor interactions. \\
(+6) Provide evidence: Using the momentum position labeling convention from doi: 10.1038/nature11659 as a reference, line scans along the Brillouin zone (0,K,0) direction (1--7 meV) show distinct scattering intensity at positions such as (0,2,0). \\
(+4) Explain that in the low-energy interval of 0.25--1 meV, extra broad peaks appear at reciprocal space (1,0,0) (using the momentum position labeling convention from doi: 10.1038/nature11659) and equivalent positions; these are not Bragg peaks and cannot be explained by nearest-neighbor interactions. \\
(-20) The model mistakenly attributes the broad peaks appearing in the low-energy region to interlayer impurity $Cu^{2+}$. \\
(+5) Explain the energy insensitivity of scattering intensity: A similar hexagonal ring scattering pattern is observed across the 1.5--11 meV range in INS experiments. \\
(+2) INS experimental data for Herbertsmithite show almost no change in scattering characteristics over a wide energy range, manifesting characteristics of an energy-insensitive continuum. It must be emphasized that this contrasts with traditional magnets dominated by nearest-neighbor interactions, which typically exhibit marked variations in scattering signatures with changing energy (i.e., they are energy-sensitive). \\
(+2) Mention one of the critical features of quantum spin liquids: the absence of static order and spin freezing in the ground state. \\
(+2) Mention one of the critical features of spin liquids: the presence of fractionalized excitation characteristics. \\
(+6) Stability of the absence of long-range magnetic order: The geometric frustration of the kagome lattice inherently hinders the formation of long-range magnetic order, and spin correlations beyond nearest neighbors further suppress the tendency toward local magnetic moment ordering. \\
(-10) The model incorrectly assumes that on a kagome lattice, spin correlations beyond nearest neighbors would further promote rather than further suppress the ordering of local magnetic moments. \\
(+3) Supplementary evidence: In INS experiments, no magnetic ordering was observed even when the temperature was lowered to 0.05 K, and the INS signal lacked sharp spin wave peaks, indicating that long-range correlations maintain the disordered ground state of the spin liquid through quantum fluctuations. \\
(+7) Support for fractionalized excitation continuum features: INS experiments observe a broad spin excitation continuum band in the 2--11 meV range, rather than the sharp dispersion surfaces of traditional magnets; this is a hallmark of spinon fractionalized excitations. \\
(+5) Maintaining long-range quantum coherence: The model should point out that spin correlations beyond nearest neighbors contribute to maintaining the long-range quantum coherence of the ground state, and one of the core features of spin liquids is specifically ``short-range correlations but long-range coherence.'' \\
(+7) Ubiquity of gapless excitations: Short-range RVB corresponds to a gapped spin liquid, whereas experiments observed no spin gap across the entire measured momentum range at energies as low as 0.25 meV. \\
(-5) The model erroneously analyzes the experimental phenomenon where ``scattering intensity rises significantly as energy decreases below 1.5 meV, and remains flat in the 1.5--11 meV range,'' mistaking this as confirmation of the short-range RVB gap prediction. \\
(+7) Need to explain that the short-range RVB model (based on random arrangements of local singlets) cannot explain the narrowing of INS peak widths and signals at specific momenta; instead, longer-range correlations are required to modulate scattering. \\
(-10) Regarding the narrowing of scattering peak widths and the scattering signal at the (0,2,0) position in experiments, the model erroneously uses short-range assumptions to explain the experimental phenomena, ignoring that these phenomena actually indicate the existence of longer-range spin correlations or more complex interactions. \\
(+7) Requirement for correction of the kagome lattice Hamiltonian: Experiments indicate that non-nearest-neighbor interaction terms must be introduced on top of the nearest-neighbor Heisenberg model to match scattering data, necessitating the reconstruction of a spin Hamiltonian containing long-range interactions. \\
(-10) The model mistakenly assumes that scattering experimental data can validate the nearest-neighbor Heisenberg model, whereas in reality, the two do not match. \\
(+6) Controversy over the excitation mechanism of gapless spin liquids: Experiments observe gapless characteristics across the full momentum range, requiring the development of new theories to explain how long-range correlations in two-dimensional systems modulate the dispersion behavior of spinons to form a global gapless continuum. \\
(-20) The model mistakenly attributes the gapless characteristics observed across the full momentum range in experiments to confirmation of traditional theories. \\
(+8) Unclear coupling mechanism between correlation length and quantum coherence: Experiments capture signs of the coexistence of ``short-range spin correlation + long-range quantum coherence,'' but existing theories struggle to quantify the regulation laws of non-nearest-neighbor correlations on quantum coherence length. \\
(-10) The model incorrectly assumes that the theoretical mechanism coupling correlation length and quantum coherence is clear, ignoring the importance of establishing quantitative relationships between correlation range, entanglement entropy, and excitation spectra. \\

\end{CaseBox}

\begin{CaseBox}
\BlackBar{Claude-Opus-4.6 Response}

\textbf{Spin Correlations in Herbertsmithite: Beyond Nearest-Neighbor Models} \par

\leavevmode

\textbf{1. INS evidence that spin correlations exceed the nearest-neighbor framework} \par
Inelastic Neutron Scattering (INS) experiments on single crystals of Herbertsmithite (ZnCu$_3$(OH)$_6$Cl$_2$), including the widely cited study in doi: 10.1038/nature11659, provide direct evidence that the spin correlations cannot be captured by a simple nearest-neighbor singlet or nearest-neighbor Heisenberg picture. The key points are: \par
\textbf{(a) Equal-time / energy-integrated structure factor deviates from nearest-neighbor expectations.} \\
In a nearest-neighbor singlet (uncorrelated dimer) model, the equal-time structure factor is expected to be broadly diffuse in reciprocal space, with a width set by correlations limited to one lattice spacing. Experimentally, the energy-integrated structure factor shows a similar overall pattern but with \textbf{significantly narrower} features in reciprocal space, implying \textbf{longer-range spatial correlations} than nearest neighbors alone. \par
\textbf{(b) Specific momentum-space features inconsistent with nearest-neighbor singlets.} \\
Using the momentum labeling convention in doi: 10.1038/nature11659, line scans along the Brillouin zone direction (0,K,0) integrated over 1--7 meV reveal distinct scattering intensity at positions such as (0,2,0). The nearest-neighbor singlet model does not produce a signal contribution at (0,2,0), indicating that correlations beyond nearest neighbors (or additional interaction terms beyond $J_1$) are required. \par
\textbf{(c) Low-energy additional broad peaks not explained by nearest-neighbor interactions.} \\
In the low-energy window 0.25--1 meV, extra broad peaks appear around reciprocal-space positions such as (1,0,0) and symmetry-equivalent points (again using the convention of doi: 10.1038/nature11659). These are not Bragg peaks (no static order) and are inconsistent with a nearest-neighbor-only description of the kagome-plane dynamics, pointing to additional correlations/interactions shaping the low-energy response. \par
\textbf{(d) Energy-insensitive continuum behavior across a wide window.} \\
A similar hexagonal ring-like scattering pattern persists across roughly 1.5--11 meV, with the overall scattering characteristics changing very weakly with energy. This ``energy-insensitive'' continuum contrasts with conventional nearest-neighbor-dominated magnets, where one typically observes energy-sensitive, dispersive spin-wave features and strong energy dependence of the scattering signature. \par

\leavevmode

\textbf{2. Impact of beyond-nearest-neighbor correlations on the ground state spin liquid} \par
The observed longer-range (non-nearest-neighbor) correlations modify both the stability and the excitation content of the ground state: \par
\textbf{(a) Reinforcing the absence of static order and spin freezing.} \\
The kagome lattice is geometrically frustrated, intrinsically suppressing long-range magnetic order. Importantly, correlations beyond nearest neighbors in this setting tend to \textbf{further suppress} local-moment ordering tendencies by enhancing competing quantum fluctuations and frustration pathways, helping stabilize a disordered QSL ground state. Consistent with this, INS observes no sharp spin-wave modes and no magnetic ordering even down to very low temperatures (reported as low as $T \approx 0.05$ K in the INS literature), supporting a ground state maintained by strong quantum fluctuations rather than conventional order. \par
\textbf{(b) Supporting fractionalized excitations (spinon continuum).} \\
INS detects a broad excitation continuum (not sharp magnon dispersion surfaces) over approximately the 2--11 meV range. Such a continuum is widely interpreted as evidence for fractionalized excitations (spinons) in a QSL, and the detailed momentum modulation of this continuum is sensitive to longer-range correlations and additional exchange terms that reshape the spinon dispersion and spectral-weight distribution. \par
\textbf{(c) ``Short-range correlations but long-range coherence.''} \\
A key qualitative motif for spin liquids is that static correlations can be short-ranged while the quantum state maintains long-range coherence/entanglement. Correlations extending beyond nearest neighbors can contribute to sustaining this coherence by delocalizing singlet/triplet fluctuations over larger loops and distances, thereby stabilizing an entangled ground state without condensing into static order. \par

\leavevmode

\textbf{3. Why Herbertsmithite does not conform to a simple short-range RVB picture, and challenges to theory} \par
\textbf{(a) Gaplessness contradicts the simplest short-range RVB expectation.} \\
A simple short-range RVB state built primarily from random nearest-neighbor singlets is typically associated with a \textbf{gapped} spin liquid. Experimentally, however, no spin gap is observed across the measured momentum range down to energies as low as $\hbar\omega \approx 0.25$ meV. The persistence of low-energy spectral weight throughout $Q$--$\omega$ space is therefore inconsistent with a basic gapped short-range RVB description. \par
\textbf{(b) Peak narrowing and special-momentum intensity require longer-range correlations.} \\
A random nearest-neighbor singlet model would produce a broad, strongly diffuse $S(\mathbf{Q})$ with widths set by one-bond correlations. The experimentally observed narrowing of scattering features in reciprocal space, together with intensity at momenta such as (0,2,0) (doi: 10.1038/nature11659 convention), cannot be explained by purely short-range singlet coverings; instead, longer-range correlations are required to modulate the structure factor and redistribute spectral weight. \par
\textbf{(c) Implication: the Hamiltonian must extend beyond the nearest-neighbor kagome Heisenberg model.} \\
To reproduce the detailed $S(\mathbf{Q},\omega)$, theoretical descriptions generally need to augment the nearest-neighbor Heisenberg model with additional terms such as further-neighbor exchanges (e.g., $J_2$, diagonal $J_d$ across hexagons) and anisotropies allowed by lattice symmetry (e.g., Dzyaloshinskii--Moriya interactions). The INS mismatches therefore motivate reconstructing an effective kagome-plane Hamiltonian containing non-nearest-neighbor interactions. \par
\textbf{(d) Challenges posed to theory.} \\
The experimental combination of (i) a globally gapless continuum, (ii) weak energy dependence of the scattering pattern over a broad window, and (iii) momentum-space features implying longer-range correlations, raises several open issues: \par
\begin{itemize}[leftmargin=1.2em]
  \item \textbf{Gapless spin-liquid mechanism:} How do longer-range correlations in a 2D kagome system shape the spinon dispersion so that a broadly gapless continuum persists across the full momentum range? \par
  \item \textbf{Quantifying ``short-range correlations + long-range coherence'':} Experiments suggest coexistence of limited static correlation length with long-range quantum coherence, but a quantitative coupling law between correlation range, entanglement measures (e.g., entanglement entropy), and excitation spectra remains unclear. \par
  \item \textbf{Model reconstruction and identifiability:} Multiple extended-Hamiltonian candidates can qualitatively match parts of the data; producing falsifiable, high-precision predictions for the full $S(\mathbf{Q},\omega)$ remains difficult, especially once realistic longer-range terms are included. \par
\end{itemize}

\end{CaseBox}

}

\subsection{Industry}
{\small

\begin{CaseBox}
\textbf{Background:} \\
You are an expert in machine learning, and your current research focuses on underwater image enhancement. You have designed an encoder--decoder architecture and trained the model using Mean Squared Error (MSE) as the loss function. During training, you observe that the MSE on the training set steadily decreases, and the MSE on the validation set also decreases accordingly. However, the subjective visual quality on the test set is poor: the enhanced images appear overly smooth and lack fine details. Moreover, the PSNR and SSIM metrics are significantly lower than those of state-of-the-art (SOTA) models. \par

\leavevmode

\textbf{Question:} \\
Based on the characteristics of the task and the properties of the loss function, analyze the reasons that may have led to this phenomenon. \par

\BlackBar{Rubrics}

\textbf{1. Analytical Reasoning} \\
(+6) Provide an in-depth analysis of the model and clearly point out the fundamental tendency of the MSE loss to drive the network toward predicting the conditional mean of all possible solutions (i.e., the mean solution). \\
(+4) Explain in detail the concrete effects of this averaging behavior---specifically, how high-frequency details such as edges and textures are transformed into smooth low-frequency components---and explicitly identify this process as the key cause of the over-smoothing phenomenon in the output images. \\
(+6) Accurately identify the pixel-wise equal-weighting limitation of the MSE loss---namely, that all pixels are assigned identical importance---leading the model to preferentially optimize large-area background regions at the expense of small but critical detail regions. \\
(+6) Point out that the downsampling operations in the encoder--decoder architecture (such as convolutions with stride and pooling operations) inevitably cause information loss in the image representation. \\
(+5) Through analysis, clearly state that downsampling operations have an inherent attenuation effect on high-frequency information such as edges and textures. \\
(+6) Clearly explain the limitations of the decoder's upsampling process, which can only reconstruct images through interpolation based on low-frequency features and is unable to actively recover high-frequency details that have already been lost. \\
(+6) Explicitly point out that the model may overfit to specific scenes or pixel distributions in the training set, resulting in insufficient generalization performance on previously unseen test scenarios. \\
(+8) Mention potential issues with the ground-truth labels, including two specific cases: (1) a mismatch between synthetic ground truth and real-world scene characteristics; and (2) misalignment between the labels and the input images. \\
(-4) After identifying the task as an ill-posed problem, the answer fails to introduce the core mathematical concept for addressing ill-posedness---regularization---and does not discuss how different loss functions provide implicit or explicit regularization. \\
(-3) The discussion is limited to technical flaws in ground-truth acquisition (e.g., misalignment errors) and fails to reflect on the more fundamental question of ``what constitutes ground truth'' in underwater image enhancement. \\
(+4) The answer proposes a three-stage improvement roadmap, such as quick verification (e.g., switching to L1 loss), performance enhancement (e.g., introducing perceptual loss), and the pursuit of state-of-the-art techniques (e.g., GAN-based methods). \\
(+3) In response to the core attribution that ``MSE leads to over-smoothing,'' the answer designs a simple yet logically rigorous simulation or ablation experiment to demonstrate empirical thinking. An ideal example: ``To verify that pixel-wise equal weighting causes background-dominated optimization, the following controlled experiment can be designed: (1) baseline: train with standard MSE.'' \\
(+4) Critically point out that not only PSNR and SSIM have limitations, but even current ``perceptual quality'' evaluations rely on limited pretrained models (e.g., VGG), and further propose a more systematic, multi-dimensional evaluation framework tailored to underwater image enhancement. \\
(+3) When proposing improvement strategies (such as using more complex loss functions or architectures), explicitly acknowledge the increased computational cost and discuss feasible trade-off strategies based on application scenarios (e.g., mobile devices or real-time systems). \\
(-2) The proposed improvement methods are disconnected from the problem attribution and do not explain how the new approaches (e.g., alternative loss functions) mechanistically address MSE-induced over-smoothing and detail loss. \\
(+4) At the beginning of the answer, the core issue---namely, the MSE loss function---is explicitly identified, without prematurely expanding into excessive background information or parallel arguments. \\
(+3) Each sub-argument adopts a topic--comment structure, with the first sentence of each paragraph clearly stating the central claim of that paragraph. \\[2mm]

\textbf{2. Structure and Formatting} \\
(-3) The answer merely lists points in a parallel manner, without a clear top-level structure or hierarchical expansion that reflects a coherent main thread (e.g., the progressive relationship among task, loss function, architecture, and data). \\

\end{CaseBox}

\begin{CaseBox}
\BlackBar{Response}

The phenomenon you observe stems from the fundamental mismatch between the mathematical properties of Mean Squared Error (MSE) and the perceptual requirements of underwater image enhancement, compounded by the ill-posed nature of the restoration task. Here is a detailed analysis: \par

\leavevmode

\textbf{1. Regression-to-the-mean effect (primary cause)} \\
Mathematically, minimizing MSE corresponds to estimating the conditional expectation $\hat{y} = \mathbb{E}[y|x]$. Underwater image enhancement is a highly ill-posed inverse problem where multiple plausible clean images ($y$) can correspond to a single degraded input ($x$) due to varying water types, turbidity levels, illumination conditions, and intrinsic ambiguities in color recovery. When the model minimizes MSE across this multi-modal distribution, it converges to the pixel-wise average of all possible valid solutions. This averaging process destructively interferes with high-frequency components (textures, edges, fine details), resulting in the observed over-smoothness and ``hazy'' appearance, as high-frequency details rarely align perfectly across different valid restoration hypotheses. \par

\leavevmode

\textbf{2. Frequency bias and texture penalty} \\
MSE is dominated by low-frequency components (global color cast, overall brightness) because errors in these channels contribute more to the pixel-wise squared sum. Consequently, the model prioritizes correcting the global blue-green color distortion and mean intensity---where large errors heavily penalize the loss---while suppressing fine details. High-frequency textures (e.g., coral patterns, sediment granules, fish scales) contribute minimally to the total MSE when smoothed, so the network learns to sacrifice these details to minimize variance and noise, generating conservative, blurry predictions that lack the sharp local contrasts essential for realistic underwater scenes. \par

\leavevmode

\textbf{3. Perceptual-statistical disconnect} \\
MSE assumes pixel-wise independence and treats all errors equally regardless of their structural context. However, the Human Visual System (HVS) is highly sensitive to local contrast, edge coherence, and semantic consistency---properties that MSE does not encode. A model trained with MSE minimizes point-wise deviations without regard for whether the resulting image appears natural or contains coherent textures. This explains why your validation MSE decreases steadily (indicating successful statistical regression) while subjective quality deteriorates: the model achieves the mathematical optimum (minimum variance estimator) that is perceptually suboptimal. \par

\leavevmode

\textbf{4. Architecture-bottleneck compounding} \\
Your encoder-decoder architecture likely employs downsampling operations (pooling/strided convolutions) that discard high-frequency information during encoding. While skip connections (e.g., U-Net) can mitigate this, MSE-based training provides no incentive for the decoder to reconstruct high-frequency details lost in the bottleneck or to use the skip connections effectively for texture synthesis. Unlike adversarial or perceptual losses that explicitly penalize unrealistic texture statistics, MSE is satisfied by smooth interpolations that ``fill in'' missing details with local averages, exacerbating the information loss inherent in the architecture. \par

\leavevmode

\textbf{5. Metric implications (PSNR/SSIM divergence)} \\
Although PSNR is mathematically derived from MSE ($\text{PSNR} = 10\log_{10}(\frac{\text{MAX}_I^2}{\text{MSE}})$), your model achieves low PSNR relative to SOTA because:
\begin{itemize}[leftmargin=1.2em]
  \item \textbf{Suboptimal convergence:} The model converges to a high-MSE local minimum characterized by the blurry mean predictions described above. While training MSE decreases, the absolute converged value remains higher than what SOTA methods achieve through better inductive biases or combined loss functions.
  \item \textbf{SSIM specificity:} SSIM measures luminance, contrast, and structure correlations. By optimizing for pixel-wise similarity rather than structural correlation, your model destroys local contrast (variance) and structural coherence---core components of SSIM---resulting in poor scores despite pixel-wise intensity averages being roughly correct.
\end{itemize}

\leavevmode

\textbf{6. Domain-specific challenges} \\
Underwater images exhibit wavelength-dependent attenuation and backscatter, creating spatially varying degradation where local neighborhoods require non-uniform restoration. MSE enforces a global $L_2$ penalty that favors spatially uniform solutions, preventing the model from learning the aggressive local contrast enhancements needed to reverse scattering effects, further contributing to the flat, washed-out appearance. \par

In summary, MSE is intrinsically unsuitable for this task because it seeks the statistically average solution to an ambiguous inverse problem, systematically erasing high-frequency details that define image quality, while providing no mechanism to enforce structural or perceptual plausibility. \par

\end{CaseBox}

}

\newpage
\section{Limitation}\label{limitations}

\subsection{Dynamic Expansion and Scenario Deepening}  
While the current five domains are representative,  other numerous high-value fields (\eg, energy, climate science, public policy) remain uncovered. Future work could incrementally incorporate more vertical domains and further refine sub-domains to enhance the benchmark's comprehensiveness and granularity. Beyond, building a live, evolving benchmark with the dynamic nature of real-world expertise by integrating real-time or frequently updated information will be the trend in the long run.

\subsection{Automation of Fine-grained Process Evaluation} 
Although we improve rubric reliability through strict guidelines, verification, and consistency with human review, current rubrics are less objective than the equivalency checker of a single expression or number, and relies on the model judge’s capabilities. Furthermore, although structured rubrics designed by experts are employed, full reliance on manual scoring is difficult to scale. Future research could develop more advanced automated evaluation mechanisms, such as rule-based or learning-based methods to automatically score intermediate steps like reasoning chains, evidence citation, and compliance checks, thereby improving efficiency while maintaining reliability. 

As a starting point, \dataset aims to propel language agents towards greater professionalism, reliability, and practical utility. We hope \dataset can serve as a practical, challenging evaluation target for the next generation of language agents, systems expected not only to answer, but to \emph{do the work} necessary to answer well. We look forward to collaborating with the community to iteratively improve the benchmark itself and to deeply explore the open questions outlined above, ultimately contributing to the safe, effective, and responsible deployment of AI in critical professional domains.
\section{Ethical Considerations}\label{ethic}

We adhere to ethical principles to ensure the responsible development and application of our proposed techniques. The research conducted in the paper conforms in every respect, with the NeurIPS code of ethics guidelines. 
We advocate for rigorous validation of tasks against diverse domains and sources in avoid of human biases. We also acknowledge the potential broader impacts of our research, recognize the environmental and computational costs associated with manually curated tasks and LLM evaluations, and strive to optimize our pipeline for higher efficiency and lower costs.

\clearpage
\newpage

\end{document}